\documentclass{book}

\usepackage[dvips]{graphicx}
\usepackage{array}
\usepackage{amsmath}
\usepackage{latexsym}
\usepackage{longtable}
\usepackage{algorithm}
\usepackage{algorithmic}
\usepackage{boxedminipage}
\usepackage{alltt}
\usepackage{named}

\newcommand{\argmax}[2]
 {\mbox{$\makebox[0mm]{\hspace{8ex}\raisebox{-1.5ex}{\scriptsize $#1$}}{\rm argmax}\,\left[#2\right]$}}
\newcommand{\prob}[1]{P(#1)}

\newcommand{\inputitem}[1]{\tt \refstepcounter{#1} \raggedright \item \makebox[0mm]{\hspace{144ex} {\rm (\arabic{#1})}}}
\newcommand{\outputitem}[1]{\refstepcounter{#1} \tt \raggedright \item \makebox[0mm]{\hspace{143ex} {\rm (\arabic{#1})}}}

\newcommand{\xpl}[1]{{\ttfamily `#1'}}
\newcommand{\dxpl}[1]{{\ttfamily #1}}
\newcommand{\hgl}[1]{\textit{#1}}
\newcommand{\prognm}[1]{\textsc{#1}}

\begin{document}

\thispagestyle{empty}

\vspace{\baselineskip}
\vspace{\baselineskip}
\vspace{\baselineskip}
\vspace{\baselineskip}
\vspace{\baselineskip}
\vspace{\baselineskip}
\vspace{\baselineskip}
\noindent{\LARGE\bf Abstract}
\vspace{\baselineskip}

\noindent
This thesis addresses automatic lexical error recovery and
tokenization of corrupt text input. We propose a technique that can
automatically correct \emph{misspellings}, \emph{segmentation errors}
and \emph{real-word errors} in a unified framework that uses both a
model of language production and a model of the typing behavior, and
which makes tokenization part of the recovery process.

The typing process is modeled as a \emph{noisy channel} where
\emph{Hidden Markov Models} are used to model the channel
characteristics. Weak \emph{statistical language models} are used to
predict what sentences are likely to be transmitted through the
channel. These components are held together in the \emph{Token
  Passing} framework which provides the desired tight coupling between
orthographic pattern matching and linguistic expectation.

The system, \prognm{ctr} (Connected Text Recognition), has been tested
on two corpora derived from two different applications, a natural
language dialogue system and a transcription typing
scenario. Experiments show that \prognm{ctr} can automatically correct
a considerable portion of the errors in the test sets without
introducing too much noise. The segmentation error correction rate is
virtually faultless.

\newpage
\thispagestyle{empty}
\mbox{}
\newpage

\thispagestyle{empty}

\vspace{\baselineskip}
\vspace{\baselineskip}
\vspace{\baselineskip}
\vspace{\baselineskip}
\vspace{\baselineskip}
\vspace{\baselineskip}
\vspace{\baselineskip}
\noindent{\LARGE\bf Acknowledgments}
\vspace{\baselineskip}

\noindent
First and foremost I would like thank my supervisor Lars Ahrenberg,
for his support, helpful discussions and encouragement to get me
through this process and reach a result. It would not have been
possible without his assistance. I would also like to thank my two
other supervisors Arne J\"{o}nsson and Ulf Nilsson.

The NLP-group is a stimulating group to work in, it provides a
positive atmosphere that I am grateful to be part of. I would like to
thank Lars Ahrenberg, Arne J\"{o}nsson, Nils Dahlb\"{a}ck, Magnus Merkel,
Lena Str\"{o}mb\"{a}ck and Lena Wigh.

Many people have been helpful to me in my research in one way or the
other. Many thanks to the brave secretaries that sustained the
ordeal of having to re-type one of the most boring texts ever written:
Lena Wigh, Eva-Britt Berglund, Lise-Lott Svensson, Barbara Ekman, Eva
Hejdeman, Carita Lilja, Britt-Inger Karlsson and Anne Eskilsson. Thank
you Bernt Nilsson for reprimanding the machines when they are
unfriendly and Ivan Rankin for valuable comments on an earlier draft
of this work.

\vspace{\baselineskip}
\noindent Thank you all!

\vspace{\baselineskip}
\vspace{\baselineskip}
\noindent
Peter Ingels

\noindent
Link\"{o}ping, December 1996

\newpage
\thispagestyle{empty}
\mbox{}
\newpage

\pagenumbering{roman}
\tableofcontents
\cleardoublepage

\pagenumbering{arabic}

\chapter{Introduction}
\label{ch:intro}

Practical Natural Language Processing (NLP) systems can not expect all
the input to conform to the grammars encoded in them. An NLP system
that is able to handle input that in some way deviates from the
language defined by the grammar encoded in the system is called
\emph{robust}. The unexpectancy of the input is due to the input being
\emph{ungrammatical} or \emph{extragrammatical}. Ungrammatical input
is judged by humans as being erroneous or strange in some way whereas
extragrammatical input contains no errors, it is just that
the input entered to the system happens to lie outside of the
grammar's coverage. The distinction is not unimportant, e.g. on the
lexical level, it would be a great help to know whether an
unrecognized string is a misspelling or a correctly spelled unknown
word. The problem of distinguishing the two cases is, however, in
general impossible to solve. In theory it is possible to write a
grammar and a lexicon that fully cover a particular language and
exclude everything that is not in the language, so researchers in
robust natural language processing tend to adopt the view that
unparsable input is ungrammatical.

Robustness is called for in all modes of language communication and in
virtually all applications that one can think of. The traditional NLP
applications with machine-readable texts and keyboard-entered input
include machine translation, information retrieval, grammar/style
checkers, text/code editing, and other NLI (Natural Language
Interface) applications such as computer-aided authoring,
computer-based language learning/tutoring and NL dialogue
systems. Speech processing applications require robustness, especially
speech recognition (speech-to-text). Pen-based interfaces (handwriting
readers) and optical character recognition (OCR) devices have to have
their output further processed to improve recognition performance. The
latter cases, where a media shift (recognition) takes place, are
particularly troublesome since the machinery performing the
recognition introduces errors which add to the human-generated errors
that were already in the first medium.

The action taken by the robust text processor in the face of
ill-formedness of course varies from application to application. A
grammar- or style-checker may highlight a portion of the input and
suggest a better way to formulate the particular passage. A
computer-based language learning system should be equipped with good
diagnostics abilities so that whatever is fed back to the learner is
informative and relevant. An NL dialogue system should be able to
perform simpler corrections to the input and enter into clarification
sub-dialogues when more serious conditions arise in order to make the
dialogue flow more naturally and minimize interruptions.

The development of the robust text processing techniques presented in
this thesis takes as a starting-point keyboard-entered input to an NL
dialogue system. An examination of a dialogue corpus showed that the
lexical errors are more urgent than other error types, where the
overall goal is to facilitate as many as possible of the user's
inputs being interpretable. Furthermore, the majority of the lexical
errors can be automatically corrected, which means that the user would
not be bothered by simple typing mistakes. The scope of this thesis is
thus the automatic correction of lexical errors, and although the
techniques presented here are not limited to text-based dialogue
systems, this application is the one assumed.

The lexical errors are broadly divided into the two major error
categories of \emph{misspellings} and \emph{segmentation errors}.
\begin{list}{\texttt ==>}{\setlength{\rightmargin}{.5cm}}
        \inputitem{utterance}What is the maintenance-cost for the
        respective models in the \hgl{aboue} table\label{utt:above}
\end{list}
Utterance\footnote{Some of the example utterances in this and
  subsequent chapters originate from the dialogue corpus mentioned
  above (These utterances have the precursor \texttt{==>}.) The
  corpus is in Swedish, so the utterances here are literal translations
  where the crucial aspect (often a lexical error) of the utterance
  has priority over good linguistic style. Utterances with the
  \texttt{-->} precursor are not literal translations,
  but have been tampered with slightly, or simply invented to better
  illustrate a particular phenomenon that is hard to literally
  translate. Hyphens that do not wrap a line indicate that
  the Swedish source token is a noun compound, e.g. the Swedish source
  token for \xpl{maintenance-cost} is \xpl{underh\aa
    llskostnad}}~(\ref{utt:above}) displays a typical
misspelling, (the highlighted \xpl{aboue}). It is an example of a so
called \emph{nonword misspelling}, i.e. the erroneous token is not to be
found in the system's vocabulary. The majority of the techniques for
automatic spelling error correction developed over the years deal with
this error type only. It is possible to form a correction
hypothesis for \xpl{aboue} by comparing it to the valid words in the
vocabulary. This approach is called \emph{isolated word error
  correction} and utilizes lexical information (the vocabulary)
only. Lexical information is, of course, of paramount importance when
recovering from lexical errors, but in general the whole range of
linguistic information is useful when correction hypotheses are
generated, particularly syntax and to some extent semantics. It is
pretty obvious that the proper correction for \xpl{aboue} in
utterance~(\ref{utt:above}) is \xpl{above}, but without the use of
contextual information there is little evidence to distinguish this
hypothesis from, say, \xpl{about}. Whereas syntactic information is
useful for handling close calls as in this example, it is absolutely
crucial when dealing with so-called \emph{real-word misspellings}.
\begin{list}{\texttt ==>}{\setlength{\rightmargin}{.5cm}}
        \inputitem{utterance}show price for volvo 300 from the \hgl{rear}
        1988\label{utt:real-word-error}
\end{list}
A word has been substituted for a token that is a valid word in the
vocabulary. It is likely that the user in~(\ref{utt:real-word-error})
intended to type \xpl{year} but it accidentally came out
\xpl{rear}. The real-word errors are obviously harder to come to terms
with than the nonword errors. Information other than lexical is
necessary just to detect the problem spot. Few researchers have
addressed the real-word error problem, and some of these have
tended to focus exclusively on this problem, forgetting the easier
nonword error problem.

The other major error category is the segmentation error
category. Segmentation errors somehow involve word boundaries. There
are two types: \emph{run-ons} and \emph{splits}.
\begin{list}{\texttt ==>}{\setlength{\rightmargin}{.5cm}}
        \inputitem{utterance}finally, can I have a list of these cars
        with information \hgl{onspaciousness}\label{utt:run-on}
        \inputitem{utterance}just the ones with \hgl{coup\'{e} space}
        3-4\label{utt:split}
\end{list}
Utterances~(\ref{utt:run-on}) and~(\ref{utt:split}) illustrate a
run-on and a split respectively. In a run-on two (or more) words have
been run together into a single token. In a split one word has been
split into two (or more) tokens. The split in~(\ref{utt:split}) should
have been written \xpl{coup\'{e}-space}\footnote{The Swedish source is
  \xpl{kup\'{e} utrymme}, which is not the same as
  \xpl{kup\'{e}utrymme}.}. As is evident from
utterances~(\ref{utt:run-on}) and~(\ref{utt:split}), the real-word --
nonword error distinction applies to run-ons and splits as well as to
misspellings. Errors involving word boundary infractions are more
difficult to handle than those that do not. Virtually all systems that
process text in any way rely on a \emph{tokenizer} to split the text
up into word tokens, where the assumption is that a sequence of
characters surrounded by space characters corresponds to a word. When
this assumption is violated things go awfully wrong since an unknown
token, like \xpl{onspaciousness} in utterance~(\ref{utt:run-on}) for example,
is assumed to be a misspelling of a single word. Furthermore, a
segmentation error can generally be `repaired' in more ways than can a
misspelling and in general more linguistic information is needed to
distinguish the good hypotheses from the bad ones. Recovering from
segmentation errors is obviously hard, and this is one of the reasons
why this problem has scarcely been addressed at all before.

This thesis aspires to take a collected approach to the entire range
of lexical errors, utilizing syntactic and to some extent semantic
information in addition to the essential lexical knowledge source. This
has not been done before and it requires in particular that correction
hypotheses pertaining to different error types can be compared for
discrimination in a meaningful way.

Inspiration can be gained from \emph{Connected Speech
  Recognition}. Consider the fictional utterance below where
  especially the segmentation problem has been exaggerated \emph{in
  absurdum}.
\begin{list}{\texttt -->}{\setlength{\rightmargin}{.5cm}}
        \inputitem{utterance}\hgl{Whtisthemaitneancecostofrtherespetivemdelsintheaboetalbe}\label{utt:absurd}
\end{list}
Finding the words in the heavily distorted~(\ref{utt:absurd}) is in
many ways similar to finding (phonemes and) words in continuous
speech. There is no indication in continuous speech as regards the end
of one word and the start of the next, i.e. there is no space
character counterpart in speech. Furthermore, since the `alphabet' in
the speech case is made up of an infinite number of `characters'
(real-valued feature vectors), a word is never `spelled' the same way
twice. The point made here is that if we agree to carry the error
types of text processing over to speech processing, we see that speech
is virtually littered with misspellings and segmentation errors. It is
therefore close at hand to see what the methods used in the difficult
speech recognition task can offer in the relatively simple
text recognition problem\footnote{One plausible interpretation
  of~(\ref{utt:absurd}) could be~(\ref{utt:above}) (with \xpl{aboue}
  substituted for \xpl{above}).}.

The following chapter describes and delineates the problem domain and
introduces the terminology used in this
thesis. Chapter~\ref{ch:corpus} ``Error Profile of a Dialogue Corpus''
describes the types and frequencies of errors found in a Natural
Language dialogue corpus. Chapter~\ref{ch:back} gives a brief review
of prior work in the area of lexical error
correction. Chapter~\ref{ch:rtr} ``An Algorithm for Robust Text
Recognition'' holds the technical contributions of this thesis. It
describes probabilistic methods used in speech processing, and how
they can be incorporated, adapted and put to use for the present
task. Chapter~\ref{ch:ae} ``Experimental Evaluation'' evaluates the
techniques described in Chapter~\ref{ch:rtr} on two error corpora
extracted from two different applications, or scenarios. The first is
a dialogue application and the other is a transcription typing
scenario. The thesis concludes with Chapter~\ref{ch:fw} ``Future
Work''.

\chapter{The Lexical Error Problem}
\label{ch:problem}

A problem faced by the analysis component of any NLP-system is that
the input sometimes does not conform to the expectations of the
system's developers. One of the reasons for this may be that the input
is erroneous, ill-formed. In this situation the system needs to react
in some way, and the proper reaction depends, amongst other things, on
the application at hand and the type of error encountered. Errors that
occur in natural language text can violate linguistic expectations on
all levels: the lexical, syntactic, semantic and the
pragmatic/discourse level. The class of lexical errors is the one that
has been most thoroughly studied, and in some respects the least
problematic one. In many cases it is possible to guess at what the
user intended to write and hence corrections can be automatically
proposed. Some of the syntactic error types can be corrected with
varying degree of success, but in general this error type calls for
alternative reactions on the system's part. Semantic and pragmatic
errors are the hardest.

In this thesis we are primarily concerned with the lexical error
problem. We look at this problem from the perspective of a Natural
Language Dialogue system, from which we have gathered a dialogue
corpus. The study of this corpus (presented in
Chapter~\ref{ch:corpus}) involves not only the lexical errors but also
certain syntactic problems. Thus the lexical error problem formulation
of this chapter in Sections~\ref{sec:prob-missp}
and~\ref{sec:prob-segm} is supplemented with
Section~\ref{sec:prob-synerr} ``Syntactic Errors'' to provide the
taxonomy of syntactic errors used in Chapter~\ref{ch:corpus}. Semantic
and pragmatic errors are not addressed
here. V\'{e}ronis~\shortcite{veronis:91} gives a
relatively comprehensive account of semantic problems
and~Carberry~\shortcite{carberry:84} describes some of
the pragmatically ill-formed phenomena that can appear in the context
of a dialogue system.

\section{Misspellings}
\label{sec:prob-missp}
Any misspelling can be described as a transformation from a correctly
spelled word performed by one or several of the \emph{basic error
  operations}:
\begin{itemize}
\item \emph{deletion} (e.g. \xpl{deltion})
\item \emph{insertion} (e.g. \xpl{innsertion})
\item \emph{substitution} (e.g. \xpl{subatitution})
\item \emph{transposition} (e.g. \xpl{trasnposition})
\end{itemize}
The basic error operations are not primitive operations, nor do they
provide a unique path from word to misspelling. Rather, their
usefulness lies in their correspondence to real world error-creating
operations and their ability to interconvert any pair of strings. The
basic error operations can be used to describe lexical errors but they
are not very good at explaining them. Distinctions are made between
\emph{typographic errors} (performance errors), \emph{cognitive
errors} and \emph{phonetic errors} (competence errors). In the case of
typographic errors it is assumed that the typist knows the correct
spelling of the word but makes a simple motor coordination slip. The
substitution of \xpl{teh} for \xpl{the} is a typical typographic
error. In the case of cognitive errors there is a lack of linguistic
competence on the part of the typist. An example of a cognitive error:
\xpl{receive} $\rightarrow$ \xpl{recieve}. A phonetic error, which is
really a sub-class of the cognitive errors, is a word that is
phonetically correct but orthographically incorrect (\xpl{memories}
$\rightarrow$ \xpl{memerys}). Although the distinctions are useful for
sorting out human spelling behavior, they are seldom used when it
comes to designing a spelling correcting program. The reason is that
it is generally quite hard to determine the underlying cause of the
error; \xpl{recieve} for example may just as well be an accidental
transposition error, and even if the cause can be established it is
not certain that it will help the spelling corrector. Another, more
important distinction is the one between nonword errors and real-word
errors.
\begin{itemize}
\item A \emph{nonword misspelling} is one that results in a string not
  in the vocabulary (of the system)
\item A \emph{real-word misspelling} is one where a valid (correctly
  spelled) word is substituted for the intended word.
\end{itemize}
The substitution of \xpl{wether} for
\xpl{whether} in \xpl{\hgl{wether} to be or not to be\ldots} is an
example of a real-word error. By definition a real-word error can not
be detected by the use of a system's lexical knowledge. A syntactic
analysis may or may not detect the error depending on the relationships
between the words. Apart from detecting real-word errors, the context
is often helpful in deciding amongst alternative correction candidates
for nonword errors. The traditional way of correcting the detected
nonword error usually involves some sort of distance metric,
cf.~\cite{kukich:90}. The metric is used to compare the erroneous
token to the valid words in the dictionary, and to choose the word
that is closest to the misspelling. The problem with many of these
algorithms is that in many cases there are several candidates that are
equally `close'. In \xpl{pass me the \hgl{salr} please}, the erroneous
\xpl{salr} should probably be \xpl{salt}, but viewed in isolation
\xpl{sale} is just as likely. In these situations contextual
dependencies are useful.

The example of \xpl{\hgl{wether}} above hints at a problem related to
spelling errors, namely that of the \emph{unknown word problem}. In
most NLP-systems `wether' would not be part of the vocabulary, and hence
it would not be a real word but rather a nonword error. In this
particular example it causes no problem since the intended word was
`whether', but if `wether' was the intended word, as in: \xpl{Is that our
  wether grazing over there?}, problems arise. The tricky bit is to
decide whether `wether' is a misspelling of a known word or if it is a
correctly spelled word that just happens to lie outside of the
dictionary's coverage. Extragrammatical problems such as the unknown
word problem is not further addressed in this thesis, except in the
discussion in Chapter~\ref{ch:fw} ``Future Work''.

Another distinction that is often made between misspellings is that of
single error misspellings and multiple error misspellings.
\begin{itemize}
\item A \emph{single error misspelling} is an error where one of the
four basic error types has occurred exactly once 
\end{itemize}
and consequently
\begin{itemize}
\item A \emph{multiple error misspelling} contains more than one
  instance of the four basic error types.
\end{itemize}
The distinction may seem a bit strange, but it was discovered already
in the sixties (see below) that a large portion of all spelling errors
was due to single error misspellings, and thus many of the techniques for
automatic spelling error correction developed over the years have
focused on this error type. If for no other reason, this makes the
distinction interesting for comparative purposes.

When examining studies of spelling error corpora it is important to
notice that spelling error frequencies and error patterns vary
significantly between different applications.

In an early study Damerau~\shortcite{damerau:64} found that as many as
80\% of the words rejected by the list of acceptable terms in an
information retrieval system were single error misspellings. The
term-list will only reject nonword errors of course, so the remaining
20\% are presumably multiple error misspellings
(Damerau does not mention segmentation errors).

Kukich~\shortcite{kukich:92a} made an error profile of a 40,000 word
TND\footnote{TND is a service that AT\&T provides for their speech-
  and hearing-impaired customers. A person can have a TDD
  (Telecommunications Device for the Deaf) device with which she can
  communicate with other people with speech- and/or
  hearing-impairments who also have a TDD. A TDD is much like a
  terminal that can be hooked up to the wire via a modem, and
  lets the user type messages on a keyboard and receive messages on a
  screen. A TDD user can also communicate with a voice phone user by
  calling a deaf relay center, where a relay operator, using both a
  TDD and a voice phone, reads the text typed by the TDD user to the
  voice phone user and listens to the voice phone user's response and
  types the words spoken by the voice phone user back to the TDD user.}
(Telecommunications Network for the Deaf) transcript corpus. She found
78\% of the nonword errors in the corpus to be single error
misspellings, corroborating the findings of Damerau. Some 18\% of the
misspellings contained a mistake in the first character position. It
is generally believed that errors tend not to occur in the first
character position, and the figure may be unrepresentatively
high. Twentyseven percent contained errors involving an adjacent
key. The layout of the keyboard is, of course, relevant when motor
coordination slips occur. Two percent were phonetic errors: \xpl{cuz},
\xpl{becuz}, \xpl{u}, \xpl{ur} and \xpl{rite}. Phonetic error is
perhaps not the most adequate characterization for these words. The
slang-like abbreviations can be seen in informal conversations over
the net, and unknown word may be a better term.

Pollock and Zamora~\shortcite{pollock+zamora:83} collected
over 50,000 nonword misspellings from around 25,000,000 words of text
from a number of different scientific and scholarly
databases. They found, amongst other things,
that the multiple error frequency was as low as 7.5\%. The rest of the
nonword misspellings were distributed over the four basic error
types: deletions 34\%, insertions 27\%, substitutions 19\% and
transpositions 12\@.5\%. The low error frequency (0.2\%) is probably
partly due to the nature of the text source.

Mitton~\shortcite{mitton:87} studied the spelling behavior of
secondary school pupils (15 years of age) in a corpus of short essays
on the subject of ``Memories of my primary school''. The essays were
hand-written. The corpus included 924 essays and a total of 170,016
words. Mitton recorded 4,218 errors, an error rate of only 0.25\%. The
low error frequency can partly be explained by the fact that the
essays were written by hand, thus excluding the keyboard
error-source. It is not clear from Mitton's report, but it appears as
though the pupils were aware of the fact that they were being tested on
their spelling skills. This may also have had an affect on the error
frequency (perhaps in either way). Mitton reports a relatively high
percentage of real-word errors. He found 40\% to be real-word errors
and consequently 60\% nonword errors. In the real-word error category
Mitton includes errors that are generally considered to be syntactic
errors, such as present-tense verbs in place of past (\xpl{the best
thing I \hgl{like} was to play}) and number agreement errors (\xpl{five
other primary \hgl{school}}). Excluding these errors from the
real-word error category, the percentage is still as high as
30\%. When a system's (incomplete) dictionary is used there will be a
smaller amount of real-word errors and instead a corresponding amount
of unknown words. Another remarkable finding in Mitton's report is the
unprecedented high frequency of phonetic errors. Homophones and
near-homophones made up close to 60\% of the errors in the
corpus. Near-homophones are words like \xpl{where} and \xpl{were}
which are homophones to some people and \xpl{have} and \xpl{of}
which can be homophones in \xpl{I might \hgl{of} done}. The high frequency
of homophones is hard to explain, but perhaps it has to do with the fact
that the pupils were given a spelling transcription test before they
were asked to write the essay. This might have put the pupils in a
``sound-to-text spelling mode'', which might explain the high
homophone frequency.

Peterson~\shortcite{peterson:86} set out to determine the probability
of a word being mistyped as another word, as a function of the size
of the word-list. With the terminology introduced in this section this
means: What is the probability that a single basic typographic
error will result in a real-word error when the size of the word-list
varies? Due to lack of reliable statistics the four basic error
types were assumed to be equally likely. For an $n$-letter word and 28
characters in the alphabet (26 letters, hyphen and apostrophe) there
are $n + 28(n+1) + 27n + n-1 = 57n+27$ possible single error
misspellings\footnote{deletions, insertions, substitutions and
transpositions respectively.}. A word-list of 369,546 words was run
through a program that generated all the possible mistypings for each
word. Of the possible 205,480,845 mistypings 988,192, or 0.5\%, turned
out to be another valid word. Variably sized word-lists were
constructed by stripping off the most infrequent words from the
original word-list. The smallest word-list would thus contain only the
most frequent words, which are generally shorter words: \xpl{the},
\xpl{of}, \xpl{and}, \xpl{to}, \xpl{a}, \xpl{in}, \xpl{is},
\xpl{that}, \xpl{it} and \xpl{he} make up the top ten. The word
frequencies were estimated from different corpora. Short
words are more likely to be real-word errors if mistyped than longer
words. By weighing the fraction of all possible errors that are
real-word errors against the expected frequency of occurrence, and
varying the size of the word-list, Peterson found that in running text
the expected frequency of single error real-word errors caused by
typographic mistakes could be as high as 16\%. The effect of the more
frequent shorter words is apparent in that the 100,000 most frequent
words resulted in 13\% real-word error probability, while the remaining
300,000 words only added a further 3\%.

\section{Segmentation Errors}
\label{sec:prob-segm}
Segmentation errors are errors in which word boundary markers are
involved. The most prominent of the boundary markers is the space
character, whose sole objective is to delimit words. The space
character is a member of the white-space characters which also include
tab, carriage return and line-feed.  These characters also have other
functions besides delimiting words, and the same goes for parenthesis,
period, comma, slash etc.

The basic error operations of deletion, insertion, substitution and
transposition when applied to word boundaries result in things like:

\begin{list}{\texttt -->}{\setlength{\rightmargin}{.5cm}}
        \inputitem{utterance}He \hgl{gaveher} roses\label{utt:prob-del}
        \inputitem{utterance}He \hgl{ga ve} her roses\label{utt:prob-ins}
        \inputitem{utterance}He \hgl{gavehher} \hgl{ro es}\label{utt:prob-subs}
        \inputitem{utterance}He \hgl{gav eher} roses\label{utt:prob-trans}
\end{list}

The segmentation errors in~(\ref{utt:prob-del})
through~(\ref{utt:prob-trans}) are more or less likely to appear in
actual texts. In general~(\ref{utt:prob-del}) and (\ref{utt:prob-ins})
are more common than~(\ref{utt:prob-subs}) and (\ref{utt:prob-trans})
since they are not only caused by accidents, but also arise from cognitive
and phonetic misconceptions. Segmentation errors can, just as regular
misspellings, be diagnosed as cognitive, phonetic and typographic
errors. The examples~(\ref{utt:prob-del}) through
(\ref{utt:prob-trans}) are obviously all typographic errors. The
accidental substitution of \xpl{h} for \xpl{\symbol{32}} and
\xpl{\symbol{32}} for \xpl{s} in~(\ref{utt:prob-subs}) is probably
unlikely to occur very frequently in actual texts because of the form
and placement of the space-bar; it is unlikely that one accidentally
hits the space-bar instead of one of the character keys or vice
versa. Example~(\ref{utt:prob-del}) and the first erroneous token
in~(\ref{utt:prob-subs}) are called
run-ons. Example~(\ref{utt:prob-ins}), the second erroneous token
in~(\ref{utt:prob-subs}) and example~(\ref{utt:prob-trans}) are called
splits.
\begin{itemize}
\item A \emph{run-on} segmentation error has occurred when two or more words
  are written as one token, and it is a \emph{multiple run-on} if more
  than two words are involved.
\item A \emph{split} segmentation error has occurred when a word is
  written as two or more tokens, or, when two words are split up into
  two erroneous tokens, and it is a \emph{multiple split} if
  more than two tokens are involved.
\end{itemize}
The Example~(\ref{utt:prob-trans}) requires special
consideration. From the point of view of the basic error operations it
is quite arbitrary whether~(\ref{utt:prob-trans}) should be counted as
a run-on or as a split, or as a combination of both, but since the
result of the transposition is two tokens the best way to describe
example~(\ref{utt:prob-trans}) is as a split and hence the definition
above.

Run-ons and splits may, of course, like misspellings result in
real-word errors. Particularly in the case of splits that are caused
by cognitive and/or phonetic misconceptions it is quite likely that at
least one of the tokens is a real word
(\xpl{already}$\rightarrow$\xpl{al ready}).

\begin{itemize}
\item A \emph{real-word run-on} has occurred when the token is a valid
    word in the vocabulary.
\item A \emph{real-word split} has occurred when at least one of the
  affected tokens is a valid word in the vocabulary.
\end{itemize}

The problem of detecting and correcting segmentation errors has
received much less attention than that of misspellings. A reason for
this is that segmentation errors are less frequent than spelling
errors in naturally occurring texts. Another reason, and most likely
the main reason, is the complexity inherent in the segmentation
problem. In the case of segmentation errors context is not only
helpful, as with most misspellings, but essential.

Virtually all spelling correction techniques rely on a tokenizer to
split the character input stream into tokens. Tokenizers are often
simple programs that just look for white-space characters and other
delimiters to determine token boundaries. The assumption is that a
token corresponds to a word in the vocabulary. When the input stream
has been tokenized, each token is checked against a dictionary of
valid words and if the token is not in the word-list, it is flagged as
an error, i.e. an error involving \emph{one word}. In actuality this
means that \emph{the tokenizer defines the problem-spot}. Assume for a
moment that the tokenizer is right in its assumption that the
unrecognized token is the misspelling of exactly one word. As
mentioned above, the traditional approach to spelling error correction
relies on the computation of the \emph{distance} between the erroneous
token and the words in the vocabulary. With an $M$ word vocabulary,
$M$ distances have to be computed. However, if the constraint enforced
by the white-space during tokenization is relaxed, the error
correction task becomes considerably more complex. Turning again to
utterances~(\ref{utt:prob-del}) to~(\ref{utt:prob-trans}), if a single
erroneous token appears in the input stream it might be a run-on. If
the token contains $n$ characters it can be split in $n$ ways and
each splitting results in two candidate tokens. If any of the
splittings results in two perfect matches,
(utterance~(\ref{utt:prob-del})) this yields a plausible
correction. However, it is possible that perfect matches are
unobtainable (the first erroneous token in
utterance~(\ref{utt:prob-subs})). In this case both candidate tokens
of the hypothetical run-on have to be compared to the $M$ words of the
vocabulary resulting in $2nM$ distance computations. The same line of
reasoning can, of course, be applied to splits. The problem that faces
the error recovery program is that it can not make any assumptions
regarding the type of the error if it wants to be sure to find the
best correction alternative. When misspellings, run-ons and splits,
along with single as well as multiple errors, and even combinations of
misspellings and segmentation errors, are considered, the traditional
approach to lexical error recovery has lost its applicability. This is
most likely the main reason why so little interest has been shown in
segmentation errors.

In addition to this there is the real-word segmentation error problem which
means that legal words surrounding the unrecognized token(s) have to
be taken into account in the generation of corrections as
well. Consider the pathological laboratory sentence:

\begin{list}{\texttt -->}{\setlength{\rightmargin}{.5cm}}
        \inputitem{utterance}Her an together be fore shere ached these
        a\label{utt:prob-path}
\end{list}

This mumbo-jumbo sentence contains one nonword, \xpl{shere}, the rest
of the words are all legal. The problem of
resegmenting~(\ref{utt:prob-path}) is quite hard, even for humans. The
white-space characters carry a significant amount of information and
when they are misplaced, interpretation gets hard. Although the
segmentation problem never gets as hard as in~(\ref{utt:prob-path})
in processing of actual texts, there are other applications that need
to deal with similar situations. In speech recognition, OCR and
handwriting decoding the segmentation problem is a primary
concern. (In the two latter cases character segmentation is actually
more acute than word segmentation.) Speech recognition suffers from
problems of coarticulation, homophones and, of course, there is no
space character counterpart in speech. A speech-like reformulation
of~(\ref{utt:prob-path}) could be written as:
 
\begin{list}{\texttt -->}{\setlength{\rightmargin}{.5cm}}
        \inputitem{utterance}Herantogetherbeforeshereachedthesea\label{utt:prob-speech}
\end{list}

The segmentation of~(\ref{utt:prob-path}) and (\ref{utt:prob-speech})
that is likely to be the `intended' one is:

\begin{list}{\texttt -->}{\setlength{\rightmargin}{.5cm}}
        \inputitem{utterance}He ran to get her before she reached the
        sea\label{utt:prob-pathcorr}
\end{list}

Segmentation problems are less frequent in texts than misspellings. In
many applications, however, it is a problem that needs to be
addressed.

In her study of the nonword errors in the 40,000 word TND corpus
Kukich~\shortcite{kukich:92a} found that 13\% of the errors were
run-ons and 2\% were splits. Apparently the majority of the
segmentation errors were caused by typographic slips such as
\xpl{yesthisis} and \xpl{sp ent}. The investigation also indicates
that a high percentage of the run-ons involve a relatively small set
of high-frequency function words.

In Mitton's~\shortcite{mitton:87} study of errors in 15-year-olds'
essays, segmentation errors were also considered. Contrary to Kukich,
Mitton found splits to be more common than run-ons. Fourteen percent
of the errors were splits but there were only 3\% run-ons. Whereas it
was very rare for the run-ons to result in a real word, there were
only five cases (0.12\%) where a split word did \emph{not} result in
neither of the halves being a real word. In 13\% of the cases both
halves were real words and in the remaining cases (1\%) one of the
halves was a real word (\xpl{to gether} and \xpl{evry body}
e.g.). Mitton also found that the (generally shorter) function words
were more liable to result in real-word errors than the content
words. The real-word error portion of the function words in error was
66\%, and the corresponding figure for the content words was 33\%.

\section{Syntactic Errors}
\label{sec:prob-synerr}
The aim of this section is to introduce the error categories that are
used to diagnose the error corpus in Chapter~\ref{ch:corpus}, not to
fully cover all the syntactic errors and peculiarities that can appear
in texts. For a more in-depth and linguistically-oriented account of
these issues see, for example,
Baker~\emph{et al.}~\shortcite{baker+al:90},
V\'{e}ro\-nis~\shortcite{veronis:91},
Carbonell and Hayes~\shortcite{carbonell+hayes:83},
Kwasny and Sondheimer~\shortcite{kwasny+sondheimer:81},
Hayes and Mouradian~\shortcite{hayes+mouradian:81}.

Syntactic errors depend on the structural relationships between words
in a sentence. At the surface level however, a sentence is a sequence
of words just as a word is a sequence of characters. From this
standpoint it is natural to simply upgrade the basic lexical error
operations to the syntactic level. This results in \emph{the basic syntactic
error operations}:
\begin{itemize}
\item \emph{missing word} e.g. \xpl{the plays in the backyard}
\item \emph{extra word} e.g. \xpl{the \hgl{the} boy plays in the backyard}
\item \emph{substituted word} e.g. \xpl{the boy plays \hgl{but} the backyard}
\item \emph{transposed words} e.g. \xpl{the boy plays \hgl{the in} backyard}
\end{itemize}
Just as any misspelled word can be transformed into a legal word by
combinations of the basic lexical error operations, so can an
ungrammatical sentence with the basic syntactic error operations. For
practical purposes, however, this classification is too crude. In the
typology used here the basic error operations will account primarily
for performance errors. Most of the examples above are likely to be
accidental, possibly with the exception of the substituted word error. A
variant of the substituted word error, which is generally a competence
error, is the agreement error.
\begin{itemize}
\item \emph{Agreement errors} include:
\begin{itemize}
\item \emph{subject-verb number and person errors}, e.g. \xpl{\hgl{they
      plays} in the backyard}
\item \emph{wrong case of pronouns}, e.g. \xpl{\hgl{them} play in the
    backyard}
\item \emph{noun phrase problems}, e.g. \xpl{they play in \hgl{a backyards}}
\end{itemize}
\end{itemize}

A subtle distinction has been introduced here regarding the
real-word misspelling (see Section~\ref{sec:prob-missp}) and the
substituted word error. Both error types are concerned with the
substitution of one word for another and both are performance
errors. The difference is that the real-word misspelling is a
\emph{misspelling}, i.e. it is an error on the character level. The
substituted word error is something other than a misspelling and it is
an error on the vocabulary level. Hopefully the confusion can be
reduced by an example of a substituted word error:
\begin{list}{\texttt ==>}{\setlength{\rightmargin}{.5cm}}
        \inputitem{utterance}Is there a medium-sized-car in the
        50000-70000\\ price-range \hgl{and} has year-of-production
        1988?\label{utt:prob-subst}
\end{list}
This strange looking utterance would read perfectly well if \xpl{and}
was replaced with \xpl{that} or \xpl{which}. It is hard to say what
the user was thinking about when she typed~(\ref{utt:prob-subst}),
but one thing is clear and that is that \xpl{and} is not \xpl{that} or
\xpl{which} misspelled. Thus~(\ref{utt:prob-subst}) is an example of
the substituted word error type. Utterance~(\ref{utt:real-word-error})
in Chapter~\ref{ch:intro} is an example of a real-word error.

A varied and wide-ranging phenomenon, that is also quite frequent in
the corpus examined in the following chapter, is the elliptic
utterance. The elliptic utterances are not really errors, since they
are generally unproblematic to understand in the context in which they
appear, at least for humans. In some sense, however, they are errors,
they violate an imagined \emph{core grammar} of how legal declarative,
imperative and interrogative sentence are formed in a
language. Henceforth these sentences will simply be referred to as
elliptic, avoiding having to classify them as being ill- or
well-formed.

Various distinctions can be made between different types of ellipses
(cf. \cite{carbonell+hayes:83,lavelli+stock:90}). For our purposes,
however, two types will suffice: the \emph{telegraphic ellipsis} and
the \emph{contextual ellipsis}. In a telegraphic ellipsis one or, more
often, several words have been \emph{intentionally} left out. It is
generally easy to state that a fragment has been left out on purpose,
and this makes it easy to distinguish the telegraphic ellipsis from
the missing word error type. Further, the elliptic fragment, the
left-out portion, is always semantically redundant. Two examples of
telegraphic ellipses, one at each end of the scale:

\begin{list}{\texttt ==>}{\setlength{\rightmargin}{.5cm}}
        \inputitem{utterance}looking for other models in the same
        price-range that would not be expensive to
        maintain\label{utt:prob-synte1}
        \inputitem{utterance}mercedes fuel-consumption\label{utt:prob-synte2}
\end{list}
The subject and the copula have been left out
in~(\ref{utt:prob-synte1}) and many people would not consider it an
error. In~(\ref{utt:prob-synte2}) it is more obvious that fragments
have been left out. However, in the dialogue context in which it
appears, it is unproblematic to interpret the input: \xpl{What is the
fuel-consumption for mercedes}.

In contextual ellipses the elliptic fragment is also left out
intentionally, but is not semantically redundant, it can be inferred
from the context. In a dialogue system, it can usually be inferred
from one of the immediately preceding utterances, which may be system
or user generated. An example of a contextual ellipsis:

\begin{list}{\texttt ==>}{\setlength{\rightmargin}{.5cm}}
        \inputitem{utterance}What is the impact-safety for the cars
        with rust-protection better than 3\label{utt:prob-synce1}
\end{list}

{\ttfamily\textit{The system responds\ldots}}

\begin{list}{\texttt ==>}{\setlength{\rightmargin}{.5cm}}
        \inputitem{utterance}better than 4\label{utt:prob-synce2}
\end{list}

When the second utterance is entered, it is not hard to understand what
is meant, basically the portion \xpl{What is the impact-safety for
  the cars with rust-protection} should be put first in the second
utterance to yield the intended query.

\chapter{Error Profile of a Dialogue Corpus}
\label{ch:corpus}
The need for robust text processing techniques became apparent during
the development of the natural language dialogue system
\textsc{linlin}~\cite{AJD:90,Jonsson:95}. \textsc{linlin} is a
dialogue interface to a database, and it accepts queries in natural
language (Swedish) and produces \textsc{sql}-queries that are fed to
the DBMS\@. \textsc{linlin} can be adapted to different application
databases. In order to try to determine the required functionality of
the dialogue management module, a series of experiments was conducted
using the \emph{Wizard of Oz} data collection scheme,
cf.~Dahlb{\"a}ck~\emph{et al.}~\shortcite{DJA:93}. The idea behind the Wizard
of Oz technique, in short, is to let an \emph{operator}, a human
being, perform the task of the dialogue interface. In this case the
operator interprets the NL input, she then accesses the
\textsc{sql}-database and relays the database output to the subject,
or she herself replies with canned text or manually types back a short
reply.

In the following two sections error profiles are given of two
different corpora collected using the Wizard of Oz technique. The
motivation for the study is to get an idea of the error frequencies in
this sort of application. How are the errors distributed over the
error types? What is the most `urgent' robustness functionality? Which
linguistic knowledge sources are in the foreground for
resolving/interpreting the different types of ill-formedness?

The corpora (and the sections) are named \textsc{cars} and
\textsc{travel}. Slightly more attention will be devoted to
\textsc{cars} since it is one of the corpora that have been used for
evaluating the techniques presented in this thesis (see
Chapter~\ref{ch:ae}). In the \textsc{cars}-application the database
contains information on different car models. The subject can retrieve
information about the model's price, fuel consumption, top speed
etc. The task given to the subject is to decide which car to buy given
certain financial restrictions. The \textsc{travel}-application is a
travel agency scenario. The subject's task is to decide on a charter
trip to the Greek archipelago. She has to choose an island and a
hotel, when to travel etc. The experimental setups differ slightly in
the two cases. In \textsc{cars} the operator reads the subject's input
from the screen and constructs the corresponding \textsc{sql}-query
that is submitted to the DBMS. The output from the database is simply
forwarded to the subject's screen. The output is thus generally in
table format. Occasionally the subject asks for clarification of table
output, and sometimes the subject queries the `system' for information
that is not included, in which case the operator replies using canned
text or hand-typed short messages. In \textsc{travel} no actual
database system is involved, the operator simply manipulates a large
collection of canned texts organized in such a way that the operator
can simulate a database system interface. The \textsc{travel}
interface also exhibits some degree of multimodality in that the
operator is able to display maps of many of the tourist locations.

In both experimental setups and for all subjects the operator is
instructed to be `forgiving' with respect to ill-formed input,
i.e. the operator will respond to all queries that she can
understand. Note that the data collection was not carried out with
ill-formed input as a topic of investigation.


\section{\sc cars}
\label{sec:corpus-cars}
The \textsc{cars} corpus contains 20 dialogues gathered from 20
different subjects. Ten of the subjects were led to believe that they
were communicating with an actual natural language interface, and the
other ten were informed of the fact that the interface was simulated
by an operator. As far as ill-formed input is concerned the two
sub-corpora are quite similar. Below the sub-corpus collected with the
misled subjects is called \textsc{machine}, and the other half is
called \textsc{operator}, the distinction reflecting the subject's
beliefs. The two sub-corpora are presented separately in the tables
below for transparency. \textsc{cars} contains 369 user utterances,
3,139 word tokens and 584 word types. Table~\ref{tab:c-utt} shows how
many of the utterances are well-formed, elliptic and how many contain
at least one error.
\setlength{\extrarowheight}{2pt}
\begin{table}[h]
\centerline{
\begin{tabular}[H]{||l|rr|rr|>{\bfseries}r>{\bfseries}r||}
\hline
& \multicolumn{6}{c||}{Corpus} \\
\raisebox{1.5ex}[0cm][0cm]{Utterances} & \multicolumn{2}{c}{\textsc{machine}}
& \multicolumn{2}{c}{\textsc{operator}} &
\multicolumn{2}{c||}{\textsc{cars}}\\ \hline
Well-formed       & 116 & 70\% & 132 & 65\% & 248 & 67\% \\ 
Elliptic          & 13  & 8\%  & 24  & 12\% & 37  & 10\% \\ 
Ill-formed        & 37  & 22\% & 47  & 23\% & 84  & 23\% \\ \hline\hline
{\bfseries Total} & {\bfseries 166} & {\bfseries 100\%} & {\bfseries
  203} & {\bfseries 100\%} & {\bfseries 369} & {\bfseries 100\%} \\ 
\hline
\end{tabular}
}
\caption{The distribution of utterances\label{tab:c-utt}}
\end{table}

Of the 121 ($37+84$) ill-formed and elliptic utterances 94 (78\%)
contain one elliptic or erroneous construction and 27 (22\%) contain
more than one. In the 121 ill-formed and elliptic utterances there is
a total of 161 individual errors/ellipses in \textsc{cars}. The distribution
over the error types is displayed in Table~\ref{tab:c-errs}.

\begin{table}[ht]
\centerline{
\begin{tabular}[H]{||lll|rr|rr|>{\bfseries}r>{\bfseries}r||}
\hline
&&& \multicolumn{6}{c||}{Corpus} \\
\multicolumn{3}{||c|}{\raisebox{1.5ex}[0cm][0cm]{Error Types}} & \multicolumn{2}{c}{\textsc{machine}}
& \multicolumn{2}{c}{\textsc{operator}} & \multicolumn{2}{c||}{\textsc{cars}}\\ \hline
& \multicolumn{2}{l|}{Misspellings}              & 22 & 31\% & 40 & 45\%  & 62 & 39\% \\
& \multicolumn{2}{l|}{Run-ons}                   & 14 & 19\% &  3 &  3\%  & 17 & 11\% \\
& \multicolumn{2}{l|}{Splits}                    &  7 & 10\% &  6 &  7\%  & 13 &  8\% \\
\multicolumn{2}{||l}{Lexical errors $(\sum)$}   && 43 & 60\% & 49 & 55\%  & 92 & 57\% \\ \hline\hline
& \multicolumn{2}{l|}{Missing Constituent}       &  4 &  6\% &  0 &  0\%  &  4 &  2\% \\
& \multicolumn{2}{l|}{Extra Constituent}         &  0 &  0\% &  0 &  0\%  &  0 &  0\% \\
& \multicolumn{2}{l|}{Substituted Constituent}   &  1 &  1\% &  4 &  4\%  &  5 &  3\% \\
& \multicolumn{2}{l|}{Transposed Constituent}    &  0 &  0\% &  0 &  0\%  &  0 &  0\% \\
& \multicolumn{2}{l|}{Agreement Error}           &  6 &  8\% &  9 & 10\%  & 15 &  9\% \\
\multicolumn{2}{||l}{Syntactic errors $(\sum)$} && 11 & 15\% & 13 & 15\%  & 24 & 15\% \\ \hline\hline
& \multicolumn{2}{l|}{Telegraphic Ellipsis}      &  7 & 10\% & 18 & 20\%  & 25 & 16\% \\
& \multicolumn{2}{l|}{Contextual Ellipsis}       & 11 & 15\% &  9 & 10\%  & 20 & 12\% \\
\multicolumn{2}{||l}{Ellipses $(\sum)$}         && 18 & 25\% & 27 & 30\%  & 45 & 28\% \\ \hline\hline  
\multicolumn{2}{||l}{{\bfseries Total}} && {\bfseries 72} & {\bfseries 100\%} & {\bfseries 89} & {\bfseries 100\%} & {\bfseries 161} & {\bfseries 100\%} \\
\hline
\end{tabular}
}
\caption{The distribution of errors and ellipses in \textsc{cars}\label{tab:c-errs}}
\end{table}

The relatively small size of the corpus makes the error frequencies
sensitive to individual subjects' strange behavior. Ten of the 14
run-ons in \textsc{machine}, for example, are the work of two
particularly careless subjects. One of these utterances reads:

\begin{list}{\texttt ==>}{\setlength{\rightmargin}{.5cm}}
        \inputitem{utterance}I want to see cars \hgl{inprice-range} 20000 to
        70000 of \hgl{makesaudi},bmw,ford,mazda,\hgl{toyotapeugeotvolkswagen}\label{utt:prob-topevo}
\end{list}
The fluctuations between \textsc{machine} and \textsc{operator} are
probably due more to the choice of subjects for the respective
experimental setups than to the setups themselves.

Utterance~(\ref{utt:prob-topevo}) contains three run-ons, the first
two being single errors and the last a multiple error. Note that the
absence of spacing in the comma-separated enumeration of the makes
\xpl{bmw}, \xpl{ford} and \xpl{mazda} does not constitute an
error. Although the enumeration violates typographic conventions
regarding spacing in conjunction with punctuation characters, a
tokenizer can resolve this problem by simply triggering on characters like
comma.

A system with no robustness built into it, and without the ability to
deal with elliptic constructions, could theoretically parse 248 (67\%)
of the utterances in \textsc{cars} (from Table~\ref{tab:c-utt}). It is
interesting to see how the theoretic performance of the system improves
when robustness functionality is added.

Envisage three robust modules: an automatic spelling and segmentation
error correction module, a module that interprets telegraphic and
contextual ellipses and a module that can recover from the basic
syntactic errors of missing constituent, extra constituent, substituted
constituent, transposed constituent and also the agreement errors. The
figures in Table~\ref{tab:c-module} show the theoretical performance
improvements brought on by these modules had they been used on
\textsc{cars}. A small number of utterances are hard to interpret for
reasons other than those discussed here. One utterance contains a
definite reference to objects not displayed or mentioned before, and
there are three occurrences of prematurely ended inputs. The single
word utterance~(\ref{utt:prob-singleword}) is an example of the latter
case.

\begin{list}{\texttt ==>}{\setlength{\rightmargin}{.5cm}}
        \inputitem{utterance}show\label{utt:prob-singleword}
\end{list}

All utterances are included in Table~\ref{tab:c-module}
although these oddities are disregarded.

\begin{table}[ht]
\centerline{
\begin{tabular}[H]{||l|rr||}
\hline
Robust Module & \multicolumn{2}{c||}{\textsc{cars}}\\ \hline
None        & 248     & 67\%  \\ \hline
Lexical     & +58=306 & 83\% \\ \hline
Syntactic   & +14=262 & 71\% \\ \hline
Ellipses    & +36=284 & 77\% \\ \hline\hline
{\bfseries All Modules} & {\bfseries +121=369} & {\bfseries 100\%} \\ \hline
\end{tabular}
}
\caption{Utterances theoretically parsable with different robustness
  capacities\label{tab:c-module}}
\end{table}

Note that the figures in Table~\ref{tab:c-module} require the modules
to have 100\% both recall and precision. The figures do not properly
add up because there are utterances that contain instances of
different error types, and hence can not be completely corrected by
any single module in isolation.

Table~\ref{tab:c-module} strongly emphasizes the need for robustness
in this type of application. The non-robust system has a theoretic
parsability maximum of 67\%, and if the system can resolve elliptic
constructions it is 77\%. The single, potentially most productive
module is the lexical error recovery module. It is noteworthy that a
system with the ability to handle ellipses and lexical errors has its
maximum at 94\% ($(248+58+36+6)/369$). (There are six utterances that
contain both lexical errors and ellipses but no syntactic error.) The
lexical errors are therefore the most interesting error category, at
least from the point of view of building a useful dialogue system.

The lexical error rate, the number of error tokens per token, in
\textsc{cars} is $2.9\%$. The lexical errors can be orthogonally
divided into non/real words and single/multiple
errors. Table~\ref{tab:c-lexcloseup} shows how the lexical errors are
distributed over these categories. Misspellings are naturally divided
into the four categories since a misspelled word is always realized as
a single token regardless whether it is a single, multiple, nonword or
real-word error. Segmentation errors are not as straightforward
(cf. Section~\ref{sec:prob-segm}). The ratio of single error
misspellings among the nonwords (57\%) is considerably lower than the
80\% reported by Damerau~\shortcite{damerau:64}. The real-word error
frequency (7\%) could be on the lower side and it is certainly below the
frequency reported by Mitton~\shortcite{mitton:87}, but then again,
Mitton's findings may not be representative for an application such as the
present one. The real-word error splits are of two kinds: strange use
of slash
(\xpl{petrol/mile}$\rightarrow$\xpl{\hgl{petrol\symbol{32}/\symbol{32}mile}},
the multiple error in the table, and
\xpl{dollars/year}$\rightarrow$\xpl{\hgl{dollars/\symbol{32}year}}),
and split number notation
(\xpl{40000}$\rightarrow$\xpl{\hgl{40\symbol{32}000}}). Since
\xpl{petrol}, \xpl{mile}, \xpl{year} and \xpl{40} are all in the
vocabulary, these are all real-word errors. This of course raises the
question of what words to put in the vocabulary. Special characters
like \xpl{/}, \xpl{(}, \xpl{)} are troublesome because they do
not belong in the vocabulary and yet they carry meaning. Numbers
should definitely not be included in the vocabulary. These issues are further
discussed in Section~\ref{sec:ctr}.

\begin{table}[ht]
\centerline{
\begin{tabular}[H]{||ll|rr|rr|>{\bfseries}r>{\bfseries}r||}
\cline{3-8}
\multicolumn{2}{c}{}& \multicolumn{2}{|c}{Nonword error} & \multicolumn{2}{c}{Real-word
  error} & \multicolumn{2}{c||}{{\bfseries Total}}\\ \hline
& Single error                                        & 49 & 57\% & 1 & 17\% & 50 & 54\% \\
\raisebox{1.5ex}[0cm][0cm]{Missp.} & Multiple error   & 12 & 14\% & 0 & 0\%  & 12 & 13\% \\ \hline
& Single error                                        & 14 & 16\% & 0 & 0\%  & 14 & 15\% \\
\raisebox{1.5ex}[0cm][0cm]{Run-ons}      & Multiple error   & 3  & 3\%  & 0 & 0\%  & 3  & 3\% \\ \hline
& Single error                                        & 8  & 9\%  & 4 & 67\% & 12 & 13\% \\
\raisebox{1.5ex}[0cm][0cm]{Splits} & Multiple error   & 0  & 0\%  & 1 & 17\% & 1  & 1\% \\ \hline\hline
& Single error                                        & 71 & 83\% & 5 & 83\% & 76 & 83\% \\
Total  & Multiple error                               & 15 & 17\% & 1 & 17\% & 16 & 17\% \\
& {\bfseries Total}             & {\bfseries 86} & {\bfseries 100\%} & {\bfseries 6} & {\bfseries 100\%} & {\bfseries 92} & {\bfseries 100\%} \\
\hline
\end{tabular}
}
\caption{Breakdown of lexical errors in \textsc{cars}\label{tab:c-lexcloseup}}
\end{table}

Looking at Table~\ref{tab:c-lexcloseup}, the `easy' errors are the 49
misspellings that are nonwords and singletons. Although it is the
single largest class of errors, it only accounts for slightly more
than half of all the lexical errors. Error profiles taking
segmentation errors into account are hard to find and so there is not
much to compare with, but it seems that the 33\% (30/92) segmentation
error rate is high compared to what others have
found. Kukich~\shortcite{kukich:92a} reports that
15\% of the lexical errors in her TND corpus are segmentation errors.


\section{\sc travel}
\label{sec:corpus-travel}
The \textsc{travel} corpus contains 20 dialogues gathered from 20
subjects who were not aware of the role played by the
operator. \textsc{travel} consists of 717 utterances, 3,882 word tokens
and 941 word types. There are 424 (59\%) well-formed utterances, 184
(26\%) elliptic and 109 (15\%) ill-formed utterances. There is a
comparatively larger portion of ellipses in \textsc{travel} compared to
\textsc{cars}, particularly telegraphic ellipses. The explanation for
this probably lies in the different structures of the two domains and
the fact that the \textsc{travel} domain is supplied with maps. The
\textsc{travel} domain is more hierarchically structured than the
\textsc{cars} domain. There are islands in the archipelago, resorts on
the islands, hotels in the resorts and so on. The maps also naturally
focus the dialogue, and let the subject express herself in a
telegraphic manner, without the utterance being hard to interpret for
the operator. If the subject has a map of a village on Rhodes on the
screen with the hotels marked out, an utterance
like~(\ref{utt:prob-travel-ellipsis}) seems to come naturally.

\begin{list}{\texttt ==>}{\setlength{\rightmargin}{.5cm}}
        \inputitem{utterance}standard these hotels\label{utt:prob-travel-ellipsis}
\end{list}

The way that different modalities are exploited in a dialogue system
certainly has an effect on the dialogue itself and on the occurrences
of, for example, ellipses, but that is not our prime interest here.

\begin{table}[ht]
\centerline{
\begin{tabular}[H]{||lll|rr||}
\hline
& & & & \\
\multicolumn{3}{||c|}{\raisebox{1.5ex}[0cm][0cm]{Error Types}} &
\multicolumn{2}{c||}{\raisebox{1.5ex}[0cm][0cm]{\textsc{travel}}} \\ \hline
& \multicolumn{2}{l|}{Misspellings}              & 81  & 24\%  \\
& \multicolumn{2}{l|}{run-ons}                   & 21  & 6\%  \\
& \multicolumn{2}{l|}{splits}                    & 14  & 4\%  \\
\multicolumn{2}{||l}{Lexical errors $(\sum)$}   && 116 & 34\%  \\ \hline\hline
& \multicolumn{2}{l|}{Missing Constituent}       & 6   & 2\%  \\
& \multicolumn{2}{l|}{Extra Constituent}         & 0   & 0\%  \\
& \multicolumn{2}{l|}{Substituted Constituent}   & 12  & 4\%  \\
& \multicolumn{2}{l|}{Transposed Constituent}    & 0   & 0\%  \\
& \multicolumn{2}{l|}{Agreement Error}           & 9   & 3\%  \\
\multicolumn{2}{||l}{Syntactic errors $(\sum)$} && 27  & 8\%  \\
\hline\hline
& \multicolumn{2}{l|}{Telegraphic Ellipsis}      & 159 & 47\%  \\
& \multicolumn{2}{l|}{Contextual Ellipsis}       & 36  & 11\%  \\
\multicolumn{2}{||l}{Ellipses $(\sum)$}         && 195 & 58\%  \\ \hline\hline
\multicolumn{2}{||l}{{\bfseries Total}} && {\bfseries 338} & {\bfseries 100\%} \\ \hline
\end{tabular}
}
\caption{The distribution of errors and ellipses in \textsc{travel}\label{tab:t-errs}}
\end{table}

Apart from ellipses \textsc{cars} and \textsc{travel} are very
homogeneous. The distribution of the lexical errors for example among
the error types is almost exactly the same for the two corpora.

\begin{table}[ht]
\centerline{
\begin{tabular}[H]{||ll|rr|rr|>{\bfseries}r>{\bfseries}r||}
\cline{3-8}
\multicolumn{2}{c}{}& \multicolumn{2}{|c}{nonword error} & \multicolumn{2}{c}{real-word
  error} & \multicolumn{2}{c||}{{\bfseries total}}\\ \hline
                                     & Single error   & 64 & 62\% & 0  & 0\%  & 64 & 55\% \\
\raisebox{1.5ex}[0cm][0cm]{Missp.}   & Multiple error & 17 & 17\% & 0  & 0\%  & 17 & 15\% \\ \hline
                                     & Single error   & 16 & 16\% & 0  & 0\%  & 16 & 14\% \\
\raisebox{1.5ex}[0cm][0cm]{Run-ons}  & Multiple error & 5  & 5\%  & 0  & 0\%  & 5  & 4\% \\ \hline
                                     & Single error   & 1  & 1\%  & 11 & 85\% & 12 & 10\% \\
\raisebox{1.5ex}[0cm][0cm]{Splits}   & Multiple error & 0  & 0\%  & 2  & 15\% & 2  & 2\% \\ \hline
                                     & Single error   & 81 & 79\% & 11 & 85\% & 92 & 79\% \\
Total & Multiple error               & 22 & 21\% & 2  & 15\% & 24 & 21\% \\
& {\bfseries Total}          & {\bfseries 103}  & {\bfseries 100\%} & {\bfseries 13}  & {\bfseries 100\%} & {\bfseries 116}  & {\bfseries 100\%} \\
\hline
\end{tabular}
}
\caption{Breakdown of lexical errors in \textsc{travel}\label{tab:t-lexcloseup}}
\end{table}

The lexical error rate in \textsc{travel} (3\%) is only slightly
higher than that of \textsc{cars}. Table~\ref{tab:t-lexcloseup} shows
that again the nonword single error ratio $(64/103 = 62\%)$ is far
below the `agreed upon' 80\% reported by Damerau and
others. The segmentation error ratio is still high, 30\% is only
slightly down from the 32\% found in \textsc{cars}. The real-word
errors, of which there are only splits, are almost exclusively errors
due to lack of linguistic competence. \xpl{boat-trips} $\rightarrow$
\xpl{boat\symbol{32}trips} and \xpl{shark-attack} $\rightarrow$
\xpl{shark\symbol{32}attack} are examples of these cases. These errors
do not translate very well into English; there is, however, one error in
the corpus that more clearly demonstrates the nature of the error in
the original language:

\begin{list}{\texttt ==>}{\setlength{\rightmargin}{.5cm}}
        \inputitem{utterance}the water \hgl{drink able}\label{utt:prob-travel-split}
\end{list}

Besides the real-word error split,
utterance~(\ref{utt:prob-travel-split}) is also a telegraphic ellipsis.

\section{Conclusions}
\label{sec:corpus-concl}
The motivation for adding robustness functionality to an NL dialogue
system is to increase the number of utterances that the system can
accurately analyze, and for this purpose
\begin{itemize}
\item \emph{Lexical errors are more urgent than other errors}.
\end{itemize}
The results in Table~\ref{tab:c-module} show that this is true for
\textsc{cars}. Having to choose one of the robust modules, the system
would benefit the most from the lexical error recovery module. This is
also the case in the \textsc{travel} domain as far as lexical and
syntactic errors are concerned, although there are a greater number of
elliptic expressions compared to \textsc{cars}, especially telegraphic
ellipses. Lexical errors can, and should, be automatically corrected
in a dialogue system. As reagrds syntactic errors this is not as
obvious, and for ellipses this is probably not a very good idea at
all. Telegraphic ellipsis utterances, the largest ellipsis category,
contain all of the semantic information that is needed to interpret
them (cf. utterance~(\ref{utt:prob-synte2})). The fact that these
utterances are often syntactically erroneous should not be a major
consideration in trying to find the proper interpretation of the
utterance. It is probably better to circumvent the syntactic
constraints than attempting to exploit them when this sort of
phenomenon occurs. The same sort of reasoning can be applied to some
of the syntactic errors. The lexical errors, however, can not be
circumvented in any way. Out of the 91 lexical errors in \textsc{cars}
82 (90\%) are content words and 9 (10\%) are function words; the
numbers for \textsc{travel} are 103 (89\%) and 13 (11\%) out of a
total of 116 lexical errors. A large portion of the content words are
so-called \emph{domain words}, i.e. words with a strong domain
association. Fiftynine percent of the content words in \textsc{cars}
are also domain words. In \textsc{travel} the domain word ratio is as
high as 73\%. There is a high degree of unfamiliar words in
\textsc{travel}, (Greek) names of resorts and such, which can explain
the differences. It is difficult to produce a meaningful
interpretation of an utterance containing a lexical error without the
use of lexical error recovery methods, and if the error involves a
content word, or even worse a domain word, it is generally speaking
impossible.

The corpus investigation clearly shows that
\begin{itemize}
\item \emph{The lexical error situation here is more severe than
    normal}.
\end{itemize}
The dialogue scenario puts a higher cognitive load on the subject,
compared to many other applications. The subject is concerned with
extraction of information, not orthography. This, together with the
fact that the operator is forgiving with respect to errors, (the
subject notices that she can get away with lexical errors and becomes
less careful,) can explain the frequent errors which are also hard to
correct. In Kukich's~\shortcite{kukich:92a}
application, which is quite similar to the present one, the error rate
is as high as 5-6\%. The other applications cited in the previous
chapter all have lower error rates than the 3\% found in \textsc{cars}
and \textsc{travel}, with
Pollock and Zamora~\shortcite{pollock+zamora:83} reporting
the lowest error rate $0.2\%$. Taking the real-word errors into
account as well the error rate is close to 3\%. Apart from the high
error rate there is also a relatively small portion of `easy cases' in
the dialogue corpora compared to what others have found. The least
difficult lexical errors are the nonword single error misspellings,
which have been found by several researchers to make up 80\% of the
nonword errors. The figures here are 57\% (\textsc{cars}) and 62\%
(\textsc{travel}). The remainder of the lexical errors are
consequently distributed over the harder cases and it is evident that
\begin{itemize}
\item \emph{Wide error scope is crucial in a dialogue application}.
\end{itemize}
The consequence of not addressing the segmentation errors, for
example, will be that these are treated as misspellings, which is a
foolproof way of providing a lexical error recovery module with poor
performance. Researchers generally do not pay much attention to
segmentation errors, though
Kukich~\shortcite{kukich:92a} reports that run-ons
and splits make up 15\% of all the lexical errors in her corpus. The
ratio is considerably higher here, 32\% of the lexical errors in
\textsc{cars} being segmentation errors and the corresponding figure
for \textsc{travel} is 30\%. The hard cases, the segmentation errors,
the multiple errors and the real-word errors emphasize the fact that
\begin{itemize}
\item \emph{Contextual dependencies are crucial}.
\end{itemize}
There are generally more alternatives to be considered where
segmentation errors and multiple errors are concerned. Local
contextual preferences may then be used to distinguish the good
hypotheses from the bad ones. A lexical recovery module is needed that
can handle both misspellings and segmentation errors and has the
ability to deal with some of the problems in the smaller but harder
class of real-word errors.

Many automatic spelling error correction techniques have been
developed to address the nonword
misspellings. Kukich~\shortcite{kukich:92b} performed a comparative
study of some of the most well-known isolated word error correction
algorithms. The test set contained 170 human-generated misspellings,
of which 25\% were multiple error misspellings. The best algorithms
scored around 80\% correction accuracy. Applied to the \textsc{cars}
corpus (see Table~\ref{tab:c-lexcloseup},
Section~\ref{sec:corpus-cars}), this means that the best isolated word
error corrector can correct slightly more than 50\% of the lexical
errors in the corpus ($.8\times(49+12)/91 = .54$). This limit needs to
be raised.

\chapter{Background}
\label{ch:back}
The research area of automatic spelling error detection and correction
is nearly as old as the
computer. Kukich~\shortcite{kukich:92b} provides an
excellent review of the varied and wide-ranging area spanning the last
few decades.

The (incomplete) survey of the research field presented in this
chapter will review some of the more influential and/or interesting
contributions. Special interest is devoted to the error scope of the
proposed technique, whether or not contextual information is used and,
of course, the performance of the algorithms (to the extent that
authors quantify their results).

The review below is divided into three sections roughly corresponding
to three general `schools-of-thought' in automatic spelling
correction. Section~\ref{sec:back-classical} ``The Classical Method''
describes the string edit distance idea and techniques akin to
it. Section~\ref{sec:back-noisy} ``Noisy Channel Methods'' reviews the
probabilistic approach and Section~\ref{sec:back-nlp} ``Error
Correction in Rule-Based NLP Systems'' surveys some of the attempts
at robust natural language processing where spelling correction is
generally just one of several problem areas that are addressed.

\section{The Classical Method}
\label{sec:back-classical}
One of the earliest and probably the most influential contributions to
the area of human-generated spelling error correction techniques was
that of Damerau in 1964. Damerau~\shortcite{damerau:64} found that
approximately 80\% of all nonword misspellings were also single error
misspellings. Based on these findings Damerau subsequently implemented
what was later to be named the \emph{minimum edit distance}
algorithm. The algorithm detects errors by comparing the input word to
a dictionary. When an error is detected, the program tries to transform
the input word into a legal word in the dictionary relying on the
assumption that exactly one of the basic error operators has
`produced' the error. When a match is found, the process is halted
and the dictionary word is suggested as correction for the input word.

Independently
of~Damerau, Levenshtein~\shortcite{levenshtein:66}
developed a similar technique in the research discipline of error
correcting binary codes. The \emph{Levenshtein Distance} (LD) is the
distance between two words in terms of deletions, insertions and
reversals (transpositions). The idea of the LD metric algorithm is to
choose the word in the dictionary that is closest to the erroneous
input word. The LD metric algorithm is sometimes used synonymously to
the minimum edit distance algorithm, or rather, the term minimum edit
distance algorithm refers to both ideas nowadays.

Several authors have extended the LD metric
algorithm. Okuda~\emph{et al.}~\shortcite{okuda+al:76} introduced
the \emph{Weighted Levenshtein Distance} (WLD). This is a
generalization of the LD algorithm that can correct garbled words
containing multiple instances of character substitutions, insertions
and deletions. The weights can be used to give preference (shorter
distance) to one error type over the others. The authors were
satisfied with their algorithm, stating that: ``our method achieved
higher error correcting rates than any other method tried to
date''. Okuda~\emph{et al.}, however, noticed that short words
constitute a serious problem.

The character $n$-gram technique is another class of methods used for
spelling correction of various
kinds. Angell~\emph{et al.}~\shortcite{angell+al:83} implemented
a technique that computes a similarity measure between an input word
and the words in the dictionary based on trigrams. The similarity is
computed as the fraction of trigrams that are common to the two
strings.

The method achieved an overall accuracy of 76\% on a test set of 1,544
misspellings using a dictionary of 64,636 words. The authors noticed
that the correction rates for transpositions were very poor (36\%), it
was actually worse than for multiple errors (55\%). It was noticed
that more transpositions occurred in shorter words on the average and
short words were also problematic for the trigram similarity
technique.

The \prognm{speedcop} system~\cite{pollock+zamora:84} is one of the
techniques devised to correct single error misspellings by ways of
\emph{similarity keys}. Each word in the vocabulary is given a key and
when a misspelling is detected, its key is also computed and compared
to the set of precompiled keys corresponding to the words in the
dictionary. The words with identical or similar keys are considered as
correction alternatives. The similarity keys can be computed in a
number of ways, in this case the ordering of the consonants are
emphasized~\cite{pollock+zamora:84}. The authors tested their program
on over 50,000 misspellings gathered from seven different scientific
databases with a 40,000 word dictionary. \prognm{speedcop} scored an
overall correction rate of 74-88\% correction rate for the
different databases, counting only the misspellings whose
corresponding word was in the dictionary. \prognm{speedcop} uses some
complementary correction aids besides the similarity keys, (it
actually uses two sets of similarity keys). One such complementary aid
is the so-called ``function word routine'' which looks for
concatenations (run-ons) of frequent function words
(e.g. \xpl{oftheir}, \xpl{inthe}). It is not completely clear exactly
how the function word routine operates, but it is claimed that it
improves the overall correction rate by 1-2\%.

What the ``classical'' techniques above have in common is that they
look at words in isolation, which puts real-word errors outside the
scope of these techniques. Furthermore, based on the findings of
Damerau, most of the isolated word error correction
programs only consider singletons. Segmentation errors are not
addressed at all, with \prognm{speedcop}, which addresses a subset of
the run-ons, as the outstanding exception. The striking
observation is that segmentation errors are usually not even
mentioned, and this goes for the entire area, not just the classical
methods. It is often unclear whether or not errors due to defective
tokenization are included in the test sets.

There seems to be a limit of somewhere around 80\% correction rate for
the isolated word correction algorithms
(cf. Kukich~\shortcite{kukich:92b}). One of the major
problems with the classical methods is that the ranking of alternative
correction candidates is fairly imprecise, i.e. there will generally be
a number of correction alternatives that are equally `good' and there
is no way to decide which should be preferred. A more fine-grained
(although not necessarily better) measure of `goodness' is provided
via probabilities.

\section{Noisy Channel Methods}
\label{sec:back-noisy}
The noisy channel idea is based on the metaphor that communication is
relayed via an imperfect medium, the channel. A word is
inserted in one end of the channel and from the other end comes the
distorted version of that word. Given the distorted word and the
characteristics of the channel, the job is to calculate the most
likely word to have been inserted originally. This likelihood is
estimated via conditional probabilities. If there is also a language
model that has information on what words are likely to be inserted,
it provides what are known as the \emph{prior} probabilities. (See
Chapter~\ref{ch:rtr} for more details.)

The noisy channel model has a long tradition in the neighboring
research areas of optical character recognition (OCR) and speech
recognition. Especially in OCR the approach seems natural since the
channel (OCR-device) is there, open for inspection. In automatic
correction of human-generated spelling errors it is not as
straightforward since the channel characteristics are more
elusive. However, with the increasing availability of large corpora, the
noisy channel model has found its way into computational linguistics
in recent years.

Researchers who look to the noisy channel model for automatic
spelling error correction can roughly be divided into two groups:
those who emphasize the prior probabilities with the intention of
correcting real-word errors, and those who emphasize the channel's
distribution with nonword errors as prime target.

Kernighan, Church and Gale (KCG) were among the first to use really
large text-corpora to estimate the channel characteristics for the
purpose of spelling
correction. KCG~\shortcite{kernighan+al:90,church+gale:91a,gale+church:90}
constructed a program, \prognm{correct}, that generates and ranks
corrections for words rejected by \prognm{spell}\footnote{The Unix
  \prognm{spell} program~\cite{mcilroy:82} is a fast wide-covering
  spelling error \emph{detection} program that uses elaborate
  hash-functions to store the vocabulary for random access. See
  also~Domeij~\emph{et al.}~\shortcite{domeij+al:95}.}. The program
generates correction candidates for a typo by applying a single
instance of one of the basic error operators to the typo, and then it
ranks the candidates using a Bayesian scoring function. The spelling
corrector can thus handle single error nonword misspellings. The
scoring function is based on the modeling of the typing process as a
noisy channel. The most likely correction for a typo $t$ is the
correction candidate $c$ that maximizes $\prob{c} \prob{t|c}$. The
probabilities are estimated from the 1988 AP newswire corpus ($44\cdot
10^{6}$ words). The conditional probability, $\prob{t|c}$, is computed
from four confusion matrices that contain the number of deletions,
insertions, substitutions and transpositions that were recorded for
the individual characters in the words rejected by \prognm{spell}.

\prognm{correct}'s scoring function was tested on 329 misspellings
where the program generated two correction candidates. Three human
judges were asked to choose among the alternatives, given the rejected
word, the two alternatives and a few concordance lines of
context. \prognm{correct} agreed with the majority of the judges in
87\% of the cases.

During the work on the \prognm{correct} program it was noticed that
human judges were reluctant to decide between alternative candidate
corrections given only the information available to the program,
i.e. the typo and the candidate corrections. The judges were much
happier if they could see a line or two of the typo's surrounding
context. Consequently KCG went on to furnish \prognm{correct} with a
bigram model of local context. The Bayesian scoring function of a
correction candidate $c$ given a typo $t$ used in the original
\prognm{correct}, $\prob{c} \prob{t|c}$, the prior probability and the
channel probability, is complemented with the probability that the
word to the left of $c$ is $l$, $\prob{l|c}$, and the probability that
the word to the right is $r$, $\prob{r|c}$. The intention of KCG was
to see whether the updated scoring function: $\prob{c} \prob{t|c}
\prob{l|c} \prob{r|c}$ would enhance the performance of
\prognm{correct}. The test set of the 329 nonword errors rejected by
\prognm{spell} for which \prognm{correct} generated exactly two
correction candidates was again used for testing. Using only the prior
and channel probabilities \prognm{correct} agreed with the judges in
87\% of the cases. Using the bigram probabilities, performance rose to
almost 90\% which the authors found to be significant. They also
discovered that the bigram parameter estimator was crucial to the
behavior of the program. Maximum likelihood estimation and expected
likelihood estimation actually degraded performance or made no
difference, respectively, to the original \prognm{correct}. The only
estimation technique to improve performance was the Good-Turing
estimation technique \cite{good:53,church+gale:91b}.

Kashyap and Oommen~\shortcite{kashyap+oommen:81,kashyap+oommen:84}
had investigated the same approach some years earlier although they had
used subjective estimates of the parameters describing the
channel. Their results range from 30\% to 92\% correction accuracy
depending on the word length and number of errors per word. They also
report that the figures compare favorably with those reported
by~Okuda~\emph{et al.}~\shortcite{okuda+al:76} which ranged from
28\% to 64\% for words with similar characteristics.

Mays~\emph{et al.}~\shortcite{mays+al:91} took the other route,
the one that focuses on the correction of real-word errors. The prior
distribution is modeled with a trigram language model:

\[\prob{W} = \prod_{i=1}^{n} \prob{w_{i}|w_{i-2},w_{i-1}}\]

The trigram language model has relatively good predictive power but
the channel-model used by Mays~\emph{et al.} is quite
simple. Each word $w_{i}$ in the vocabulary has a precompiled
\emph{confusion set} $w_{i}^{c}$. The confusion set is generated by
applying the four basic error operators to each character position of
the word exactly once and adding the resulting string to the confusion
set if it is a legal word in the vocabulary. $w_{i}$ is also added to
$w_{i}^{c}$. The confusion set of a word thus contains all the single
error misspellings that are legal words in the
vocabulary. $\prob{y_{i}|w_{i}}$, where $y_{i}$ is the real-word
error, is then simply computed as:
\[
\prob{y_{i}|w_{i}} = 
\left\{
  \begin{array}{ll}
    \alpha & \mbox{if}\ y_{i}=w_{i}\\
    (1 - \alpha)/(|w_{i}^{c}| - 1) & \mbox{otherwise}
  \end{array}
  \right.
\]
where $\alpha$ is a constant determined by experimentation ($\alpha =
.9, .99, .999, .9999$ in the reported experiments). Note that all
correction candidates for $y_{i}$ in $w_{i}^{c}$ have equal
probabilities, except for $w_{i}$. The technique was tested on 8,628
sentences using a 20,000 word vocabulary. The test sentences were
generated from 50 sentences of AP newswire and 50 sentences from the
Canadian Parliament transcripts. These 100 original sentences were
manipulated so that each of the 8,628 test sentences contained exactly
one single error real-word error. The technique proposed by
Mays~\emph{et al.} detected and corrected 73\% of the 8,628 single
error real-word errors. It is safe to say that it is the trigram model
that does the work, the channel model merely enumerates the candidates
where the trigram scores
low. Mays~\emph{et al.}~\shortcite{mays+al:91} used the trigram
language model employed in the IBM speech recognition
project~\cite{bahl+al:83}.

Golding and Schabes~\shortcite{golding+schabes:96,golding:95}
took the idea of Mays~\emph{et al.} in a slightly different
direction. They also used confusion sets but these were not based on the
basic error operations. Rather, the confusion sets were selected on
the basis of the fact that they occurred frequently in the Brown
Corpus~\cite{kucera+francis:67} and sampled from the list of ``Words
Commonly Confused'' in Random House~\cite{flexner:83}. The confusions
represent different types of errors: homophone confusions
\{\xpl{peace},\xpl{piece}\}, grammatical confusions
\{\xpl{among},\xpl{between}\} and the authors also added some
typographical confusions that were not found in Random House,
e.g. \{\xpl{being},\xpl{begin}\}. Golding and Schabes
contrasted three different language models: a part-of-speech (POS)
trigram language model, a feature-based Bayesian scoring function and
a hybrid of the two. The POS trigram worked better than the Bayesian
scoring function and the hybrid outperformed both. The technique was
tested on 20\% of the Brown Corpus. The evaluation is rather
small-scale in that only 18 confusion sets are used. The confusion
sets are also quite small, they contain only two or three words
each. Two tests were conducted, one to see whether the system would
wrongfully change the correct usage of a word in a confusion set
into the incorrect usage, and one to see if the system could restore
the correct usage of the word if presented with an error. The
performance varies for the 18 different confusion sets, results ranging
from 35\@.3\% to 98\@.4\% on the corrupted words, and from 87\@.8\% to
100\% on the uncorrupted words.

Atwell and Elliott~\shortcite{atwell+elliott:87} took
\prognm{claws}~\cite{garside:87} as a starting point to detect
real-word errors. \prognm{claws} is a program that assigns POS tags to
a text using a POS bigram language model. The idea is simply to assign
POS to the text of interest using \prognm{claws} and when the
probability of two consecutive tag-pairs fall below a predefined
threshold, the word whose tag is present in both tag-pairs is marked as
an error. The authors collected a 13,500 word corpus and extracted 502
real-word errors. The real-word error detector scored a 62\% recall
and 35\% precision. The way that the threshold is set is obviously
crucial to the performance of the method. In an attempt to better the
poor precision rating Atwell and Elliott raised the
threshold slightly and precision rose from 35\% to 38\%, but in doing
so recall dropped from 62\% to 47\%. These results indicate that the
discriminative powers of the POS bigram probabilities alone are too
weak. This is particularly evident when an absolute value (threshold)
is used for making the decisions. Comparing alternative tag-pair
sequences to each other is surely a more promising approach.

There is obviously a greater interest in the more challenging
real-word errors among the researchers adopting the noisy channel
approach. The focus is consequently on the language model, often at
the expense of the channel characteristics. The confusion set idea
works satisfactorily for a small set of precompiled errors introduced
in the channel, but is left without a chance to correct an error that
does not belong to one of the confusion sets. One can, of course,
argue that one should have a complementary program that performs
nonword error correction and that the real-word error correction
algorithm should be applied on the output from this program. The
problem, however, is that if the real-word error correction module
proposes corrections based solely on the preferences of the language
model, regardless of orthographic similarity between error and
correction, it will introduce errors that were not originally there.

The introduction of contextual properties in the spelling correction
algorithms is certainly beneficial. Contextual properties are useful
not only for the purpose of real-word error correction. KCG use the
contextual information to be able to better discriminate between
competing correction alternatives to nonword errors. Although there is
nothing in principal that prevents one from using contextual
dependencies also in the ``classical approaches'', it seems as though
they are more naturally incorporated into the noisy channel model.

None of the authors cited above mentions segmentation errors.

\section{Error Correction in Rule-Based NLP Systems}
\label{sec:back-nlp}
This section takes a wider view on errors in natural language
text. The aim of an NLP-system is to analyze, interpret, translate or
whatever the application might be, as many input sentences as
possible, and for this reason errors other than just spelling errors
are, of course, of interest.

The need for error-handling capabilities became evident as real
users were let loose on the early prototype NLP-systems. As a
consequence the early
eighties saw a range of NLP-systems with robustness functionality
built into them. These systems can be divided into three general
categories~\cite{kukich:92b}: the relaxation-based
(e.g.~\cite{heidorn+al:82,weischedel+sondheimer:83}),
expectation-based (e.g.~\cite{carbonell+hayes:83,granger:83}) and the
acceptance-based techniques (e.g.~\cite{fass+wilks:83}).

The acceptance-based approach works under the assumption that errors
can be ignored as long as an interpretation can be found that is
meaningful in the given application. The approach is grounded in the
observation (hypothesis) that this is the way humans deal with erroneous
or disturbed input. Acceptance-based approaches tend to make extensive
use of semantic information and little use of any other level of
linguistic information.

The expectation-based technique derives its expectations from various
linguistic knowledge
sources. \prognm{caspar}/\prognm{dypar}~\cite{carbonell+hayes:83} use
syntactic and semantic expectations expressed in a case frame whose
slots expect to be
filled. Carberry~\shortcite{carberry:84} generates
responses to pragmatically ill-formed dialogue input based on the
user's goals and plans that have been inferred from the preceding
dialogue.

Contrary to the acceptance-based approach the relaxation-based
technique assumes that no errors can be ignored. This assertion is
largely based on the fact that many early NLP-systems depended heavily
on syntactic rules, and when the input violated the rules, the parsing
process simply came to a halt. The relaxation-based technique will try
to find the rule that when relaxed will allow the parser to
succeed. If this scheme works, it means that the error is both
localized and diagnosed and can consequently also be corrected. The
relaxation-based approach is the one that has been favored in the
research community, compared to the other two approaches. However,
there are still problems. When only syntactic constraints are applied, there
are many rules that when relaxed will lead to a complete parse and
there will be a large amount of correction
alternatives. Mellish~\shortcite{mellish:89} noticed
this problem trying to correct unknown/misspelled words, omitted words
and spurious words with relaxation techniques using a
CFG. Ingels~\shortcite{ingels:92,ingels:93} tried the
same approach, but with a richer grammar formalism that allowed also
for feature values to be relaxed, and experienced the same
problem.

After the hectic years in the early eighties interest in robust NLP
seemed to have faded somewhat, but only to rise again in the late
eighties and early nineties. Two important systems from this period
are \prognm{critique}~\cite{richardson+braden-harder:88} and a
text-editing system for Dutch~\cite{kempen+vosse:90,vosse:92}. Both
systems are relaxation-based but only the latter takes spelling errors
seriously.

The \prognm{critique} system~\cite{richardson+braden-harder:88}, with
its ancestor the IBM \prognm{epistle} system~\cite{heidorn+al:82}, is
one of the few systems to sincerely address robustness issues on a
large scale, i.e. wide covering, natural language text processing
system. The dictionary includes more than 100,000 entries and the
words also carry information used in syntactic processing. The grammar
contains several hundred phrase structure rules. The \prognm{critique}
system accepts text as input and delivers critique on grammar and
style at levels of detail that can be adjusted by the user. The
original \prognm{epistle} system could cope with different types of
grammatical errors whereas the rule-based parser in \prognm{critique}
``\ldots provides a unique approximate syntactic parse for a large
percentage of English text and diagnoses over 100 grammar and style
errors''. After an initial preprocessing phase the robust parsing
proceeds along the following lines: Lexical processing identifies
words not in the dictionary and assigns them default morphological and
syntactic information to avoid parsing failure. After the lexical
analysis the text is passed to the parser. If the parser fails to
produce a complete parse, the parser goes over the text once more,
this time with some of its constraints relaxed. Certain lexical
substitution rules are also activated during this second pass. The
substitution rules involve easily confused words such as
whose$\leftrightarrow$who's or its$\leftrightarrow$it's. If parsing
now succeeds, the relaxed rules will serve as a basis for the critique
fed back to the user. If this still does not succeed in a complete
parse after the relaxation phase, the `parse fitting' procedure is
invoked~\cite{jensen+al:83}. The parse-fitting procedure relies on
heuristics to choose a head constituent to which fragments produced by
the parser in the relaxation phase can be attached to form an
approximate parse. Even when parse-fitting is applied, grammar and
style error critique can be produced for the incomplete fragments. In
any of the processing phases described multiple parses may result and
in this case the system selects one based on a parse metric which
favors trees in which modifying words and phrases are attached to the
closest qualifying constituent~\cite{heidorn:82}.

The accuracy of \prognm{critique} was tested on 10 essays from each of
four groups: freshman compositions, business writing, ESL (English as
Second Language) and professional writing. The diagnoses made by
\prognm{critique} were classified as correct, useful or wrong, useful
meaning that the detection of the error was satisfactory but that the
critique was off the target. The authors did not consider errors that
the system missed, strangely enough. \prognm{critique} produced the
correct advice in 39\% (professional), 54\% (ESL), 72\% (freshman) and
73\% (business) of the critiques. When useful critiques were taken
into account, figures rose in all four categories, particularly for the
ESL-group (54\%$\rightarrow$87\%), but as the authors point out,
useful critique may not be that useful to users who lack native
intuitions about English.

Another parser-based text proof-reading system is that developed for
Dutch
by~Kempen and Vosse~\cite{kempen+vosse:90,vosse:92}. The
system can correct nonword spelling errors, real-word syntactic errors
such as agreement errors, it can also handle word doubling errors,
problems with idiomatic expressions and compounds. Some structural
errors such as punctuation errors and strange word order errors can
also be dealt with. The robustness is achieved mainly in two
processing phases, the word-level processing and the sentence-level
processing. In the word-level processing the linear order of the text
is abandoned for a lattice structure. The lattice reflects ambiguities
that arise from compounds, idiomatic phrases and word doublings. If
the text contains spelling errors a correction module is invoked and a
limited number of correction alternatives are added to the
lattice. The correction algorithm is based on a variation of trigram
analysis~\cite{angell+al:83} and triphone
analysis~\cite{vanberkel+desmedt:88} extended with a ranking
mechanism. The dictionary contains 250,000 word-forms. After the word
level processing the sentence level processing can proceed (on the
lattice). The parser is a shift-reduce parser working with an
Augmented Context-Free Grammar (ACFG) of some 500 rules. If an
agreement constraint is violated during parsing, the constraint is
relaxed and appropriately marked as such, and parsing
continues. Structural errors are dealt with in a different
manner. These (unusual according to the authors) errors are parsed
with error rules included in the grammar. When parsing is finished,
the job is to select among the (sometimes very frequent) alternative
parses. The most straightforward method is to simply count the number
of errors in the parses and choose the one with the least number of
errors in it. However, this instrument is too blunt since there may be
many parses with the same number of errors. To make the ranking of
alternatives more fine-grained the grammar rules have weights added to
them. For example, verb transitivity violation is more heavily
penalized than incorrect subject verb agreement. After the selection
phase the text can be regenerated with suggested corrections and
diagnostics messages.

The word level processor was tested on 1,000 lines of text randomly
chosen from two large texts submitted for publication, one on
employment legislation and the other concerning collective wage
legislation~\cite{vosse:92}. The sample contained almost 6,000 words
with 30 nonword misspellings. Twentyeight of the misspellings were
detected and 14 were given the proper correction. The two missed
misspellings were assumed by the system to be proper names. Eighteen
false alarms were produced. Compared to an elementary spell-checker
(supposedly the sort of spell-checker that comes with word processors)
the word-level processor performed well. A simple spell checker, on
the same text, marked 217 words as misspelled, which amount to 187
false alarms, 37 abbreviations and proper names and 150 compounds. The
authors report that the word level can process in excess of 25 words
per second. The sentence-level, however, requires considerably more
time. Processing time ranges from four or more words per second for
short error-free sentences to several seconds per word for longer and
more error-prone sentences. On a 150 sentence spelling test for
secretaries and typists, the system was able to correct 72 out of 75
errors without any false alarms. The errors corrected were spelling
errors, agreement errors and errors in idiomatic expressions. The
three errors missed involved semantic violations. The 150 sentences
were processed in under nine minutes. Although the spelling correction
algorithm
of~van~Berkel and de~Smedt~\shortcite{vanberkel+desmedt:88}
works in an environment rich in linguistic information, it is
actually an isolated word correction technique. That is, the
correction alternatives are generated without contextual
information. The contextual information is supplied afterwards to
disambiguate among the multiple alternatives.

The problem of disambiguation has a long history in NLP research. The
problem is to find the correct word-sense for a word in a particular
context where the word has more than one interpretation in the
vocabulary. This problem is very hard and requires large amounts
of linguistic knowledge on all levels, and in the general case also
extra-linguistic knowledge of the world.

The problem of lexical error recovery can also be seen as a problem of
disambiguation, but on a different level, the string level. There is
certainly no conflict of interests here, both problems are important
and need to be solved. The question is rather: what are the crucial
knowledge sources in the respective problems?

There is obviously a fair amount of randomness involved in how lexical
errors are produced. Probabilistic techniques are well suited
(although not always exploited) to capture the randomness of spelling
behavior, which is one of the fundamental ideas behind the noisy
channel approach. Still, there are clearly patterns there as well.
van~Berkel and de~Smedt~\shortcite{vanberkel+desmedt:88}
focus on phonetic resemblance which is one of several
distinguishable patterns. Emphasizing one type of pattern is bound to
obscure others. Techniques based on probabilistic or statistical
methods usually derive their model's parameters from a corpus, and in
the case of lexical error recovery, the corpus would consist of
errors. This implies that as long as we have a large enough set of
example errors, preferably produced by real users, we can train or
derive a model that can describe any pattern that might be in the
corpus. The model would then `encode' a randomness/pattern mix that
reflects the contents of the corpus. The problem is of course that a
large enough corpus is quite hard to come by. Probabilistic techniques
also provide a sharp disambiguation instrument, there will always be
one alternative that is better (more likely) than the rest (for better
or for worse). The voluminous NLP-systems reviewed in this section
contain a large set of hand-crafted rules. The proof-reading system
of~Vosse~\shortcite{vosse:92} uses this rule-base (of
primarily syntactic constraints) to \emph{disambiguate} among the
multiple words and word-senses produced by the isolated word spelling
correction program. That is, it uses information that really pertains
to a different level of description. There is nothing wrong with that,
it obviously produces useful results. The point, however, is that
information pertaining to the string disambiguation problem is
overlooked. Another problem related to the use of large rule-bases for
lexical error disambiguation is that it is hard to move to another
domain or language. The isolated word spelling correction technique is
not very useful without the syntactic constraints, so a new set of
rules must be manually constructed before the spelling correction
technique can perform satisfactorily. This is a laborious task.

Although algorithms dealing with isolated word error detection and/or
correction work on words in isolation, the program usually processes
chunks of running text, depending on tokenization before the detection
phase. The tokenization process is often simplified so that errors
involving word boundaries will be more or less impossible to
correct.

Carter~\shortcite{carter:92} implemented an elaborate tokenization and
error-correcting algorithm in the \prognm{clare} system which explicitly
considers segmentation errors. Carter maintains a lattice of
overlapping word hypotheses and uses syntactic and semantic
constraints to select the best alternative. \prognm{clare} was tested on
102 artificially generated sentences containing 108 errors without the
use of syntactic and semantic constraints. The system found a single
correct repair in 59 cases and 24 of these involved segmentation
errors. It is not clear what the ratio of segmentation errors was
in the source text.

In spite of the work of Carter and others,
Kukich~\shortcite{kukich:92b} states that ``the general problem of
handling errors due to word boundary infractions remains one of the
significant unsolved problems in spelling correction research''
(p. 385). This is one of the main problems we deal with in the
following chapters.

\chapter{An Algorithm for Robust Text Recognition}
\label{ch:rtr}

This chapter includes the theoretical contributions of this
thesis. The contributions are principally condensed into two
algorithms, one for isolated word error correction
(Section~\ref{sec:iwr} ``Isolated Word Recognition''), and one for the
correction of lexical errors in general in running text
(Section~\ref{sec:ctr} ``Connected Text
Recognition''). Sections~\ref{sec:fhmm} ``Fundamentals of Hidden
Markov Models'' and~\ref{sec:tp} ``Token Passing'' introduce the
building blocks and tools with which these algorithms are built.


\section{Fundamentals of Hidden Markov Models}
\label{sec:fhmm}
This section gives a brief account of the theory behind the
\emph{discrete observation Hidden Markov Model} (HMM). We present the
components of the model and the basic algorithms that can be applied
to it. These can of course be found in several other
places~\cite{rabiner:89,levinson+al:83}, but the account below is
slightly different from the `standard model'. In the theoretical
underpinnings of Markov models there is no notion of \emph{final
states} as in the case of the FSA, for example. In real life, however,
observation sequences are finite, at least the ones we are interested
in. One way of thinking of final states of a Hidden Markov Model would
be to somehow decide that a subset of the states are legal final
states and that those of the complementary set are not. This approach
can be found in, for
example,~Deller~\emph{et al.}~\shortcite{deller+al:93}(p. 690). Another
way would be to assign a probability to each state stating how likely
it is that the particular state is a final state. This approach
certainly blends better with the general theory, and more importantly
it is trainable as will be shown below.

The hidden Markov model presented here has two additional states
compared to the standard model. These are called the \emph{entry
state} and the \emph{exit state}. (The terms are borrowed
from~Young~\emph{et al.}~\shortcite{young+al:89}. The terms entry
and exit are preferred over start and final for reasons that will be
become clear in Section~\ref{sec:ctr}.) As mentioned before, the
notion of final states in Markov models is not new and the difference
between the standard model and the one presented here is quite small,
although not insignificant. An account of this type of Markov model
and the fundamental algorithms associated with it has to my knowledge
not been published before.

A random process $\underline{X}$ is a sequence of random variables
\[\underline{X}=\{\ldots X_{t-1},X_{t},X_{t+1}\ldots\}\]
If the value of $X_{t}$ is dependent on the value of $X_{t-1}$ but
independent of earlier variables, i.e.
\[P(X_{t}=q_{t}|X_{t-1}=q_{t-1},X_{t-2}=q_{t-2},\ldots)=P(X_{t}=q_{t}|X_{t-1}=q_{t-1})\]
we say that the process is a (first order) Markov process, and when the
variables take discrete values, a Markov Chain. Given that the range
space of the variables is finite the Markov Chain can be modeled by
a finite state network where the states are associated with the
outcomes of the random variables. Arcs connecting the states of the
network impose transition probabilities between the states. The
transition probabilities are often denoted
\[
a_{ij}=\prob{X_{t}=q_{j}|X_{t-1}=q_{i}}
\]
If the transition probability in a
state is independent of time $t$, the Markov Chain is said to be
homogeneous.

The \emph{Hidden Markov Model} (HMM) models two parallel homogeneous
random processes where one is the state transition sequence just
described and the other is a sequence of observation symbols
\[\underline{Y}=\{\ldots Y_{t-1},Y_{t},Y_{t+1}\ldots\}\]
The variables in \(\underline{Y}\) take values from a discrete set $V$
of observations or observables and the observation symbol
probabilities are denoted
\[b_{j}(k)=P(Y_{t}=v_{k}|X_{t}=q_{j})\]
where \(v_{k}\in V\). The model is thus extended with an observation
symbol distribution for each state. The HMM $\mathcal{M}$ is thus
determined by:
\begin{itemize}
        \item a finite set of states \(Q = \{q_{1},q_{2}, \ldots
          ,q_{N}\}\), where $q_{1}$ is the non-emitting entry state
          and $q_{N}$ is the absorbing non-emitting exit state
        \item a finite set of observables \(V = \{v_{1},v_{2},\ldots ,v_{K}\}\)
        \item an $(N-1)\times (N-1)$ transition matrix $\mathbf{A}$ where \(a_{ij}\) denotes\\
          \(P(X_{t}=q_{j}|X_{t-1}=q_{i})\)
        \item an $(N-2)\times K$ observation matrix $\mathbf{B}$ where \(b_{j}(k)\) denotes\\
          \(P(Y_{t}=v_{k}|X_{t}=q_{j})\)
\end{itemize}
There are no transitions out of $q_{N}$ and no transitions into
$q_{1}$, thus the dimensions of the transition matrix
$\mathbf{A}$. The states $q_{1}$ and $q_{N}$ do not emit any symbols
and this explains the dimensions of the observation matrix
$\mathbf{B}$. The parameters $N$ and $K$ along with a
specification of the observation symbol alphabet/vocabulary and the
two distributions $\mathbf{A}$ and $\mathbf{B}$ determine the HMM
$\mathcal{M}$. The shorthand for this is
$\mathcal{M}=\langle\mathbf{A},\mathbf{B}\rangle$.

The HMM above can be used as a generator of an observation sequence
\[O = o_{1},o_{2},\ldots ,o_{T}\]
where $o_{t}, 1\leq t\leq T$ is chosen from $V$. The workings of this abstract
machine in generative mode is as follows:
\begin{enumerate}
      \item Start in the designated entry state $q_{1}$ at
        time\footnote{In the speech application the notion of time refers
          to actual time, it has to do with sample rates and
          suchlike. In the text case, time is merely a metaphor. The
          discrete time points should be thought of as an ordering of
          events.} $t=0$.
      \item Transit to a new state, say $q_{j}$ according to the state transition
        distribution in the current state $q_{i}$ ($a_{ij}$).\label{item:loop-start}
      \item Set $t=t+1$.
      \item Choose $o_{t}=v_{k}$ according to the observation symbol
        distribution in state $q_{j}$ ($b_{j}(k)$).
      \item If $t=T$ terminate by taking the transition to the exit state $q_{N}$
        according to $a_{jN}$ and set $t=T+1$. if $t<T$ go
        to~\ref{item:loop-start}.
\end{enumerate}
Setting $t=T+1$ might seem strange when the observation sequence is
only of length $T$. It is, however, theoretically unpleasant that
$X_{T}$ can assume two different values, so in the algorithmic
descriptions below $X_{T+1}$ will occasionally appear and its value
will always be $q_{N}$. In the implementation of the algorithms this
fix is not necessary.

Given an HMM $\mathcal{M}=\langle\mathbf{A},\mathbf{B}\rangle$ and an
observation sequence \(O = o_{1},o_{2},\ldots\,o_{T}\) three important
tasks can be performed.
\begin{description}
        \item[{\mdseries\textsf{Task 1.}}] Calculate the probability of the observation sequence
          $O$ given the model $\mathcal{M}$,
          i.e. \(P(O|\mathcal{M})\). This can be done with the
          \emph{Forward-Backward procedure}.\label{it:fhmm-task1}
        \item[{\mdseries\textsf{Task 2.}}] Calculate the joint probability of the observation
          sequence $O$ and the optimal state sequence $Q^{*}$ given
          the model $\mathcal{M}$,
          i.e. $\prob{O,Q^{*}|\mathcal{M}}$. The \emph{Viterbi
          algorithm} computes this probability.\label{it:fhmm-task2}
        \item[{\mdseries\textsf{Task 3.}}] Reestimate the model parameters $\mathbf{A}$ and
          $\mathbf{B}$ so as to maximize the probability of a given
          observation sequence, the training material. The training
          procedure is called the \emph{Baum-Welch reestimation
            algorithm}.\label{it:fhmm-task3}
\end{description}
To achieve Task~1 in an efficient way we need the
\emph{forward variables} and the \emph{backward variables}. These
variables are used to store intermediate results of the
forward--backward algorithm. Actually it would suffice with the
forward variables \emph{or} the backward variables to calculate
$\prob{O|\mathcal{M}}$, but since we need both sets later in the
training procedure, both sets are defined here. The forward variables
$\alpha_{t}(i)$ hold the probability of being in state $q_{i}$ at time
$t$ having observed the partial sequence $o_{1},\ldots ,o_{t}$. That
is
\[\alpha_{t}(i) = \prob{o_{1},\ldots ,o_{t}, X_{t}=q_{i}|\mathcal{M}}\]
The forward variables are recursively defined:
\begin{description}
\item[Initialization]
\begin{eqnarray}
1\leq i\leq N & & \nonumber \\
\alpha_{0}(i) & = &
\left\{ \begin{array}{rl}
    1 & \mbox{if $i=1$} \\
    0 & \mbox{otherwise}
  \end{array}
\right.
\end{eqnarray}
\item[Induction]
\begin{eqnarray}
0\leq t\leq T-1 & , & 2\leq j\leq N-1 \nonumber \\
\alpha_{t+1}(j) & = & \left[ \sum_{i=1}^{N-1} \alpha_{t}(i) a_{ij} \right]
b_{j}(o_{t+1})
\label{eq:fhmm-alphainduct}
\end{eqnarray}
\item[Termination]
\begin{eqnarray}
\prob{O|\mathcal{M}} & = & \sum_{i=1}^{N-1} \alpha_{T}(i) a_{iN}
\end{eqnarray}
\end{description}
The purpose behind the entry and exit state should now start to become
clearer. The vector $\mathbf{a_{1}}$, the transitions out of the entry
state, is the equivalent of the initial state distribution usually
denoted by the vector $\boldsymbol{\pi}$ in the standard model. The
number $a_{1i}$ for example, gives the probability that the
observation sequence starts in state $q_{i}$. The entry state
modification is there for practical purposes only, which will be
explained below.

The exit state has also a practical purpose, but it also conveys the
final state idea. The number $a_{iN}$, for example, gives the
probability that the observation sequence ends in $q_{i}$. It should
be noted that the induction step does not include $q_{N}$ in any
way. The transition to the exit state is taken when the entire
observation sequence has been observed. One can think of this as
having some sort of end-of-sequence marker ($o_{T+1}$) after the last
symbol that triggers the final transition to the exit
state. $\alpha_{T+1}(N)$ would then equal the summation in the
termination step and $\alpha_{T+1}(i)=0$ for all $i\neq N$. There is,
however, no point in adding the $T+1$ column to the
$\alpha$-matrix. (The $\alpha$ variables are typically kept in a
matrix, as is the case here.)

The backward variables $\beta_{t}(i)$ hold the probability of making
the partial observation $o_{t+1},\ldots ,o_{T}$ and then taking the
transition to the exit state given state $q_{i}$ at time $t$. That is
\[\beta_{t}(i) = \prob{o_{t+1},\ldots ,o_{T}, X_{T+1}=q_{N} |
  X_{t}=q_{i}, \mathcal{M}} \] Again the backward variables are
recursively defined:
\begin{description}
\item[Initialization]
\begin{eqnarray}
1\leq i\leq N-1 & & \nonumber \\
\beta_{T}(i) & = & a_{iN}
\end{eqnarray}
\item[Induction]
\begin{eqnarray}
T-1\geq t\geq 0 & , & 1\leq i \leq N-1 \nonumber \\
\beta_{t}(i) & = & \sum_{j=2}^{N-1} a_{ij} b_{j}(o_{t+1})
\beta_{t+1}(j)
\end{eqnarray}
\end{description}
One can think of the backward variables as being recursively computed
going from last to first in the observation sequence. Note that
$\beta_{T}(i)$ means the probability of being in the exit state next,
given that the present state is $q_{i}$,
i.e. $\beta_{T}(i)=\prob{X_{T+1}=q_{N}|X_{T}=q_{i}} = a_{iN}$. Note
also that $\beta_{0}(1)=\prob{O|\mathcal{M}}$.

The word \emph{Hidden} in Hidden Markov Model comes from the fact
that for any given sequence of observation symbols there can be many
different underlying state sequences and it is impossible to say which
one generated the particular observation at hand. The state sequence is
hidden. There must, however, be one state sequence that is at least as likely
to have produced the observation as any other. That is, given the
observation sequence $O=o_{1},\ldots ,o_{T}$ there is a
sequence $Q^{*} = q_{1}^{*},\ldots ,q_{T}^{*}$ for which
\begin{eqnarray*}
\prob{O,Q^{*}|\mathcal{M}} & \geq & \prob{O,Q|\mathcal{M}} \hspace{2em}
  \forall Q\neq Q^{*}
\end{eqnarray*}
That is, Task~2 above. Enumerating all the $N^{T}$ possible
state sequences and choosing the best is obviously not a realistic
option. Fortunately there is an efficient solution to the problem, the
Viterbi algorithm, another dynamic programming algorithm. To keep
track of the partially optimal sequence on route to
$\prob{O,Q^{*}|\mathcal{M}}$ we define the set of variables
\begin{equation}
\phi_{t}(j) = \max_{q_{1},\ldots ,q_{t-1}} \left[ \prob{o_{1},\ldots
    ,o_{t}, q_{1},\ldots ,q_{t-1}, X_{t}=q_{j} | \mathcal{M}} \right]
\label{eq:fhmm-phidef}
\end{equation}
i.e. $\phi_{t}(j)$ is the probability of the most likely state sequence
that ends in $q_{j}$ which also accounts for the first $t$
observations. In order to retrieve the actual state sequence a second
set of variables $\psi_{t}(j)$ are needed. The $\psi_{t}(j)$ keep
track of the optimal predecessor of state $q_{j}$ in the path corresponding to
$\phi_{t}(j)$. The Viterbi algorithm proceeds as follows:

\begin{description}
\item[Initialization]
\begin{eqnarray}
1\leq i\leq N & & \nonumber \\
\phi_{0}(i) & = &
\left\{ \begin{array}{rl}
    1 & \mbox{if $i=1$} \\
    0 & \text{otherwise}
  \end{array}
\right. \\
\psi_{0}(i) & = & 0
\end{eqnarray}
\item[Induction]
\begin{eqnarray}
 1 \leq t\leq T & , & 2\leq j\leq N-1 \nonumber \\ 
 \phi_{t}(j) & = & \max_{1\leq i\leq N-1}\left[
   \phi_{t-1}(i)a_{ij} \right] b_{j}(o_{t}) \label{eq:fhmm-phiinduct}\\ 
 \psi_{t}(j) & = & \argmax{1\!\leq \!i\!\leq \!N\!-\!1}{\phi_{t-1}(i)a_{ij}}
\end{eqnarray}
\item[Termination]
\begin{eqnarray}
  \prob{O,Q^{*}|\mathcal{M}} & = & \max_{1\leq i\leq
    N-1} \left[ \phi_{T}(i)a_{iN} \right] \\
  q_{T}^{*} & = & \argmax{1\!\leq \!i\!\leq \!N\!-\!1}{\phi_{T}(i)a_{iN}} \\
  (q_{T+1}^{*} & = & q_{N}) \nonumber 
\end{eqnarray}
\item[Path Backtracking]
\begin{eqnarray}
  t = T-1,\ldots ,0 & & \nonumber \\
  q^{*}_{t} & = & \psi_{t+1}(q^{*}_{t+1},)
\end{eqnarray}
\end{description}
This algorithm is quite similar to the Forward Backward algorithm. The
only real difference is that the summation in the computation of the
forward variables is substituted for a maximization in the Viterbi
algorithm.

Recall Task~3: Given a model $\mathcal{M} =
\langle\mathbf{A},\mathbf{B}\rangle$ and an observation sequence $O$,
adjust the parameters of $\mathcal{M}$ so that $\prob{O|\mathcal{M}}$
is maximized. This task is not as simple as the two previous ones. In
fact, there is no way to optimally estimate the model parameters. An
iterative procedure such as the Baum-Welch reestimation algorithm can,
however, be used to find a model that locally maximizes
$\prob{O|\mathcal{M}}$. In each iteration step a new model
$\bar{\mathcal{M}} = \langle\bar{\mathbf{A}},\bar{\mathbf{B}}\rangle$
is estimated from the original model $\mathcal{M}
=\langle\mathbf{A},\mathbf{B}\rangle$. It can be shown that either the
model $\mathcal{M}$ defines a critical point where $\mathcal{M} =
\bar{\mathcal{M}}$, or, $\prob{O|\bar{\mathcal{M}}} >
\prob{O|\mathcal{M}}$. The iteration process stops when
$\prob{O|\bar{\mathcal{M}}} - \prob{O|\mathcal{M}} < \epsilon$, for
some suitably chosen $\epsilon$. The reestimation of $\mathbf{A}$ and
$\mathbf{B}$ can be described as:
\begin{equation}
  \bar{a}_{ij} = \frac{\text{expected number of transitions
      from\ } q_{i} \text{\ to\ } q_{j}}{\text{expected number of transitions
      from\ } q_{i}}
\label{eq:fhmm-expecttrans}
\end{equation}
\begin{equation}
\bar{b}_{j}(k) = \frac{\text{expected number of times in\ }
  q_{j} \text{\ observing symbol\ } v_{k}}{\text{expected number of times
    in\ } q_{j}}
\label{eq:fhmm-expectobs}
\end{equation}
Recall the definition of the $\alpha$ variables
\[
\alpha_{t}(i) = \prob{o_{1},\ldots ,o_{t}, X_{t}=q_{i} | \mathcal{M}}
\]
and the $\beta$ variables
\[
\beta_{t}(i) = \prob{o_{t+1},\ldots ,o_{T}, X_{T+1} = q_{N} | X_{t}=q_{i},\mathcal{M}}
\]
It is obviously not by pure chance that the $\alpha$ variables and
$\beta$ variables fit so nicely together. It is namely so that
\begin{equation}
\begin{split}
\frac{\alpha_{t}(i)\beta_{t}(i)}{\prob{O|\mathcal{M}}} &=
\frac{\prob{o_{1},\ldots ,o_{T}, X_{t}=q_{i}, X_{T+1} = q_{N} |
    \mathcal{M}}}{\prob{O|\mathcal{M}}} \\
&= \prob{X_{t}=q_{i}, X_{T+1} = q_{N} | O,\mathcal{M}} \\
\end{split}
\label{eq:fhmm-instate}
\end{equation}
i.e. the probability of being in state $q_{i}$ at time $t$ and finally
ending up in the exit state, conditioned on the observation sequence
and the model. Summed over the time index $t$, the expression
in~(\ref{eq:fhmm-instate}) can be interpreted as the expected (over
time) number of visits to state $q_{i}$, or equivalently, the expected
number of transitions from state $q_{i}$. That is
\begin{equation}
\frac{1}{\prob{O|\mathcal{M}}} \sum_{t=0}^{T} \alpha_{t}(i)\beta_{t}(i)
= \text{expected number of times in } q_{i}
\end{equation}
yields one of the sought for quantities. In a similar way it is the
case that
\begin{equation}
\frac{\alpha_{t}(i)a_{ij}b_{j}(o_{t+1})\beta_{t+1}(j)}{\prob{O|\mathcal{M}}}=
\prob{X_{t}=q_{i}, X_{t+1}=q_{j}, X_{T+1} = q_{N} | O,\mathcal{M}}
\label{eq:fhmm-fromstatetostate}
\end{equation}
i.e. the probability of being in state $q_{i}$ at time $t$ and $q_{j}$
at time $t+1$ and being able to make the exit state transition given
the observations and the model. And analogously
\begin{multline}
\frac{1}{\prob{O|\mathcal{M}}} \sum_{t=0}^{T-1}
\alpha_{t}(i)a_{ij}b_{j}(o_{t+1})\beta_{t+1}(j) =\\
\text{expected number of transitions from }q_{i}\text{ to }q_{j}
\end{multline}
can be interpreted as the expected number of transitions from $q_{i}$ to $q_{j}$
over time. Note that the exit state transition at time $T$ has to be
dealt with separately (see equation~\ref{eq:fhmm-reestexit} below). 

Using
the above equations, a convincing set of reestimation formulas would
be:
\begin{eqnarray}
1\leq i\leq N-1 & , & 2\leq j\leq N-1 \nonumber \\
\bar{a}_{ij} & = &
\frac{
{\displaystyle \sum_{t=0}^{T-1}}\alpha_{t}(i)a_{ij}b_{j}(o_{t+1})\beta_{t+1}(j)
}
{
{\displaystyle \sum_{t=0}^{T}}\alpha_{t}(i)\beta_{t}(i)
} \\
1\leq i\leq N-1 & & \nonumber \\
\bar{a}_{iN} & = &
\frac{
\alpha_{T}(i) a_{iN}
}
{
{\displaystyle \sum_{t=0}^{T}}\alpha_{t}(i)\beta_{t}(i)
} \label{eq:fhmm-reestexit}\\
2\leq j\leq N-1 & , & 1\leq k\leq K \nonumber \\
\bar{b}_{j}(k) & = &
\frac{
{\displaystyle \sum_{\substack{t=0 \\ \text{s.t.\ }o_{t}=v_{k}}}^{T}}\alpha_{t}(i)\beta_{t}(i)
}
{
{\displaystyle \sum_{t=0}^{T}}\alpha_{t}(i)\beta_{t}(i)
}
\label{eq:fhmm-reestobs}
\end{eqnarray}

The major difference in the reestimation formulas compared to the
standard model is the formula~(\ref{eq:fhmm-reestexit}), the
reestimation of the exit state transitions. The exit state transition
can only occur when time $t=T$, thus there is no summation over time
in the numerator of~(\ref{eq:fhmm-reestexit}). The effect of the exit
state transition at time $t=T$ is also that the normalization of
$\bar{a}_{ij}$ is slightly different from the standard model. In the
standard model the summation in the denominator goes from
$t=0,\ldots,T-1$. The reestimation formula~(\ref{eq:fhmm-reestobs}) is
the same as for the standard model.

The prime intent with this section has been to show the small, but yet
significant, difference between this model and the standard
model. There are a number of practical implementation issues
concerning HMMs that are consciously left out of this
presentation. Some of these will surface briefly in the remainder of
this thesis, but for a more comprehensive account the reader should
consult e.g.~Rabiner~\shortcite{rabiner:89} and
Levinson~\emph{et al.}~\shortcite{levinson+al:83} on the topics
of scaling, multiple observation sequences, choice of model type
(topology) and size and initial parameter estimation. The problem of
sparse data has been addressed in a number of papers,
including~Jelinek and Mercer~\shortcite{jelinek+mercer:80},
Church and Gale~\shortcite{church+gale:91b} and
Katz~\shortcite{katz:87}.


\section{Token Passing}
\label{sec:tp}
The term \emph{Token Passing} originates with the researchers engaged
in the speech recognition effort at the Engineering Department at
Cambridge, UK~\cite{young+al:89}. The problem of speech recognition
can simplistically be described as a recurring process of grouping
sequential items to form more meaningful items. The speech waveform is
sampled and processed for feature vectors, these are segmented into a
sequence of phones, phones are grouped into words and from there,
depending on the application of course, some sort of language model
deals with the phrases and/or sentence(s). There are obviously
interdependencies between the different levels of this hierarchically
laid out recognition process. The predictability of a certain phone
sequence, for example, is of course dependent on what words are likely
to occur in the particular context at hand. The recognition problem of
the levels varies in difficulty, and the complexity of the problem is
partly due to the strength of the dependencies on neighboring levels,
the weaker the dependencies the harder the problem. Recognizing a
sequence of phones as a word is considerably easier than finding
phones in the feature vector stream. By transitivity one can say that
all levels are dependent on each other. The main advantage of the
Token Passing framework/algorithm (TP) is the elegant way in which
dependencies between knowledge sources are maintained, i.e. the
coupling of neighboring layers. Another feature is the flexibility of
the framework, alternative recognition algorithms can be used in
various layers. Furthermore it can quite easily be adapted to generate
multiple alternative solutions and to incorporate search heuristics.

In speech recognition as well as in text recognition there is a
\emph{basic recognition unit}. In the hierarchical speech recognition
scheme laid out above the recognition of a phone as a sequence of
feature vectors is the basic recognition task, hence the phone is the
basic recognition unit. In the textual case one can have morphemes,
lemmas or word-forms as the basic recognition units. We assume from
here on that the word-form is the basic recognition unit. Each basic
recognition unit is represented by a finite state network, and each
network, or model, holds a \emph{model identifier}, the word-form
string that it models. Each state $i$, $j$ of the network is
connected by a \emph{transition cost} $p_{ij}$, and with each state
$j$ there is an associated \emph{local cost function} $d_{j}(c)$.  A
path $I = i_{1},\ldots,i_{T}$ through the network represents one
possible alignment of the network to the input characters
$C=c_{1},\ldots,c_{T}$. Assume, for the time being, that $C$ is a word
form.

\begin{figure}[ht]
\begin{center}
\includegraphics{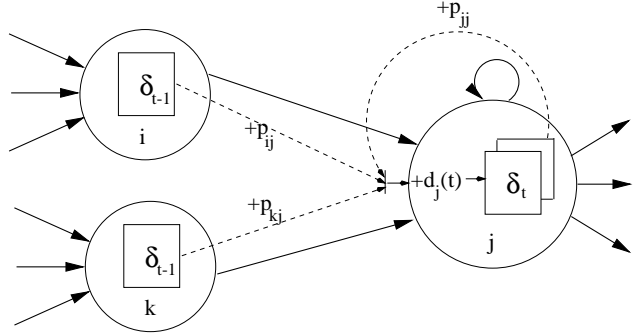}
\caption{Token propagation\label{fig:token-propagate}}
\end{center}
\end{figure}

Let each state of the network be capable of holding a passable
token. At time $t$ the token in state $j$ holds the (partial) minimum
cost alignment of $c_{1},\ldots,c_{t}$ and the network, that ends in
state $j$, i.e. the token represents the head of a path through the
network. The minimum cost alignment $\delta_{t}(j)$ can be computed by
the recursion
\begin{equation}
\delta_{t}(j) = \min_{i}\left[ \delta_{t-1}(i) + p_{ij}\right] + d_{j}(c_{t})
\label{eq:tp-mincostalign}
\end{equation}
At each discrete time point copies of tokens are propagated between states and
the minimum cost alignment is updated according to
equation~(\ref{eq:tp-mincostalign}). The algorithm is illustrated in
Figure~\ref{fig:token-propagate}. The token is the box containing the
cost $\delta_{t}$ inside each state.

The attentive reader has probably noticed the similarity between
equation~(\ref{eq:tp-mincostalign}) and the recursion
formula~(\ref{eq:fhmm-phiinduct}) of the Viterbi algorithm above. The
similarity is, of course, not accidental. This, however, should not lead
to the conclusion that the TP algorithm is just an unorthodox way of
implementing the Viterbi algorithm. It is more general than that, or
more accurately, the passing of tokens \emph{within} the basic
recognition unit, equation~(\ref{eq:tp-mincostalign}), can be seen as
a generalization of a number of other `cost functions', e.g. many of
the distance metrics used in isolated word error correction. Consider,
for example, the \emph{Weighted Levenshtein Distance} (WLD)
metric~\cite{okuda+al:76} mentioned in Chapter~\ref{ch:back}. The WLD
measures the distance between two words $X$ and $Y$ as the number of
substitutions ($k_{i}$), insertions ($m_{i}$) and deletions ($n_{i}$)
with weights $p$, $q$ and $r$ attached to the respective error
categories. Okuda~\emph{et al.} express this as
\begin{equation}
\text{\textsl{WLD}}(X\rightarrow Y) = \min_{i}\left[pk_{i} + qm_{i} + rn_{i}\right]
\label{eq:tp-wlddef}
\end{equation}
i.e. the minimum number of weighted errors it takes to transform the
word $X$ into $Y$. To
calculate~(\ref{eq:tp-wlddef})~Okuda~\emph{et al.}\footnote{The
  notation used here is somewhat different from that used
  by~Okuda~\emph{et al.}} devised a recursive algorithm that
operates on substrings of increasing length of the the two words,
where $\text{\textsl{WLD}}_{t}(j)$ denotes
$\text{\textsl{WLD}}(x_{1},\ldots,x_{j}\rightarrow y_{1},\ldots
y_{t})$.

\begin{multline}
\text{\textsl{WLD}}_{t}(j) = \min \left\{
    \begin{array}{l}
      \text{\textsl{WLD}}_{t}(j-1) + q \\
      \text{\textsl{WLD}}_{t-1}(j-1) + p_{j-1,t-1} \\
      \text{\textsl{WLD}}_{t-1}(j) + r
    \end{array} \right.\\
\text{where}\hspace{2em}p_{j-1,t-1} = \left\{ \begin{array}{ll}
        0 & \text{if\ } y_{t-1} = x_{j-1}\\
        p & \text{otherwise}
        \end{array} \right.
\label{eq:tp-okudawld}
\end{multline}

To adapt the algorithm to TP it is necessary to take a slightly
different view. In the~Okuda~\emph{et al.} algorithm they were
comparing two strings. Here we have a network and a string and it is
not appropriate to change state without consuming anything from the input,
as is done in the topmost of the three expressions
in~(\ref{eq:tp-okudawld}). (The expression adds the penalty for
insertion to the metric.) The expression~(\ref{eq:tp-okudawld}) above
has to be reformulated to suit the TP algorithm. If the word $X$ is
represented by a network in which each state $j$ corresponds to
$x_{j}$, the $j$:th character in $X$ and $Y=y_{1},\ldots,y_{T}$ is the
transformed word, the algorithm presented by~Okuda~\emph{et al.}
can be reformulated as

\begin{equation}
\text{\textsl{WLD}}_{t}(j) = \min \left\{\begin{array}{l}
\text{\textsl{WLD}}_{t-1}(i) + q \hspace{1em}(\text{if\
  }x_{j}=x_{\text{\textsl{ins}}})\\
\text{\textsl{WLD}}_{t-1}(i) + \left\{ \begin{array}{ll} 0 & \text{if\ }
    y_{t}=x_{j}\\
    p & \text{otherwise}\end{array} \right.\\
\text{\textsl{WLD}}_{t-1}(i) + (j-i-1)r \hspace{1em}j-i\geq 2
\end{array} \right.
\label{eq:tp-tpwld}
\end{equation}

\begin{figure}[h]
\begin{center}
\includegraphics{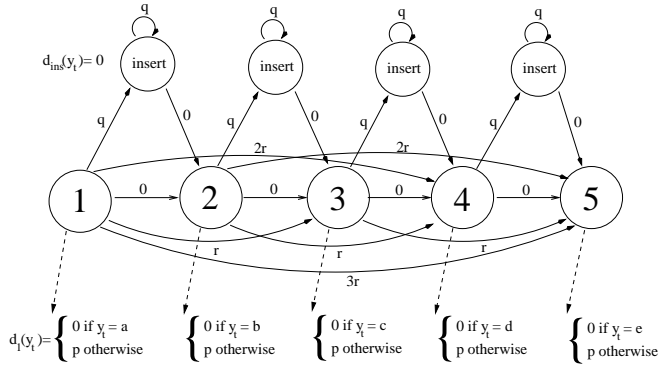}
\caption{Computation of weighted Levenshtein distance with Token Passing\label{fig:wld}}
\end{center}
\end{figure}

The algorithm for computing the WLD presented
by~Okuda~\emph{et al.} can be realized by~(\ref{eq:tp-tpwld}) and
the network topology of Figure~\ref{fig:wld}. The word represented by
the network in the figure is \xpl{abcde}\footnote{
  The model is approximate in the sense that it does not properly cope
  with insertions and deletions at the end-points of the modeled
  word. This can easily be fixed with entry and exit states. The point
  is just to show that TP can be used for a variety of computational
  tasks and to a large extent it is only a matter of altering the
  network topology.}.
The dedicated `insert' states are there to make sure that inserted
characters are penalized equally hard independent of which character
is inserted and in which position. The insert state is denoted
$x_{\text{\textsl{ins}}}$ in~(\ref{eq:tp-tpwld}). The relation between
equations~(\ref{eq:tp-mincostalign}) and~(\ref{eq:tp-tpwld}) is simply
such that the transition cost $p_{ij}$ implements the insertion weight
$(q)$ and the deletion weight $(r)$ whereas the local cost function
$d_{j}(c_{t})$ implements the substitution weight.

One of the fundamental ideas of TP is to separate out the low-level
pattern-matching algorithm(s) of the basic recognition units from the
higher level control mechanisms. In the case of textual input the
low-level task is to recognize words in the character input
stream. The hypothesizing of word occurrences in the input is
controlled by the higher level, the language model. There are many
ways in which the pattern matching function can be controlled, or
guided. These issues will be discussed further in
Section~\ref{sec:ctr} and Chapter~\ref{ch:ae}, the account below
focuses on the TP component that facilitates language modeling.

If the character input stream $C=c_{1},\ldots,c_{T}$ contains several
words, there must be a way that allows for tokens to be passed from one
basic recognition unit to another. This can be accomplished simply by
connecting subnetworks to form larger composite networks. In doing so,
however, it is necessary to record transitions between subnetworks in
some way. We are interested in the actual word sequence, not just the
cost of the best alignment of states to the input. To keep track of
what word models a token has passed through on its way to the end of
the input, the token is supplemented with a \emph{path
identifier}. The path identifier is a pointer to a \emph{Word Link
Record} (WLR) that contains word boundary information.

\begin{figure}[h]
\begin{center}
\includegraphics{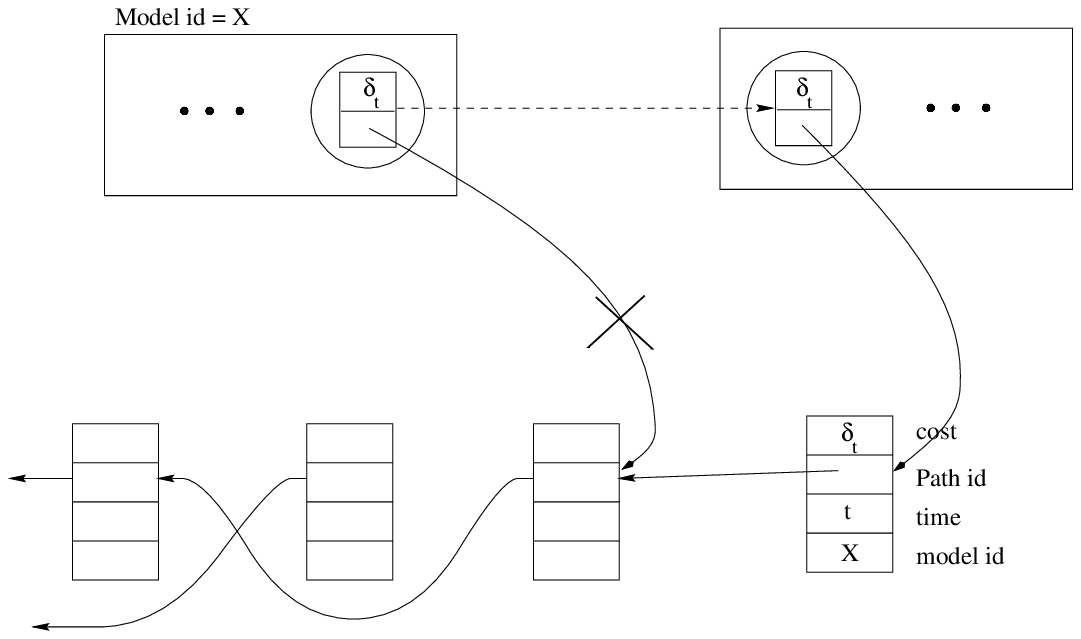}
\caption{Inter-network token propagation\label{fig:wlr}}
\end{center}
\end{figure}

As a token is propagated from one subnetwork to another, a new WLR is
created and the token is set to point to the new WLR, which in turn is
set to point to the WLR that the token was pointing to prior to the
inter-network token propagation. Figure~\ref{fig:wlr} visualizes the
process. The amount of information put in the WLR can depend on the
language model used (see below), minimally it should contain the cost,
path identifier and the model identifier. In Figure~\ref{fig:wlr} the
time (character stream position) at which the word boundary occurred
is also appended to the WLR.

The time index $t$ is not updated after the transition which means
that identifying the word boundary does not imply the consumption of
any character from the input stream. This could equally well be the
other way around. The details of exactly how, and under what
circumstances, tokens are passed between the basic recognition units
are left to Section~\ref{sec:ctr}.


\section{Isolated Word Recognition}
\label{sec:iwr}
The problem of \emph{Isolated Word Recognition} (or isolated word
error correction) is to detect and correct erroneous words without the
use of any contextual dependencies. A word is viewed in
isolation. Normally words do not just appear individually,
they most often appear in a text, a stream of characters. The
identification of words in such a stream is called tokenization, and
Isolated Word Recognition (IWR), to be successful, relies on correct
tokenization. A consequence of this implicit assumption, that is
generally not spelled out regarding IWR, is that an erroneous token is
assumed to be the misspelling of exactly one word. Under these
conditions it is only the nonword misspellings (single or multiple)
that can be corrected using IWR.

The classical scheme used to correct misspellings in isolation
involves three steps:
\begin{description}
\item[{\mdseries\textsf{Step 1.}}] Detect the erroneous token
\item[{\mdseries\textsf{Step 2.}}] Generate alternative correction candidates
\item[{\mdseries\textsf{Step 3.}}] Rank the alternative candidates
\end{description}
The detection step (almost) always means to compare the token with
words in a dictionary. If the token is not equal to any of the entries
in the dictionary, it is misspelled. The generation of candidates can
be performed in a number of ways and it is often interleaved with the
ranking process which often involves some sort of distance metric. The
dictionary word that is closest to the error-token, using the distance
metric, is the highest ranked candidate and should be chosen for
correction (cf. Section~\ref{sec:back-classical}).

The remainder of this section will describe how the Hidden Markov
Model can be used to perform Isolated Word Recognition within the
Token Passing framework. In the approach taken here the three steps are
melted down into one single process, detection, generation and ranking
of candidates is simply a matter of computing the probability of an
error token given the possible words in the dictionary. The
\emph{noisy channel} is an illustrative metaphor when using
probabilistic methods.

\begin{figure}[h]
\begin{center}
\includegraphics{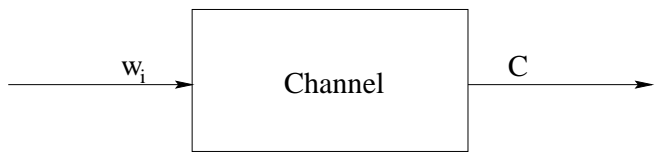}
\caption{The noisy channel\label{fig:noisy-channel}}
\end{center}
\end{figure}

A word is inserted at one end of the channel and from the other end
comes a distorted version of that word. The aim is to restore the
original.

Let $W$ be the $M$ word vocabulary $W=\{w_{1},\ldots,w_{M}\}$. Given
the character sequence $C=c_{1},\ldots,c_{T}$, which may be erroneous
or not, the most likely correction is the $w_{i}$ that maximizes
$\prob{w_{i}|C}, 1\leq i\leq M$. This number is hard to calculate, but
fortunately there is Bayes' rule
\begin{equation}
\prob{w_{i}|C} = 
\frac{
\prob{C|w_{i}} \prob{w_{i}}
}
{\prob{C}}
\label{eq:iwr-bayes}
\end{equation}
Choosing the word that best matches the character sequence is not
dependent on the probability of the sequence, which is given, so
finding the $w_{i}$ that maximizes the numerator in Bayes rule seems
like a good idea. Modeling each word $w_{i}$ of the vocabulary with an
HMM $\mathcal{M}_{w_{i}}$ and making the obviously faulty
assumption that all words are equiprobable, finding the word $w^{*}$
that best matches the character sequence $C$ is simply
\begin{equation}
w^{*} = \argmax{w_{i}}{\prob{C|\mathcal{M}_{w_{i}}}}
\label{eq:iwr-w-star}
\end{equation}
It should be noted that there is no error detection being performed
here in the normal sense of the word. It might very well be the case
that $w_{i}=C$ for some $i$, i.e. $C$ is not misspelled at all. It
would of course suffice to do a simple string match between $C$ and
the strings of the vocabulary words to find this out. The scheme
described here presupposes Step 1 above, or, one can think of the
detection of the error as something that is found out after the
ranking of the candidates. If the highest scoring candidate $w^{*}\neq
C$, then a spelling error has been detected (and corrected at the same
time).

The number $\prob{C|\mathcal{M}_{w_{i}}}$ can be efficiently computed,
as shown in Section~\ref{sec:fhmm}, but how should the words be
modeled using HMMs?

\begin{figure}[ht]
\begin{center}
\includegraphics{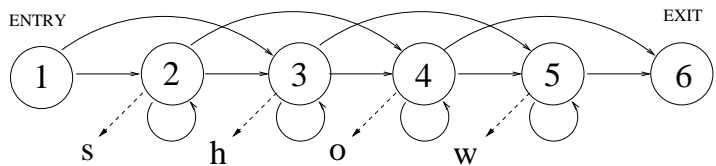}
\caption{The structure of $\mathcal{M}_{show}$\label{fig:m-show}}
\end{center}
\end{figure}

The type of model used here is the so called left-to-right model, for
which
\[
a_{ij} = 0 \hspace{1cm}\text{when\ } j<i
\]
Often additional constraints are placed on the left-to-right model,
such as
\[
a_{ij} = 0 \hspace{1cm}\text{when\ }j>i+\Delta
\]
to make sure that large changes in the state indices do not occur. In
the left-to-right model of Figure~\ref{fig:m-show} $\Delta$ is 2.

The solid arrows represent transitions with non-zero
probabilities. The dashed arrows indicate what this particular model
is biased towards. State $2$, for example, can have non-zero
probabilities for all observables, but is strongly biased towards
\texttt{s}, i.e. $b_{2}(v_{k})$ has the highest probability for
$v_{k}=$\,\texttt{s}.

The general idea is that the states of the model represent character
positions in the word modeled. Recall the four basic error types:
character insertion, deletion, substitution and transposition. The
model above can deal with all four, although perhaps not ideally with
transpositions. The transition distribution $\mathbf{A}$ makes it
possible to handle deletions and insertions. The character
distribution $\mathbf{B}$ makes it possible to handle substitutions
and for transpositions a combination of both is needed. The model of
Figure~\ref{fig:m-show} is not necessarily the best choice, some of
the restrictions imposed on it can seem a bit strange. There may be
any number of insertions (the looping arcs), but more than one
consecutive deletion will not be very well handled since
$\Delta=2$. If there were a backward chaining arc from each state to
the preceding state, transpositions would be more easily recognized. A
character sequence like \xpl{shwo} could then be generated
(recognized) with the state sequence $1\rightarrow 2\rightarrow
3\rightarrow 5\rightarrow 4\rightarrow 6$ which would then score
higher than the sequence $1\rightarrow 2\rightarrow 3\rightarrow
5\rightarrow 5\rightarrow 6$ which will probably be the highest
scoring sequence using the model in Figure~\ref{fig:m-show}. Relaxing
these constraints however, would lead to a close to ergodic model, and
the computational advantages of the left-to-right model would be
lost. Furthermore, restrictions on the number of errors that can occur
in a single word are supported by findings in samples of spelling
errors~(cf.~Damerau~\shortcite{damerau:64}).

The quantity $\prob{C|\mathcal{M}_{w_{i}}}$ can be thought of as a
measure of similarity between the string $w_{i}$, modeled by
$\mathcal{M}_{w_{i}}$, and the string
$C$. Angell~\emph{et al.}~\shortcite{angell+al:83} discuss
alternative \emph{string similarity measures} and distinguish three
types: material, ordinal and positional similarity. \emph{Material}
similarity measures the extent to which a pair of strings contains
identical characters, \emph{ordinal} similarity measures the extent to
which the characters are in the same order, and \emph{positional}
similarity measures the extent to which the characters are in
corresponding positions in the two strings. Most of the classical
techniques use one or sometimes two of these similarity measures to
perform isolated word error correction. It is interesting to see that
the approach presented here is actually a mixture of all three
types. The material similarity is expressed by the probability to
observe a particular character, and the fact that this probability is
conditioned on the states also makes it a positional similarity
measure. The transition probabilities between states capture the
ordinal similarity

The probability that a particular model generated a given character
sequence is, of course, dependent on how the model has been trained and
also what the initial (pre-training) model looked like. Ideally one
would like to have a large set of naturally occurring misspellings for
each word in the vocabulary to train the respective models with. Such
an error corpus is unfortunately not available so the corpus has to be
artificially generated. These issues will be described in some detail
in the sections below reporting the experiments.

In computing the probability of the character string for all the
$\mathcal{M}_{w_{i}}$ in the vocabulary, it is not really the number
$\prob{C|\mathcal{M}_{w_{i}}}$ as such that is interesting, it is the
probability relative to the other words in the vocabulary that
matters. To meet this end the Viterbi algorithm computes a good enough
approximation of the probability of a string\footnote{Thus
  $\prob{C,Q^{*}|\mathcal{M}_{w_{i}}}$ is used as an approximation for
  $\prob{C|\mathcal{M}_{w_{i}}}$. The Viterbi algorithm is the
  algorithm generally used in applications performing some sort of
  recognition with HMMs.}. When the length of the
character sequence grows, an arithmetic underflow condition can occur
in the forward-backward algorithm, i.e. as $t\rightarrow\infty$,
$\alpha_{t}(i) \rightarrow 0$. In the Viterbi algorithm, since it
maximizes instead of sums (see equations~(\ref{eq:fhmm-alphainduct})
~and~(\ref{eq:fhmm-phiinduct})), logarithms of the probabilities can be
used and underflow conditions do not arise. The logarithms of the
probabilities can, of course, be computed beforehand and does not
constitute an increased computational load in the actual recognition
task. As well as maximizing the logarithm of the probability of a path
(state sequence), one can of course minimize the negative logarithm of
the probability of the path, the \emph{cost} of the state
sequence. Let $\delta_{t}(j)$ be the counterpart of $\phi_{t}(j)$ in
equation~(\ref{eq:fhmm-phidef}) and denote the cheapest state sequence
that ends in $q_{j}$ and accounts for the first $t$ characters. The
reformulation of the Viterbi algorithm is straightforward:

\begin{description}
\item[Initialization]
\begin{eqnarray}
1\leq i\leq N & & \nonumber \\
\delta_{0}(i) & = &
\left\{ \begin{array}{rl}
    0 & \mbox{if $i=1$} \\
    \infty & \text{otherwise}
  \end{array}
\right.
\end{eqnarray}
\item[Induction]
\begin{eqnarray}
 1 \leq t\leq T & , & 2\leq j\leq N-1 \nonumber \\ 
 \delta_{t}(j) & = & \min_{1\leq i\leq
   N-1}\left[ \delta_{t-1}(i) + (-\log a_{ij})\right] + (-\log b_{j}(c_{t}) \label{eq:fhmm-deltainduct}
\end{eqnarray}
\item[Termination]
\begin{eqnarray}
  -\log \prob{C,Q^{*}|\mathcal{M}} & = & \min_{1\leq
    i\leq N-1}\left[ \delta_{T}(i) + (-\log a_{iN})\right]
\end{eqnarray}
\end{description}

The adaption of the Viterbi algorithm to Token Passing is quite
trivial. Equations~(\ref{eq:fhmm-deltainduct})
and~(\ref{eq:tp-mincostalign}) are virtually identical. The transition
cost is the negative logarithm of the transition distribution of the
HMM and the local cost function is the negative logarithm of the
observation symbol distribution.

Algorithm~\ref{alg:iwr-step_model} below describes how the Viterbi
algorithm is computed with TP in a single HMM network. The algorithm
works for any network topology, but it might be useful to keep the HMM
of Figure~\ref{fig:m-show} in mind.

The HMM network has $N$ states numbered $1$ to $N$, where state $1$ is
the entry state and state $N$ is the exit state. Each token $\tau$
contains only the cost, as computed
by~(\ref{eq:fhmm-deltainduct}). Let $\tau_{i}(\delta_{t})$ denote the
cost of the token in state $i$ at time $t$. The \emph{start token} has
cost $-\log 1 = 0$. The \emph{null token} has cost $-\log 0 =
\infty$. The input character sequence is $C=c_{1},\ldots,c_{T}$.

\begin{algorithm}
\caption{Viterbi decoding using Token Passing with a single HMM
  network. The boxed portion is the \emph{Step\_model}$(c_{t})$
  procedure that is reused in Algorithms~\ref{alg:iwr-iwr}
  and~\ref{alg:ctr} below.}
\begin{algorithmic}[1]
\STATE{At time $t=0$}
\STATE{\ \ Put start token in the entry state}
\STATE{\ \ Put null tokens in all other states}
\FOR{$t=1$ to $T$}
\begin{boxedminipage}{10.5cm}
\FORALL{states $i < N$}
\STATE{Pass a copy of the token $\tau_{i}$ to all connecting states $j$:\\
  $\tau_{j}(\delta_{t}) = \tau_{i}(\delta_{t-1}) + (-\log a_{ij}) +
  (-\log b_{j}(c_{t}))$}
\ENDFOR
\STATE{Discard all original tokens}
\FORALL{states $i < N$}
\STATE{Find the minimum cost token and discard the rest}
\ENDFOR
\FORALL{states $i$ connected to state $N$}
\STATE{Pass a copy of the token $\tau_{i}$ to state $N$:\\
  $\tau_{N}(\delta_{t}) = \tau_{i}(\delta_{t}) + (-\log a_{iN})$}
\ENDFOR
\STATE{In state $N$: Find the minimum cost token and discard the rest}
\end{boxedminipage}
\ENDFOR
\end{algorithmic}
\label{alg:iwr-step_model}
\end{algorithm}

In Isolated Word Recognition it is assumed that the input
$c_{1},\ldots,c_{T}$ is exactly one distorted word. This assumption
implies that there is no point in making the exit state transition
before $t=T$, i.e. lines 12 and 13 can be put outside of the main loop
in Algorithm~\ref{alg:iwr-step_model}. The \emph{step\_model}
procedure as displayed above will, however, be reused in a situation in
which this assumption does not hold, thus the exit state transition
has to be hypothesized for each value of $t$, i.e. any character
$c_{t}$ may be the last character in the word modeled by the basic
recognition unit.

When the entire input has been consumed, the cost of the token in the
exit state of the model represents the best alignment of states to the
input, i.e.
\[
\mathcal{M}_{w_{i}}\!\!\rightarrow\!\!\tau_{N}(\delta_{T}) = -\log(\prob{C,Q^{*}|\mathcal{M}_{w_{i}}})
\]
where $Q^{*}$ is the optimal path\footnote{
  The actual state sequence through the basic recognition unit is not
  recorded in any way in Algorithm~\ref{alg:iwr-step_model} and it can not
  be restored.} through the model and $w_{i}$ is the
word (the model identifier) that is modeled by the network. To perform
Isolated Word Recognition over an $M$ word vocabulary
$W=\{w_{1},\ldots,w_{M}\}$ it is, of course, necessary to execute
Algorithm~\ref{alg:iwr-step_model} for all the $M$ basic recognition
units and then choose $w_{i}$ as the correction hypothesis if
$\mathcal{M}_{w_{i}}\!\!\rightarrow\!\!\tau_{N}(\delta_{T})$ has the
lowest cost.

During the computation of the \emph{step\_model} procedure, many
models will display significantly different costs. This is, of course,
the whole idea. The point however, is that these differences will
start to show up when only a relatively small number of input
characters have been processed. Consider for example the input
$C=$~\xpl{heuristics}, and assume that both $\mathcal{M}_{heuristics}$
and $\mathcal{M}_{exhaustive}$ are in the vocabulary. Then
\[
\mathcal{M}_{heuristics}\!\!\rightarrow\!\!\tau^{*}(\delta_{t}) \ll
\mathcal{M}_{exhaustive}\!\!\rightarrow\!\!\tau^{*}(\delta_{t})
\]
after just a few characters. ($\tau^{*}(\delta_{t})$ denotes the
best token in any of the states at time $t$.) In a situation like this
the \emph{Beam Search} heuristic can be useful. All models that are
outside the beam are deactivated (pruned). The beam is defined as
the difference between the globally optimal model
$\mathcal{M}^{*}\!\!\rightarrow\!\!\tau^{*}(\delta_{t})$, where
\begin{equation}
\mathcal{M}^{*}\!\!\rightarrow\!\!\tau^{*}(\delta_{t}) =
\min_{\mathcal{M}_{w_{i}}}
\left[ \mathcal{M}_{w_{i}}\!\!\rightarrow\!\!\tau^{*}(\delta_{t})
\right]
\label{eq:iwr-globalopt}
\end{equation}
(the cost of the minimum cost token of all states of all models), and
a preset threshold $B$. The
overhead of the beam search heuristic is the computation
of~(\ref{eq:iwr-globalopt}). If at any time $t$ it is the case that
the model $\mathcal{M}_{w_{i}}$ is outside the beam, i.e.
\begin{equation}
\mathcal{M}_{w_{i}}\!\!\rightarrow\!\!\tau^{*}(\delta_{t}) >
\mathcal{M}^{*}\!\!\rightarrow\!\!\tau^{*}(\delta_{t}) + B
\label{eq:iwr-beam}
\end{equation}
$\mathcal{M}_{w_{i}}$ can be deactivated. The algorithm for isolated
word recognition using HMMs within the TP framework is given in
Algorithm~\ref{alg:iwr-iwr}.

\begin{algorithm}
\caption{Isolated Word Recognition with Beam Search}
\begin{algorithmic}[1]
\STATE{At time $t=0$}
\STATE{\ \ All models are activated}
\STATE{\ \ Put start token in the entry state of all models}
\STATE{\ \ Put null tokens in all other states}
\STATE{\ \ $\mathcal{M}^{*}\!\!\rightarrow\!\!\tau^{*}(\delta_{t}) = \infty$}
\FOR{$t=1$ to $T$}
\FORALL{models $\mathcal{M}_{w_{i}} 1\leq i\leq M$}
\IF{$\mathcal{M}_{w_{i}}$ is active}
\IF{(\ref{eq:iwr-beam})}
\STATE{Deactivate $\mathcal{M}_{w_{i}}$}
\ELSE
\STATE{\emph{step\_model}$(c_{t})$ with $\mathcal{M}_{w_{i}}$}
\ENDIF
\ENDIF
\ENDFOR
\STATE{compute
  $\mathcal{M}^{*}\!\!\rightarrow\!\!\tau^{*}(\delta_{t})$
  according to~(\ref{eq:iwr-globalopt})}
\ENDFOR
\STATE{Return \dxpl{word} as the reading of $C$ where\\
$
\text{\dxpl{word}} = 
\arg\!\!{\displaystyle\min_{w_{i}}} \left[ \mathcal{M}_{w_{i}}\!\!\rightarrow\!\!
  \tau_{N}(\delta_{T}) \right]
$\ \ \ \COMMENT{Only the active $\mathcal{M}_{w_{i}}$}
}
\end{algorithmic}
\label{alg:iwr-iwr}
\end{algorithm}

At time $T$ the exit state of all active models is inspected,
the model with the lowest cost in its exit state is the one that best
matches the character sequence, and the model's identifier denotes the
preferred reading of the character sequence.

The technique described above, although quite successful at correcting
misspellings, has some obvious shortcomings in most practical
applications. In running text the word boundaries are uncertain and
erroneous assumptions in this respect will result in segmentation
errors being treated as spelling errors. Another problem is that
real-word errors will go undetected. Further, the fact that words are
not equally likely in a given context can not be overlooked. The way
to address this problem is to look at the words in the context in
which they appear. Decisions regarding word boundaries must take into
account the fact that segmentation errors may be present in the input.


\section{Connected Text Recognition}
\label{sec:ctr}
In \emph{Connected Text Recognition} (CTR) the character sequence can
contain any number of words. The task is to find the most likely word
sequence even though the word boundaries may be obscured (segmentation
errors) and the words themselves are distorted (misspellings).
\begin{figure}[h]
\begin{center}
\includegraphics{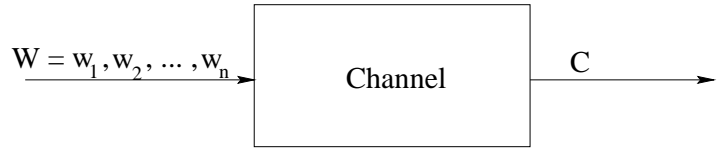}
\caption{The word sequence noisy channel\label{fig:noisy-utterance-channel}}
\end{center}
\end{figure}
Thus we want to find the word sequence $W$ that maximizes the quantity
$\prob{W|C}$. This number is impractical to compute, but again
\begin{equation}
\prob{W|C} = \frac{\prob{C|W}\prob{W}}{\prob{C}}
\label{eq:ctr-bayes}
\end{equation}
according to Bayes' rule. The denominator is given, so to find
$W^{*}$, the most likely word sequence, it suffices to maximize the
numerator of~(\ref{eq:ctr-bayes}) over the alternative word sequences.
\begin{equation}
\prob{W^{*}|C} = \max_{W}\left[ \prob{C|W}\prob{W}\right]
\label{eq:ctr-numerator-max}
\end{equation}
The first factor of the right hand side
of~(\ref{eq:ctr-numerator-max}) is the channel characteristics
(see Figure~\ref{fig:noisy-utterance-channel}) that models how word
sequences are distorted. The second factor is the prior probability of
a word sequence, the language model. Probabilistic language models
usually exploit the local context to predict the occurrence of a
word. Assume, for the sake of this formal account, that the
unspecified language model $\mathcal{G}$ can be used to predict the
probability of a word, $\prob{w_{i}|\mathcal{G}}$. In this way the
prior probability of a word sequence $W = w_{1},\ldots,w_{n}$ can be
computed as
\begin{equation}
\prob{W} = \prod_{i=1}^{n} \prob{w_{i}|\mathcal{G}}
\label{eq:ctr-word-seq-prob}
\end{equation}

Irrespective of the language model used, (and even without a language
model), the problem remains to find the most likely segmentation of
the input character stream $C = c_{1},\ldots,c_{T}$, i.e. each word in
the word sequence has to have a portion of the input characters
assigned to it. The reason is, of course, that the channel
characteristics must come to bear on the overall likelihood of the
word sequence. For a given word sequence $W = w_{1},\ldots,w_{n}$ the
most likely segmentation can be found by maximizing over all possible
word boundaries, i.e.
\begin{equation}
\prob{C|W} = \max_{\substack{1\leq t_{i},t_{i}'\leq T\\ t_{i}<t_{i}'}} \left[
  \prod_{i=1}^{n} \prob{c_{t_{i}}^{t_{i}'}|w_{i}}\right]
\label{eq:ctr-word-segm-prob}
\end{equation}
where $c_{t_{i}}^{t_{i}'}$ denotes
$c_{t_{i}},c_{t_{i}+1},\ldots,c_{t_{i}'-1},c_{t_{i}'}$, the character
sequence `assigned' to word $w_{i}$. Note that $t_{i-1}'+1 =
t_{i}$. Equations~(\ref{eq:ctr-word-segm-prob}),~(\ref{eq:ctr-word-seq-prob})
and~(\ref{eq:ctr-numerator-max}) can be used to formally define the
Connected Text Recognition problem in equation~(\ref{eq:ctr-prob}).
\begin{equation}
\prob{W^{*}|C} = \max_{\substack{W\\ 1\leq t_{i},t_{i}'\leq T\\ t_{i}<t_{i}'}} \left[
  \prod_{i=1}^{n} \prob{c_{t_{i}}^{t_{i}'}|w_{i}} \prob{w_{i}|\mathcal{G}}\right]
\label{eq:ctr-prob}
\end{equation}

Finding words in distorted text is quite similar to the speech
recognition problem. The problem of \emph{Connected Speech
Recognition} (CSR) is to recognize and segment out the elements of the
continuous speech signal without knowing the starting point or the
end-point of any of the elements. The elements modeled can be phones,
subphones or sometimes with a small vocabulary they are words.

It is interesting to note the differences and similarities between
text processing and speech processing. The primitive input symbol in
the text case is, of course, the character or the keystroke. In the
speech case, without going into the hardships of signal processing,
the primitive input symbol is the feature vector. In most speech
recognition systems the feature vector is continuous and has around
20~-~40 coefficients and the speech signal is sampled at about 100 Hz,
see for example \cite{deller+al:93}. Whatever statistical model is used to
model, say, a spoken word as a sequence of, say 50, feature vectors, it
is obvious that there can never be a perfect match. The word model
that is a closest match is chosen. Looking at a spoken word as a
sequence of feature vectors one can thus say that the word is always
`misspelled', where the norm is an imagined sequence of feature
vectors.

Looking at speech as a sequence of feature vectors, and text as a
sequence of characters, a crucial difference is that there is no
counterpart to the space character in speech. This makes segmentation
a primary concern in CSR. In CTR the segmentation task is quite easy as
long as the space character is properly placed but can get quite hard
when it is not. The point here is that if we agree to carry the
error types of text processing over to speech processing, we see that
speech is virtually littered with misspellings and segmentation
errors. It is therefore close at hand to see what the methods used in
the difficult speech recognition task can do in the relatively simple
text recognition problem.

The present computational model can straightforwardly be put to use
on the Connected Text Recognition problem using a layering scheme of
the different knowledge sources. The bottom layer, the string pattern
matcher, is called the \emph{Orthographic Decoder} (OD). The OD
consists of a set of word modeling HMMs, one for each word in the
vocabulary, very much like in the previous section. Each word modeling
HMM can assign a probability to the hypothesis that a certain
substring is the word modeled by the network, i.e. the first factor
in~(\ref{eq:ctr-prob}). Subsequent layers, on top of the Orthographic
Decoder, is called the \emph{Linguistic Decoder} (LD). The LD is the
component that hypothesizes word occurrences in the input using some
language model, i.e. the second factor in~(\ref{eq:ctr-prob}). The
information flow between the LD and the OD is such that the LD
predicts that a certain word is present at a particular point in the
character input stream, and the OD reports back the confidence of the
match. It is thus a top-down process.

The Orthographic Decoder and Linguistic Decoder to some extent have
complementary responsibilities in the recognition and error correction
process. The string matching OD is primarily responsible for nonword
errors, both regular spelling errors and segmentation errors. The LD
reduces the search space and decides on `close calls'. In an utterance
like: \xpl{\ldots in the \hgl{aboue} table}, the OD would probably
rule \xpl{above} and \xpl{about} just about equally likely, but since
\xpl{above} is more linguistically plausible in this context, the LD
will (hopefully) rule out \xpl{about}. The predictive power of the LD
is of course even more crucial when dealing with real-word errors
since the OD will assign a good match to the `wrong' word. There is
always a trade-off situation going on between the LD and the OD in the
recognition process. If both favor the same hypothesis, it will be a
clear winner. If not, as in the case with the real-word errors, there
will be several viable hypotheses.

It was mentioned above that the Linguistic Decoder may consist of
several layers. In the dialogue application described above, for
example, it could be useful to model a dialogue in terms of user
utterances and system responses. An utterance can be given a phrase
structure or a bracketing in terms of phrases, and phrases can be
modeled as word sequences. In this hypothetical scenario the LD would
consist of three layers, dialogue, utterance and phrase. The
generalization from one to several LD layers is trivial, so the Token
Passing account of Connected Text Recognition below will be restricted
to a single layer Linguistic Decoder that models utterances or
sentences in terms of word sequences.

In Connected Text Recognition, each Orthographic Decoder HMM network
models one word of the vocabulary. This is very much like the idea
behind Isolated Word Recognition in the previous section. There is,
however, one big difference. In CTR the input character stream
contains word boundaries, usually realized by the space character. In
finding the best segmentation of the input stream the space character
is of course crucial, but we do not want to make a hard decision
regarding the location of the word boundary based solely on the fact
that there is a space character in a certain position in the
input. The OD must \emph{model} word boundary locations, treating the
space character like any other character. There are different ways of
doing this. One way would be to have an HMM network to recognize word
boundaries. A single space character, for example, would then make up
the word \xpl{\symbol{32}}, modeled by
$\mathcal{M}_{<space>}$\footnote{Note that what constitutes a word is
quite arbitrary. Since the space character is just another character,
one might have \xpl{as soon as possible} e.g. as a `word'. In the
computational scheme described here, a word is merely a sequence of
characters that has a network in the OD modeling it.}. An utterance
like: \xpl{show me all cars} would then be segmented as
\begin{list}{{\tt ==>}}{\setlength{\rightmargin}{.5cm}}
        \inputitem{utterance}\underline{show}\,\underline{\
          }\,\underline{me}\,\underline{\
          }\,\underline{all}\,\underline{\
          }\,\underline{cars}\label{eq:ctr-segmentation1}
\end{list}
This is not such a good idea, however, since language modeling would
become unjustifiably expensive. For example, to model utterances with
bigrams would in fact require trigrams since each pair of proper words
is divided by the information-poor space-word. A different solution is
needed.

The space characters present in the utterances are treated as parts of
the word models. The solution is quite simple. The OD HMMs that are
used in CTR have an extra initial state added to them that is biased
towards recognizing the space character.
\begin{figure}[h]
\begin{center}
\includegraphics{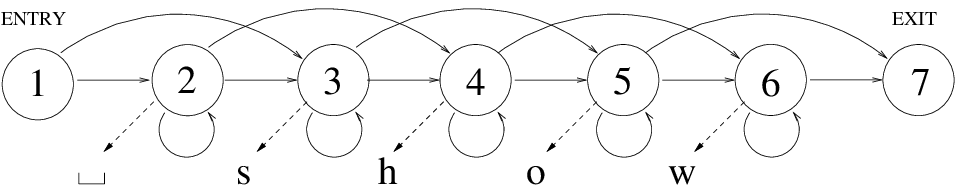}
\caption{The structure of $\mathcal{M}_{\underline{\ }\textit{show}}$
  with initial space-state\label{fig:m-show-space}}
\end{center}
\end{figure}
The HMMs of the OD are then simply trained accordingly (see
Chapter~\ref{ch:ae} below). The HMM $\mathcal{M}_{\underline{\
}\textit{show}}$, for example, will score maximum probability for the
character sequence \xpl{\symbol{32}show}. Using the approach depicted
in Figure~\ref{fig:m-show-space}, the
utterance~(\ref{eq:ctr-segmentation1}) would be segmented as
\begin{list}{{\tt ==>}}{\setlength{\rightmargin}{.5cm}}
        \inputitem{utterance}\underline{show}\,\underline{\ me}\,\underline{\ all}\,\underline{\ cars}\label{eq:ctr-segmentation2}
\end{list}
The fact that the first word's initial space character is not in the
input does not present a problem. The two types of segmentation errors
are dealt with in the following way. By taking the skip transition
past the space-state, run-ons can be handled. Space characters can
(if the network is so trained) be emitted at any state, thus splits
can be modeled as well.

The Linguistic Decoder provides the pattern matching Orthographic
Decoder with context and supervises the passing of tokens between word
models of the OD. From a computational view this can be accomplished
by connecting the set of word model HMMs together in a large super-HMM
where the connections between subnetworks determine what words can
follow others and with what probability. This solution can actually
be hard-wired into a system, but the system will be awfully rigid. The
Token Passing framework provides a much more flexible approach. The
forwarding of tokens from the exit state of one OD HMM to the entry
state of another is elevated to a higher level of control, the
Linguistic Decoder. The language model encoded by the LD may of course
vary, but it is reasonable to use a probabilistic language model since
the OD operates on probabilities. This facilitates for a clean LD-OD
interface, the LD and the OD use a common communication protocol. The
Connected Text Recognition experiments reported here employ LDs that
encode language models that can be expressed with HMM networks. This
means that the Viterbi algorithm can also be used in the LD. In the
following we will thus assume a single HMM network in the Linguistic
Decoder.

The Linguistic Decoder HMM network assigns probabilities to word
sequences. The states of the network represent different
contexts. (What is meant by a context of course varies depending on
the language model.) Specific word occurrences are more or less likely
to occur in a given context. The transition distribution of the
HMM is thus the probability of going from one context to another, and the
observation symbol distribution is the probability of a word given the
context. The Viterbi algorithm can be used to compute the probability
of the word sequence $W$ and the optimal context sequence
$\text{\textit{CON}}^{*}$ given the LD HMM
$\mathcal{M}_{\text{\textit{LD}}}$,
i.e.
$\prob{W,\text{\textit{CON}}^{*}|\mathcal{M}_{\text{\textit{LD}}}}$.
Note that $\mathcal{M}_{\text{\textit{LD}}}$ is the language model
$\mathcal{G}$ introduced in equation~(\ref{eq:ctr-word-seq-prob}), and
that
$\prob{W,\text{\textit{CON}}^{*}|\mathcal{M}_{\text{\textit{LD}}}}$ is
used as an approximation $(\approx)$ for
$\prob{W|\mathcal{M}_{\text{\textit{LD}}}}$ in
equation~(\ref{eq:ctr-prob-spec})\footnote{Remember from
  Section~\ref{sec:iwr} that $\prob{c_{t_{i}}^{t_{i}'},Q^{*}|w_{i}}$ is
  already used to approximate $\prob{c_{t_{i}}^{t_{i}'}|w_{i}}$.}.
\begin{equation}
\prob{W,\text{\textit{CON}}^{*}|\mathcal{M}_{\text{\textit{LD}}}} =
\max_{\text{\textit{CON}}} \left[ \prod_{i=1}^{n}
    \prob{\text{\textit{con}}_{i}|\text{\textit{con}}_{i-1}}
    \prob{w_{i}|\text{\textit{con}}_{i}} \right]
\label{eq:ctr-ld-context}
\end{equation}
where $\text{\textit{con}}_{0}$ is a `dummy-context', or, in the present computational
model, the entry state of $\mathcal{M}_{\text{\textit{LD}}}$.

The quantity $\prob{w_{i}|\text{\textit{con}}_{i}}$ in
equation~(\ref{eq:ctr-ld-context}) can be thought of as the interface
between the two knowledge sources, the coupling between the layers
that enables the recognizer to produce the word sequence that is
overall most likely, with respect to both orthographic evidence and
linguistic expectation. When the LD predicts that the word $w_{i}$ is
present in the input, the OD HMM $\mathcal{M}_{w_{i}}$ begins to
evaluate that hypothesis. Equation~(\ref{eq:ctr-prob}) can be made
more specific:
\begin{equation}
\prob{W^{*}|C} \approx
\max_{\substack{W,\text{\textit{CON}}\\ 1\leq t_{i},t_{i}'\leq T\\ t_{i}<t_{i}'}} \left[
  \prod_{i=1}^{n} \underbrace{\prob{c_{t_{i}}^{t_{i}'}|w_{i}}}_{\text{\textit{OD}}}
  \underbrace{\prob{\text{\textit{con}}_{i}|\text{\textit{con}}_{i-1}}
    \prob{w_{i}|\text{\textit{con}}_{i}}}_{\text{\textit{LD}}} \right]
\label{eq:ctr-prob-spec}
\end{equation}
The equation above can be visualized in the Token Passing
framework (see Figure~\ref{fig:token-propagate} in
Section~\ref{sec:tp}).
\begin{figure}[ht]
\begin{center}
\includegraphics{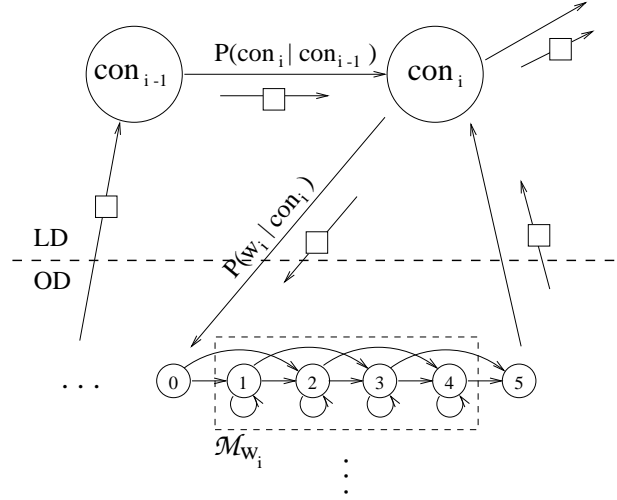}
\caption{The LD-OD interface\label{fig:ldod-interact}}
\end{center}
\end{figure}
Tokens are passed within the LD HMM according to the transition
distribution of the network and word models of the OD are hypothesized
according to the observation distribution. As tokens reach the exit
state of the word model, they are propagated back to the context/state
in which they were originally proposed. This upward propagation does
not constitute an extra cost. Note that whereas the workings of the OD
HMMs is time-synchronized the LD is not. One step of the
\emph{step\_model} procedure is executed in the OD HMMs for each
character that is read from the input. The LD simply reacts to tokens
that are passed up from beneath, and this has nothing to do with the
the time index $t$. This means for example (see
Figure~\ref{fig:ldod-interact}) that a token in the exit state of some
word model at time $t$ can get passed up to
$\text{\textit{con}}_{i-1}$, then passed from there to
$\text{\textit{con}}_{i}$, from $\text{\textit{con}}_{i}$ to the entry
state of $\mathcal{M}_{w_{i}}$ and time is still $t$. When the OD
starts processing the hypothesis $w_{i}$, and characters are read from
the input, the time index is, of course, incremented.

The Viterbi algorithm is used to compute the probability of an
observation sequence and the optimal state sequence. The state
sequence is not of great interest when computing the probability of a
character sequence given a word model
$\prob{c_{t_{i}}^{t_{i}},Q^{*}|\mathcal{M}_{w_{i}}}$, the quantity computed
by the OD HMM networks. The states of an OD network represent
character positions and this information is not really useful, the
quantity is used more like an approximation for
$\prob{c_{t_{i}}^{t_{i}}|\mathcal{M}_{w_{i}}}$. The same reasoning can be
applied to the LD (see Section~\ref{sec:tp}), but whether the context
sequence is important or not depends on what a context represents in
the language model encoded by the LD network. Irrespective of whether
the context sequence is informative or not, the word sequence
certainly is. Retrieving the word sequence is our main
objective. Since each token arriving in an LD network state from below
represents a word matched with the input, this event has to be recorded
so that backtracking is enabled. The data structure used to record the
word hypothesis is the \emph{Word Link Record}
(see Section~\ref{sec:tp}).

\begin{figure}[h]
\begin{center}
\includegraphics{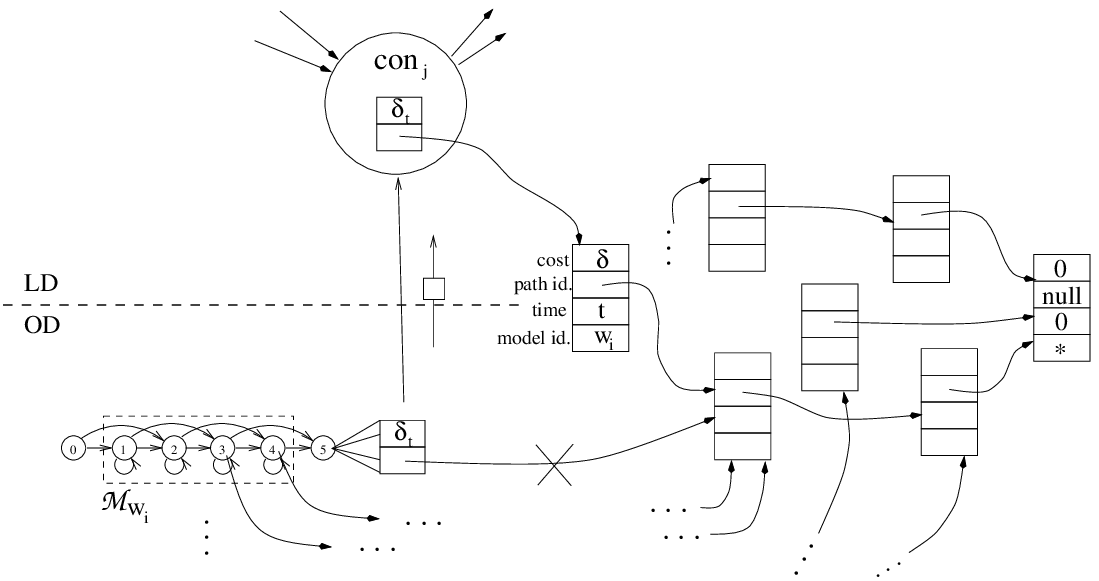}
\caption{A Word Link Record is created as a token is passed upwards\label{fig:ldod-wlr}}
\end{center}
\end{figure}

The Word Link Records (WLR) are stored in a linked list structure
where each path through the structure represents a word sequence
hypothesis\footnote{The LD layer uses the linked list structure to
keep track of what is going on in the layer beneath. In the general
case, where there might be several LD layers, each layer would need
its own WLR list structure}. Each token, besides the cost, keeps a
path identifier (pointer) to the last WLR in the word sequence
hypothesis. That WLR has a pointer to its predecessor and so on. The
token itself represents the head of the path. The WLR is created as a
token in the exit state of an OD network is propagated back to the LD
state/context that hypothesized the word, and the new WLR is
incorporated into the list structure. The WLR contains the \emph{cost}
of the path (up to the point where it was created), the \emph{path
identifier} (inherited from the token), the \emph{time index}
(character position) at which it was created and the \emph{model
identifier} of the OD HMM that the token exited from\footnote{It is
possible to store additional information in the WLRs, the context in
which the word occurrence was hypothesized, for example, might be
useful in subsequent processing of the text.}. The process is
visualized in Figure~\ref{fig:ldod-wlr}.

The `start-WLR' with * as model identifier is the root of the linked list
structure. At the start of the recognition process the start token in
the LD points to this WLR. Note that there might be several tokens
pointing to the same WLR, along with other WLRs. The creation of new
WLRs is called \emph{record\_decisions},
Algorithm~\ref{alg:record-decisions}, and it is of course crucial in
Connected Text Recognition. The four fields of the WLR in
Algorithm~\ref{alg:record-decisions} are denoted: $\delta$,
$\uparrow_{\text{\textit{wlr}}}$, $\text{\textit{time}}$ and
$\text{\textit{word}}$. The cost and path identifier fields of the
token are denoted $\delta_{t}$ (as usual) and
$\uparrow_{\text{\textit{wlr}}}$.

\begin{algorithm}
\caption{\emph{record\_decisions}}
\begin{algorithmic}[1]
\FORALL{states $i<N$ \COMMENT{of the Linguistic Decoder}}
\IF{$i$ holds a token $\tau_{i}$}
\STATE{create a new WLR $\text{\textit{wlr}}$}
\STATE{\textbf{with} $\text{\textit{wlr}}$ \textbf{do}}
\STATE{\ \ \ $\text{\textit{wlr}}(\delta) = \tau_{i}(\delta_{t})$}
\STATE{\ \ \ $\text{\textit{wlr}}(\uparrow_{\text{\textit{wlr}}}) =
  \tau_{i}(\uparrow_{\text{\textit{wlr}}})$}
\STATE{\ \ \ $\text{\textit{wlr}}(\text{\textit{time}}) = t$}
\STATE{\ \ \ $\text{\textit{wlr}}(\text{\textit{word}}) = w_{k}$
  \COMMENT{$\mathcal{M}_{w_{k}}$ is the OD HMM that propagated $\tau_{i}$}}
\STATE{\textbf{end with}}
\STATE{$\tau_{i}(\uparrow_{\text{\textit{wlr}}}) = \text{\textit{wlr}}$}
\ENDIF
\ENDFOR
\end{algorithmic}
\label{alg:record-decisions}
\end{algorithm}

The reader should not be bothered by the fact that the algorithms here
and elsewhere in this thesis do not make sense down to the last
detail. For example, how can a token know which state of the LD that
hypothesized it so that it can get propagated back up to that state
once it reaches the exit state of the OD network? There is of course a
simple technical solution to this sort of problem and issues of this
type are generally suppressed in the algorithmic outlines. The purpose
of the algorithms is to convey the basic idea.

The Token Passing algorithm is now set to recognize word sequences
instead of isolated words. The LD HMM used in the superficial
algorithmic presentation below is similar to the OD HMM in
Figure~\ref{fig:m-show-space} except that it is not limited to
left-to-right transitions (and of course, the observation symbols are
not characters but words). The Viterbi algorithm is used both in the
LD and the OD. The Beam Search heuristic has a slightly different
effect in Connected Text Recognition compared to Isolated Word
Recognition. If a word model gets \emph{deactivated} in IWR it stays
deactivated for the duration of the recognition process, in CTR a word
model may be \emph{reactivated} at any time (cf. line~10 in
Algorithm~\ref{alg:ctr}). The \emph{step\_model} procedure,
Algorithm~\ref{alg:iwr-step_model}, is reused here with only minor
changes. The token put in the entry state of the OD HMM does not have
zero cost since it has been subject to prior cost
accumulation. Recall: the \emph{start token} has cost $-\log 1 = 0$
and the \emph{null token} has cost $-\log 0 = \infty$

\begin{algorithm}
\caption{Connected Text Recognition with Beam Search}
\begin{algorithmic}[1]
\FOR{$t=0$}
\STATE{Create the start-WLR}
\STATE{Put start token in the entry state of the LD\\and let it point
  to the start-WLR}
\STATE{Put null tokens in all other states of the LD}
\STATE{Deactivate all models of the OD}
\STATE{$\mathcal{M}^{*}\!\!\rightarrow\!\!\tau^{*}(\delta_{t}) = \infty$}
\ENDFOR
\FOR{$t=1$ to $T$}
\FORALL{states $i<N$ in $\mathcal{M}_{LD}$ with a non-null token}
\STATE{Pass a copy of the token $\tau_{i}$ to the entry state of all
  $\mathcal{M}_{w_{k}}$ observable in state $j$:\\
  $\mathcal{M}_{w_{k}}\!\!\rightarrow\!\!\tau_{1}(\delta_{t}) =
  \tau_{i}(\delta_{t}) + (-\log a_{ij}) + (-\log b_{j}(w_{k}))$
  \COMMENT{Reactivation}}
\ENDFOR
\STATE{Put null tokens in all states in $\mathcal{M}_{LD}$}
\FORALL{models $\mathcal{M}_{w_{k}} 1\leq k\leq M$}
\IF{$\mathcal{M}_{w_{k}}$ is active}
\IF{(\ref{eq:iwr-beam})}
\STATE{Deactivate $\mathcal{M}_{w_{k}}$}
\ELSE
\STATE{\emph{step\_model}$(c_{t})$ with $\mathcal{M}_{w_{k}}$}
\ENDIF
\ENDIF
\ENDFOR
\STATE{compute
  $\mathcal{M}^{*}\!\!\rightarrow\!\!\tau^{*}(\delta_{t})$ according
  to~(\ref{eq:iwr-globalopt})}
\IF{the token in the exit state of $\mathcal{M}_{w_{k}}$ is non-null}
\STATE{Propagate the token up to the LD state that hypothesized it}
\ENDIF
\FORALL{states $i<N$ in $\mathcal{M}_{LD}$}
\STATE{Find the token with min \emph{cost} and discard the rest}
\ENDFOR
\STATE{\emph{record\_decisions}}
\FORALL{states $i<N$ in $\mathcal{M}_{LD}$ connected to state $N$}
\STATE{Pass a copy of the token $\tau_{i}$ to state $N$:\\
  $\tau_{N}(\delta_{t}) = \tau_{i}(\delta_{t}) + (-\log a_{iN})$}
\ENDFOR
\STATE{In state $N$ of $\mathcal{M}_{LD}$: Find the token with min
  \emph{cost} and discard the rest}
\ENDFOR
\STATE{Backtrack
  $\mathcal{M}_{LD}\!\!\rightarrow\!\!\tau_{N}(\uparrow_{\text{\textit{wlr}}})$}
\end{algorithmic}
\label{alg:ctr}
\end{algorithm}

At time $T$ the most likely word sequence can be established by
following the path identifier chain of the token in the exit state
($\mathcal{M}_{\text{\textit{LD}}}\!\!\rightarrow\!\!\tau_{N}$) back
to the start-WLR, which indicates the start of the sequence.

The algorithm is quite easily generalized to perform $N$-best
search. It suffices to let each state hold $N$ tokens instead of just
one. Lines~27 and~33 of Algorithm~\ref{alg:ctr} should be changed to:
`Find the $N$ tokens with min \emph{cost} and discard the rest'.

Algorithm~\ref{alg:ctr} returns the best word sequence, i.e. it
performs a 1-best search. Under certain circumstances, however,
Algorithm~\ref{alg:ctr} is suboptimal in the sense that it can not
guarantee that the overall best word sequence is returned. If the
algorithm is run in 1-best mode, i.e. each state of the OD HMMs can
hold only one token, and the same word model can be hypothesized from
more than one state, it can happen that the token that would score the
overall lowest cost if it was allowed to proceed, gets pruned because
there is another token with a lower cost for the left context that is
preferred in the entry state of the OD HMM\@. The part-of-speech
bigram language model is an example of a language model with this
property (the words are ambiguous with respect to their
part-of-speech). Young~\emph{et al.}~\shortcite{young+al:89} get
around this problem by having multiple instances of the same word
model, one instance for each context in which it is observable. A
variation of the same scheme is used here. Instead of having multiple
network instances, each state can hold more than one token,
i.e. $N$-best search is used to guarantee that the overall best
sequence is obtained. (See the following chapter for further comments
on this subject.)

Note that the word sequence returned is really a sequence of model
identifiers. Supposedly the model identifier of a model is exactly the
character sequence that the model models, e.g. the model identifier of
$\mathcal{M}_{\underline{\ }\textit{show}}$ is \xpl{\symbol{32}show},
but that need not necessarily be the case. For some word-forms it is
unrealistic to have one HMM network for each word-form (character
sequence). The natural numbers is one such `word group'. A solution to
this problem would be to have a single network that recognizes a
number of word-forms, e.g. dates, social security numbers and possibly
even proper nouns. The network $\mathcal{M}_{<date>}$ with model
identifier \xpl{<date>} would then recognize character sequences like
\xpl{4 of July}, and the input \xpl{On the 4 July we went to New York}
could come out as \xpl{On the <date> we went to <city>}. Since the
WLRs contain the word boundary positions, it is possible to extract
the character sequence recognized as \xpl{<date>} from the character
input stream.\label{par:word-group}

The approach to Connected Text Recognition presented here has the
potential to deal with all the lexical errors discussed in this
thesis. It has the potential to handle misspellings, run-ons and
splits, single and multiple character errors and nonword and real-word
errors. Having the potential to solve a problem is, however, not the
same thing as actually solving it. This of course ultimately depends
on the accuracy with which the Orthographic Decoder models the noisy
channel and the Linguistic Decoder the language. The performance of
this approach has to be experimentally evaluated.

\chapter{Experimental Evaluation}
\label{ch:ae}
To test our ideas of the layered HMM approach in the Token Passing
framework we have developed a system, \prognm{ctr}, to perform Connected
Text Recognition. The system is based on the algorithms presented in
the previous chapter and is thus also restricted to two layers. The
Linguistic Decoder is represented in one (the topmost) layer by a
single HMM network and the Orthographic Decoder is realized by the set
of word modeling HMM networks in the bottom layer. The Beam Search
threshold defines the level of trade-off between accuracy and
computational efficiency (speed). In the experiments reported below we
have opted for accuracy at the expense of speed. We have used an
infinitely wide Beam so that no hypotheses are ever
pruned.

Much of the work presented in this thesis is based on the findings in
the dialogue corpus profiled in Chapter~\ref{ch:corpus}. The dialogue
corpus application was also the focal point during the development of
the techniques. The \textsc{cars} part of the dialogue corpus was the
first error corpus that was given to \prognm{ctr} for testing
(Section~\ref{sec:ee-cars}). Later we came to the conclusion that we
needed to test on a larger corpus as well, and this materialized in
the \textsc{secretary} experiment (Section~\ref{sec:ee-secretary}). The
\textsc{secretary} experiment is a completely different application
compared to the dialogue scenario, it concerns transcription typing of
a software manual.

\section{\sc cars}
\label{sec:ee-cars}
\setlength{\extrarowheight}{2pt}
The \prognm{ctr} experiments reported here concern the
\textsc{cars} corpus. The corpus includes 20 dialogues. It contains in
all 369 utterances, 3,139 word tokens and 584 word types. There are 92
lexical errors distributed over 71 utterances. There are 62
misspellings, 17 run-ons and 13 splits\footnote{The figures on
  error frequencies and results presented in this thesis are not in
  complete agreement with figures presented in previous
  publications~\cite{ingels:96a,ingels:96b}. The reason is that
  different definitions have been used for multiple segmentation
  errors. A string like \xpl{toyotapeugeotvolkswagen} is treated as
  \emph{one} multiple run-on here, whereas in the other publications
  this string was regarded as two run-ons.}.

The intention with the \textsc{cars} experiment is first and foremost
to get a handle on the overall error correcting performance of the
\prognm{ctr} system. We are also interested in seeing what impact
different language models will have on the system. With sparse data,
as in the present case, it is not absolutely certain that a Linguistic
Decoder implementing a language model will have a strong positive
effect. Experiments have been conducted on three different, rather weak,
language models, a Unigram language model and two tag Bigram language
models. The difference between the two tag Bigram language models is
that they use different tag-sets. One uses a small domain-oriented
tag-set while the other employs a Part-Of-Speech (POS) tag-set. From
the point of view of the application it is of interest to note any
differences between the application-close domain oriented tag-set and the
linguistically oriented tag-set. We also have a Baseline to which
the results of these experiments can be compared. The Baseline
experiment involves no linguistic constraints so the correction of
lexical errors is performed by the Orthographic Decoder alone.

The 20 dialogues were randomly divided into five parts of four
dialogues each. In the experiments, 16 dialogues (four parts) were
used to obtain the language model and then the model was tested on the
remaining four dialogues (one part). The partitionings were rotated so
that each language model was tested on all of the five parts. The same
Orthographic Decoder was used in all the experiments.

\subsection{The Orthographic Decoder}
\label{subsec:od}
The Orthographic Decoder contains 584 word modeling HMMs, one for
each word type in the corpus. The structure of the OD HMMs can be seen
in figure~\ref{fig:m-show-space}. Ideally each HHM should be trained
on typical errors occurring in Swedish text. Unfortunately there is no
such error corpus available and we can certainly not train the HMMs on
the errors occurring in the corpus. We must find a way to generate an
error corpus so that the OD HMMs can be trained and used for other
purposes as well, not just to identify the particular errors in this
corpus.

Inspired by the basic error types one can construct four error
generating functions. Given a word, these functions will produce a set
of corrupted forms of the input word. Given the word
\xpl{\symbol{32}show} for example, the deletion function will produce:
\{\xpl{show}, \xpl{\symbol{32}how}, \xpl{\symbol{32}sow},
\xpl{\symbol{32}shw}, \xpl{\symbol{32}sho}\}. The deletion operator is
applied to each character position in the word. Although these error
types apply to the space character as well as to any other character,
we have an extra error type dealing only with the space character,
which is called \emph{white-space insertion}. There is also an error
type called \emph{double stroke}. The error types insertion and
substitution raise the question as to what to insert and what to
substitute for, respectively. One hypothesis is that keyboard
neighbors are likely to take part in, for example, substitutions. The
neighbors\footnote{The neighbor relation is limited to immediate left
and right neighbors.} of \xpl{o} are \xpl{i} and \xpl{p}, so if
substitutions are applied, the error corpus for
$\mathcal{M}_{\underline{\ }\textit{show}}$ will contain amongst
others \xpl{\symbol{32}shiw} and \xpl{\symbol{32}shpw}. The list of
error-generating operators that have been considered for training of
the OD HMMs is thus:
\begin{itemize}
\item deletion (e.g. \xpl{\symbol{32}shw})
\item insertion (e.g. \xpl{\symbol{32}shpow})
\item substitution (e.g. \xpl{\symbol{32}shpw})
\item transposition (e.g. \xpl{\symbol{32}sohw})
\item white-space insertion (e.g. \xpl{\symbol{32}sh\symbol{32}ow})
\item double stroke (e.g. \xpl{\symbol{32}shoow})
\end{itemize}
Note that the basic error-generating functions above will produce
mostly performance related errors, i.e. they are likely to be
generated by a human only by mistake. The error corpora generated for
the various OD HMMs are consequently quite poor with respect to
knowledge of spelling errors people produce for other
reasons. Phonetic resemblance and other cognitively related
difficulties are not included in the corpora. The only `external'
knowledge included in the training corpora is the layout of the
keyboard.

It is, of course, often the case that a corruption generated by one of
the above error functions turns out as another legal word in the
vocabulary. If deletion is applied to \xpl{them} for example,
\xpl{the} will be part of \xpl{them}'s error corpus. A simple little
program \prognm{filter} was devised to remove all such `real-word'
corruptions. If the corpora are filtered, the effect will be that
real-word errors are harder to recognize, but on the other
hand \prognm{ctr} will be less likely to change words that are
properly spelled.

The word models of the OD have been trained on their respective error
corpora with the Baum-Welch reestimation algorithm. A maximum
likelihood estimation procedure like the Baum-Welch algorithm will
assign zero probabilities to all unseen events. Since unexpected
events such as misspellings not included in the corpus will most
likely appear, the parameters of the model must be smoothed. Smoothing
is the term generally used for making the distributions more
uniform. Very low probabilities are adjusted upwards and high
probabilities are adjusted downwards. The models used in the
experiments reported here have all been smoothed with one of the
simplest smoothing schemes, the \emph{additive smoothing
scheme}~\cite{levinson+al:83} (p. 1053). After the model has been
trained, a small number $\epsilon > 0$ is assigned to all parameters
corresponding to unseen events and the other parameters are adjusted
downwards accordingly.

\begin{figure}[ht]
\begin{center}
\includegraphics{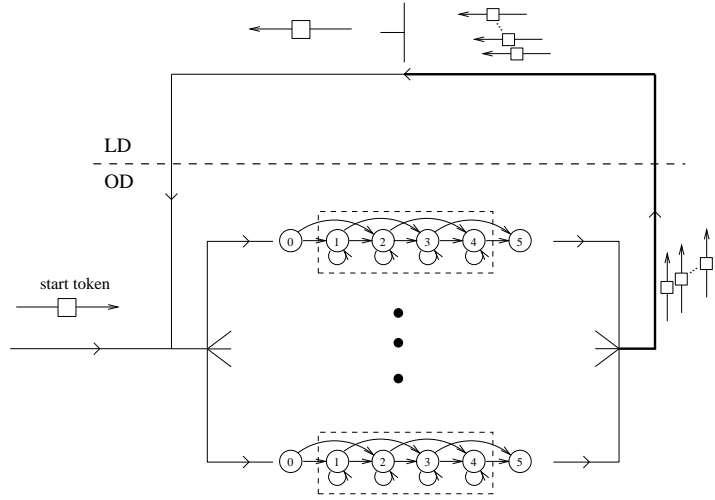}
\caption{The Baseline experiment -- \prognm{ctr} setup\label{fig:ldod-naive}}
\end{center}
\end{figure}

Because of the fact that no Swedish corpus of actually occurring
spelling and segmentation errors is available, the process of finding
a reasonable training setup for the OD HMMs is very much a matter of
trial and error. Prior to the first experiments on \textsc{cars} the
Baseline system configuration (see Figure~\ref{fig:ldod-naive}) was
used to try out different training corpora and smoothing parameters
for the OD HMMs. The Baseline system configuration runs without a
Linguistic Decoder. At each time $t$ the best token, out of the $M$
possible tokens that can be propagated out of the exit states of the
$M$ OD HMMs, corresponds to the best word hypothesis at time $t$. This
token generates a WLR corresponding to the word hypothesis and copies
of the token are inserted into the entry state of all $M$ word models,
i.e. the tokens just circulate through the LD without any cost being
added. 

In the experiments reported in Section~\ref{subsec:res}, the error
corpora were generated with the error functions deletion,
substitution, and white-space insertion. Apart from this general
strategy, some special words need specialized corpora. These words
include single character `words' such as \xpl{,}, \xpl{?}, \xpl{.} and
so on. There are seven such words in the vocabulary and the training
material is a small set of hand-made corruptions that only involve the
space character. There are 51 numbers in the vocabulary. There are car
prices, figures for fuel consumption, grades and so on. Although there
are more clever ways to handle things like this
(cf. Section~\ref{sec:ctr}~page~\pageref{par:word-group}), the numbers
all have their own individual HMM in the OD modeling it. These special
words have corpora generated with only the white-space insertion error
function. The error corpora thus generated were then filtered for
real-word errors. The OD HMMs were trained with the Baum-Welch
reestimation algorithm. After training, each HMM had its observation
symbol distribution smoothed with the additive smoothing scheme with
$\epsilon_{obs}=10^{-4}$.

Note that we are evading the unknown word problem. Even if a word type
is unseen in the training corpus of an experiment, the OD will still
contain the model corresponding to the unseen word.

\subsection{The Unigram Language Model}
\label{subsec:unigram}
The Unigram language model:
\begin{equation}
\prob{w_{1},\ldots ,w_{T}} = \prod_{i=1}^{T}\prob{w_{i}}
\label{eq:unipar}
\end{equation}
The language model's parameters are extracted from the training corpus
of the five partitionings.
\begin{equation*}
\prob{w_{i}} = \frac{Count(w_{i})}{N}
\end{equation*}
where $N$ is the number of word tokens in the training corpus. For
each of the five partitionings there will be a fair amount of unknown
words in the test corpus that have to be smoothed. Recall that one
fifth of the entire corpus is held out for testing in each
partition. It should be noted that the observables of the Linguistic
Decoder that are smoothed are exactly the words that are unseen in the
training material but are members of the vocabulary, i.e. word models
with non-zero probability after smoothing are the words of the
vocabulary and no other.

The Linguistic Decoder realizing the Unigram model is a single state
HMM (three states including the entry and exit states). The parameters
of the Unigram make up the observation symbol distribution of the LD
HMM.

The results on the Unigram Linguistic Decoder reported in
Section~\ref{subsec:res} refer to the combined results from the five
experiments with the five partitionings where the LD have been
smoothed with the additive smoothing scheme with
$\epsilon_{obs}=10^{-4}$.

\subsection{The Domain-Tag Bigram Language Model}
\label{subsec:biclassdom}
In the domain-tag Bigram language model there are 19 tags. The words
of the corpus are grouped into classes that are semantically- or
domain-oriented. Examples of classes and class members are:
\begin{itemize}
\item Object Head (OH), e.g. \xpl{\symbol{32}saab\symbol{32}900},
  \xpl{\symbol{32}all}
\item Aspect Head (AH), e.g. \xpl{\symbol{32}costs},
  \xpl{\symbol{32}acceleration}
\item Communicative Head (CH), e.g. \xpl{\symbol{32}show},
  \xpl{\symbol{32}example}
\end{itemize}
The complete tag-set is listed in Appendix~\ref{app:A-dom}.

If $\textit{tag}_{1}^{T+1}$ denotes a sequence of $T$ tags assigned to a
sequence of $T$ words, (plus the dummy tag $\textit{tag}_{T+1}$ corresponding
to the nonexistent word $w_{T+1}$), the tag Bigram language model
looks like:
\begin{equation}
\prob{w_{1},\ldots ,w_{T}} = \sum_{all
  \textit{tag}_{1}^{T+1}}\prod_{i=1}^{T}\prob{w_{i}|\textit{tag}_{i}}\prob{\textit{tag}_{i+1}|\textit{tag}_{i}}
\label{eq:bigrampar}
\end{equation}
The language model's parameters are extracted from the tagged training
corpus of the five partitionings:
\[
\prob{\textit{tag}_{i+1}|\textit{tag}_{i}} =
\frac{Count(\textit{tag}_{i},\textit{tag}_{i+1})}{Count(\textit{tag}_{i})}
\]
\[
\prob{w_{i}|\textit{tag}_{i}} =
\frac{Count(\textit{tag}_{i},w_{i})}{Count(\textit{tag}_{i})}
\]

The tag Bigram can be straightforwardly implemented as our Linguistic
Decoder HMM, see for
example~Cutting~\emph{et al.}~\shortcite{cutting+al:92}. The
second factor on the right-hand side of equation~(\ref{eq:bigrampar})
is the transition distribution of the LD and the first factor is the
observation distribution. The observables of the LD HMM are the words
of the vocabulary, or in other words, the word modeling HMMs of the
OD\@. The \prognm{ctr} setup is shown in
Figure~\ref{fig:ldod-interact} where the $\textit{contexts}$
correspond to the tags of the language model.

The tag Bigram language model models local contextual
dependencies. These dependencies are weak. There is a great deal of
uncertainty as to what the next word might be, judging from the tag
assigned to the present word. The uncertainty is emphasized by the
relatively small tag-set, each class has a relatively large amount of
potential realizations. Still, the test corpus will contain both unseen
tag-to-tag transitions and, of course, previously unseen words. This
means that both the transition distribution and the observation
distribution of the LD have to be smoothed. In the case of the
Unigram, the single state of the LD has the entire vocabulary as
observables. The states of the tag Bigram LD each have a subset of the
vocabulary as possible observables. When the observation distribution
of the tag Bigram LD is smoothed, it is only the previously unseen
observables of this subset that are assigned a non-zero probability,
the remainder stays zero.

It should be noted that we are not using the Baum-Welch reestimation
algorithm here. \prognm{ctr} can, if so instructed, return the tag
sequence along with the normalized utterance just like an ordinary POS
tagger. It was discussed above that whether or not this is desirable
depends on what the $\textit{context}$ encodes. Especially with the
domain oriented tags it can be useful to have the utterance tagged
since it will reduce the interpretation step, from input query to
SQL-query, quite substantially. For the purpose of tagging, the
language model extracted from a tagged corpus will outperform the
language model induced from an untagged corpus with a maximum
likelihood estimation procedure~\cite{elworthy:anlp:94}. For the
purpose of predicting the next word however, this need not be the
case. In the \textsc{cars} experiment we utilize the tagged corpus,
whereas in the \textsc{secretary} experiments presented in the
following section we will contrast the two approaches.

The results on the domain-tag Bigram Linguistic Decoder reported in
Section~\ref{subsec:res} refer to the combined results from the five
experiments with the five partitionings where the LD have been
smoothed with the additive smoothing scheme with
$\epsilon_{trans}=10^{-3}$ and $\epsilon_{obs}=10^{-3}$. The most
ambiguous word in the language model is three ways ambiguous, so
\prognm{ctr} was run under 3-best search to guarantee optimal
performance.

\subsection{The POS Bigram Language Model}
\label{subsec:biclasssuc}
The POS Bigram language model has 31 tags. The tag-set
originates from the SUC corpus (Stockholm-Ume\aa\ Corpus
\cite{kallgren:col:90}). The tags used in the SUC corpus are
traditional Part-Of-Speech with associated morphological features. We
have made slight modifications to the original set of SUC-tags to
obtain a set of atomic tags with different syntactic
distributions. Examples of tags and tag members are:
\begin{itemize}
\item Proper noun (PM), e.g. \xpl{\symbol{32}saab\symbol{32}900}
\item Determiner (DT), e.g. \xpl{\symbol{32}all}
\item Verb form finite (VBF), e.g. \xpl{\symbol{32}costs}
\item Noun (NN), e.g. \xpl{\symbol{32}acceleration},
  \xpl{\symbol{32}example}
\item Verb form imperative (VBP), e.g. \xpl{\symbol{32}show}
\end{itemize}
The complete tag-set is listed in Appendix~\ref{app:A-pos}.

The POS Bigram parameters are extracted from the tagged training
corpus in the same way as was done with domain-tag Bigram. Also with
POS Bigram both the state transition distribution and the observation
symbol distribution are smoothed with the additive smoothing
scheme.

The results on the POS Bigram Linguistic Decoder reported in
Section~\ref{subsec:res} refer to the combined results from the five
experiments with the five partitionings where the LD have been
smoothed with the additive smoothing scheme with
$\epsilon_{trans}=10^{-3}$ and $\epsilon_{obs}=10^{-3}$. The most
ambiguous word in the language model is three ways ambiguous, so
\prognm{ctr} was run under 3-best search to guarantee optimal
performance.

\subsection{Results}
\label{subsec:res}
When an experiment is conducted, \prognm{ctr} is run on the corpus in
batch mode, i.e. utterances are processed from an input file and
output to an output file. This creates pairs of utterances. Thus,
resulting from an experiment is a set of pairs: \mbox{$\langle${\tt
original utterance}\ ,\ {\tt normalized utterance}$\rangle$}. An
experiment is evaluated by comparing the pairs resulting from the
experiment to pairs in a result \emph{key}. The key is a hand-made set
of pairs where the first element (the original utterance) contains at
least one lexical error and the second element is the appropriate
correction of that utterance. This set is called $\mathbf{A}$. The
\emph{outcome} of an experiment are the pairs produced in the
experiment where the second element is \emph{not} identical to the
first one, and the pairs where the first element is identical to the
first element in one of the pairs in $\mathbf{A}$. In other words: The
\emph{outcome} of an experiment are the pairs produced in the
experiment where \prognm{ctr} has changed the input, and the pairs
where it should have changed the input. This set is called
$\mathbf{C}$. The pairs in the outcome that are also in the key belong
to the set $\mathbf{B}$, i.e. $\mathbf{B} = \mathbf{A}\cap
\mathbf{C}$. The outcome of an experiment can now be rated with
respect to the performance measures \emph{recall} and
\emph{precision}.

\begin{eqnarray*}
recall & = & \frac{\mid\mathbf{B}\mid}{\mid\mathbf{A}\mid}\times
100\\\\ precision & = &
\frac{\mid\mathbf{B}\mid}{\mid\mathbf{C}\mid}\times 100
\end{eqnarray*}

There is one important point that needs to be emphasized regarding
this style of evaluation. Since the outcome of an experiment does not
only include the utterances that have actually been changed, but also
those that should have been changed, this means that $\mathbf{C}$ will
always contain \emph{at least} as many pairs as $\mathbf{A}$. The
reason for this somewhat unusual evaluation scheme is of course that
we want to capture the performance of the system regarding real-word
errors. The implications of this evaluation style is that precision
can never be higher than recall. If the two metrics are the same, this
means that no unfounded changes have been made to the input. 

An example of a pair in $\mathbf{A}$: $\langle$\dxpl{rust
\hgl{protetion} \hgl{forthese}}\ ,\ \dxpl{rust protection for
these}$\rangle$. The first element of the pair contains two errors and
we would like to extend the performance measure to account for individual
errors, not just whole utterances. From the outcome of the experiment
we can extract the counterparts for $\mathbf{A}$, $\mathbf{B}$ and
$\mathbf{C}$ that apply to the respective error categories. We have
$\mathbf{A}^{m}$, $\mathbf{B}^{m}$ and $\mathbf{C}^{m}$ for
misspellings, $\mathbf{A}^{r}$, $\mathbf{B}^{r}$ and $\mathbf{C}^{r}$
for run-ons and we have $\mathbf{A}^{s}$, $\mathbf{B}^{s}$ and
$\mathbf{C}^{s}$ for splits. We are also interested in the total
number of individual errors so the key
$\mathbf{A}^{tot}=\mathbf{A}^{m}\cup\mathbf{A}^{r}\cup\mathbf{A}^{s}$
is added to the list of keys. The example pair above that was a member
of $\mathbf{A}$ also adds $\langle$\dxpl{\hgl{protetion}}\ ,\
\dxpl{protection}$\rangle$ to $\mathbf{A}^{m}$ and $\mathbf{A}^{tot}$
and $\langle$\dxpl{\hgl{forthese}}\ ,\ \dxpl{for these}$\rangle$ adds
to $\mathbf{A}^{r}$ and $\mathbf{A}^{tot}$. The five keys provide the
five performance categories in the tables below.

\begin{table}[htb]
\centerline{
\begin{tabular}{||c|l|c|c||}
\hline
Experiment            & Performance categories    & Recall        & Precision     \\ \cline{1-4}
                      & Utterances                & 72  \%        & 72  \%        \\ \cline{2-4}
                      & \hspace{1em}Misspellings  & 74  \%        & 74  \%        \\ \cline{2-4}
{\bfseries Baseline}  & \hspace{1em}Run-ons       & 100 \%        & 100  \%       \\ \cline{2-4}
                      & \hspace{1em}Splits        & 85  \%        & 65  \%        \\ \cline{2-4}
                      & Total                     & 80  \%        & 77  \%        \\ \cline{2-4}
\hline
\end{tabular}
}

\caption{Baseline experiment\label{tab:bas}}
\end{table}

In the Baseline experiment (Table~\ref{tab:bas}) there is an 80\%
total recall. The drop in precision is quite small which is not
surprising since there is no language model to `disturb' the
Orthographic Decoder. The 80\% $\rightarrow$ 77\% drop is altogether
due to the bad splits precision. In a handful of places in the corpus
there are double space characters inbetween words. Since the LD does
not add a cost to the forming of words, the superfluous space will be
changed to a single character word such as \xpl{,}. The double space
in the input utterance does not constitute an error by our definition,
so an error is introduced and the error is classified in terms of the
transformation from input to output utterance, in this case a
split. For example: \xpl{\ldots models\symbol{32}\symbol{32}and\ldots}
$\rightarrow$ \xpl{\ldots models,\symbol{32}and\ldots}.

When the LD is furnished with the Unigram language model
(Table~\ref{tab:uni}) performance is enhanced on all categories. The
total enhancement (80\% $\rightarrow$ 86\%) compared to the Baseline
is due to improved ability to deal with misspellings and splits. On
four accounts the Unigram model was able to make the right decision on
`close calls' regarding misspellings that the Baseline failed to deal
with.

\begin{table}[ht]
\centerline{
\begin{tabular}{||c|l|c|c||}
\hline
Experiment           & Performance categories        & Recall        & Precision     \\ \cline{1-4}
                     & Utterances                    & 82  \%        & 76 \%         \\ \cline{2-4}
                     & \hspace{1em}Misspellings      & 81  \%        & 76  \%        \\ \cline{2-4}
{\bfseries Unigram}  & \hspace{1em}Run-ons           & 100 \%        & 81 \%         \\ \cline{2-4}
                     & \hspace{1em}Splits            & 92  \%        & 92  \%        \\ \cline{2-4}
                     & Total                         & 86  \%        & 79 \%         \\ \cline{2-4}
\hline
\end{tabular}
}

\caption{Experiments with the Unigram language model\label{tab:uni}}
\end{table}

\begin{table}[ht]
\centerline{
\begin{tabular}{||c|l|c|c||}
\hline
Experiment                     & Performance categories    & Recall        & Precision     \\ \cline{1-4}
                               & Utterances                & 86 \%         & 79 \%         \\ \cline{2-4}
                               & \hspace{1em}Misspellings  & 89 \%         & 79 \%         \\ \cline{2-4}
{\bfseries Domain-Tag Bigram}  & \hspace{1em}Run-ons       & 100 \%        & 85 \%         \\ \cline{2-4}
                               & \hspace{1em}Splits        & 100 \%        & 100 \%        \\ \cline{2-4}
                               & Total                     & 92 \%         & 83 \%         \\ \cline{2-4}
\hline
\end{tabular}
}

\caption{Experiments with the domain-tag Bigram language model\label{tab:bidom}}
\end{table}

Both the tag Bigram experiments
(Tables~\ref{tab:bidom}~and~\ref{tab:bisuc}) show steady improvement
over both the Baseline and the Unigram\footnote{Due to the small
  test-set it is difficult to show statistically significant
  improvements from one language model to another. The only difference
  that can be statistically confirmed is that both tag Bigram language
  models are significantly better than the Baseline. This was shown
  with a $\chi^{2}$-test on the 0\@.05 level using the numbers for
  total recall.}. Mutually however, between the
domain-tag Bigram and the POS Bigram, there is not much
difference. POS Bigram seems to have a narrow advantage with respect
to precision, but the two tag Bigram language models exhibit virtually
the same results. The advantage that POS Bigram has because of the
richer class-set is possibly neutralized by the poorer estimates
resulting from the added data sparseness problem. If the result that
domain classes yield as good performance as syntactic classes would
extrapolate to a bigger corpus, we would consider this a positive
result in the context of a dialogue system since the interpretation
step (input query $\rightarrow$ SQL-query) is substantially reduced by
the domain-classification of input words.

\begin{table}[htb]
\centerline{
\begin{tabular}{||c|l|c|c||}
\hline
Experiment              & Performance categories   & Recall        & Precision     \\ \cline{1-4}
                        & Utterances               & 89 \%         & 82 \%         \\ \cline{2-4}
                        & \hspace{1em}Misspellings & 89 \%         & 83 \%         \\ \cline{2-4}
{\bfseries POS Bigram}  & \hspace{1em}Run-ons      & 100 \%        & 77 \%         \\ \cline{2-4}
                        & \hspace{1em}Splits       & 100 \%        & 100 \%        \\ \cline{2-4}
                        & Total                    & 92 \%         & 84 \%         \\ \cline{2-4}
\hline
\end{tabular}
}

\caption{Experiments with the POS Bigram language model\label{tab:bisuc}}
\end{table}

\textsc{cars} contains some `impossible' lexical errors. Examples of
these are:
\begin{list}{\texttt ==>}{\setlength{\rightmargin}{2cm}}
        \inputitem{utterance}\hgl{s}\label{eq:s}
        \inputitem{utterance}total-cost per mile \hgl{ins} of rust and
        value-decrease \hgl{ins} of motor-strength\label{eq:ins}
        \inputitem{utterance}choose the best three with respect to
        fuel-consumption total-cost and \hgl{value-decs}\label{eq:decs}
\end{list}
Utterance~(\ref{eq:s}) is a strange single character
utterance. \prognm{ctr} suggested \xpl{so} as a repair, but we had
decided that the subject probably meant
\xpl{show}. Utterances~(\ref{eq:ins})~and~(\ref{eq:decs}) are both the
work of one particular subject. The subject is obviously making up new
abbreviations. The two instances of \xpl{ins} should both be
\xpl{instead} and \xpl{value-decs} should be \xpl{value-decrease}. If
it were not for these four errors, the total recall performance for the
Bigram models would be around 96\%.

The \textsc{cars} corpus is quite small and compared to normal text
standards (not just dialogue texts) the language in \textsc{cars} is
highly irregular, full of ellipses and other oddities. This together
with the data shortage and partitioning scheme makes the language
model parameter estimation \emph{very} unreliable. On the other hand
there are virtually no real-word errors in the corpus, and here is
where a reliable language model is needed the most. The language models used
here are obviously useful in distinguishing between correction
alternatives, but the limited vocabulary of \textsc{cars} also has a
positive effect on this problem since there are relatively few
alternatives to consider. The conclusion must be that it is necessary
to test \prognm{ctr} on a larger corpus, with more training material
and a larger vocabulary.

\section{\sc secretary}
\label{sec:ee-secretary}
Eight secretaries at the Department of Computer and Information
Science were given the task to transcribe a portion of a software
manual~\cite{os/2:93} written in English with the purpose of acquiring
an error corpus. The software manual is the IBM OS/2 2\@.1
Installation Guide and the transcription part is an excerpt from pages
4-18 to 4-24. The preface and chapters one, two, three and chapter
four up to page 4-18 are used as training material in the experiments
reported below and the seven page excerpt is used for testing. The
paper copy of the excerpt given to the secretaries was freed of
formatting except for headings and paragraph delimiters. The original
text contains a lot of instruction lists formatted as enumerations,
for example:
\begin{description}
\item[{\mdseries\texttt{1.}}] \texttt{\ldots}
\item[{\mdseries\texttt{2.}}] \texttt{Press Enter to display the Options menu.}
\item[{\mdseries\texttt{3.}}] \texttt{Select Set startup values and press Enter.}
\item[{\mdseries\texttt{4.}}] \texttt{\ldots}
\end{description}
The unformatted text (relieved of the numbers) looks quite strange to
the subjects and most of the subjects said after the task had been
completed that it was hard to make any sense out of the text. Another
reason for this is that the subjects are unfamiliar with the topic.

Each of the subjects was instructed to transcribe half the
excerpt. Hence the error corpus includes four versions of the original
text. The only additional instruction given to the subject was that
``you should type as fast as you can''. The reason for this was that
we wanted a larger error sample than we would presumably otherwise
get. The subjects were not told of the purpose of the transcription
but some of them expressed the suspicion that something along the line
of error sampling was going on. All subjects used the correct
positioning of the hands on the keyboard but otherwise their
typewriting skills differed quite substantially. The time it took to
transcribe the text ranged from 16 minutes to 41 minutes and while the
most faultless typist introduced no errors at all, one made 39
spelling- and segmentation errors.

It is difficult to make any systematic comparisons regarding
\prognm{ctr}'s behavior on \textsc{cars} and \textsc{secretary}. Even
if all the free variables\footnote{Free variables such as, for the OD:
  the network topology, error generating functions used to produce the
  error corpus, smoothing of the observation distribution and
  filtering versus no filtering for real-word errors. For the LD:
  tag-set, smoothing parameters, supervised versus unsupervised
  training.} are fixed, any direct comparison
will still only be approximate; there are obviously different errors
in the two corpora, the corpora are written in different languages and
consequently the tag-sets differ. The size of the vocabulary
differs and so on.

In the experiments reported below the free variables will be kept fixed
as much as possible to facilitate for some comparative studies of
\textsc{cars} and \textsc{secretary} although the primary intention with
the \textsc{secretary} experiment is to \emph{get a better handle on
the error correcting performance} of \prognm{ctr}. We have roughly
ten times as much training material (which should give a better
language model), we have more errors and a larger vocabulary. To sum
up, it is a more realistic scenario. What \prognm{ctr} can do with the
real-word errors is also interesting. We have already concluded that
the tag Bigram language model outperforms the Unigram language model,
even with unreliable parameter estimation, so the experiments below
will only concern the Baseline and the tag Bigram.

\subsection{Error Profile}
\label{subsec:ee-secretary-ep}
The name \textsc{secretary} refers to the error corpus, the four
transcribed versions of the seven page excerpt of the manual. Although
this text has been typed by eight different subjects and is a four
times duplicate, it is regarded as one bulk of text below.

\textsc{secretary} contains 600 sentences and 8,938 word tokens. The
sentence level error rate is (somewhat surprisingly) almost as high as
that of \textsc{cars}, whereas the word error rate is considerably
lower. The figures are presented in Table~\ref{tab:sec-sen}.

\setlength{\extrarowheight}{2pt}
\begin{table}[ht]
\centerline{
\begin{tabular}{||ll|rr|rr||}
\cline{3-6}
\multicolumn{2}{c}{} & \multicolumn{2}{|c}{Sentences} & \multicolumn{2}{c||}{Word Tokens} \\ \hline
\multicolumn{2}{||l|}{Well-formed}           & 483 & 80\@.5\% & 8,788 & 98\@.3\% \\ 
\multicolumn{2}{||l|}{Lexically Ill-formed}  & 117 & 19\@.5\% & 150  &  1\@.7\% \\ \hline\hline
\multicolumn{2}{||l|}{{\bfseries Total}}     & {\bfseries 600} & {\bfseries 100\%} & {\bfseries
  8,938} & {\bfseries 100\%} \\ 
\hline
\end{tabular}
}
\caption{Error profile overview of \textsc{secretary}\label{tab:sec-sen}}
\end{table}

There is one important difference between \textsc{cars} and
\textsc{secretary} regarding the way in which the error profiles have
been produced. In the case of \textsc{cars} we have been forced to
make subjective judgments as to what was the `intended', or,
`correct' utterance. With \textsc{secretary} we do not have to do this
since we have the original text. We know the correct way to type the
text down to the last comma. This is of course practical, but it also
makes it necessary to make some distinctions. Based on the scenario,
the application, we assume that all differences between the
transcriptions and the original text are lexical errors. The
variations that are clearly not lexical errors are not considered. Our
interest is in studying the error correcting performance of
\prognm{ctr}.

On two occasions a whole chunk of text was omitted. The subject most
likely looks at the text, looks up at the screen, and then back at
the paper and continues typing from the wrong place. This sort of
phenomenon is not considered here. Probably because of the sparse
formatting in the original text, problems regarding sentence ending
punctuation are quite frequent. Sentence ending punctuations were
wrongfully deleted from regular sentences, and inserted into
subheadings. This sort of error is not considered here. Apart from
these two exceptions, string equality is the measure used to find the
errors introduced into the transcribed text. Because of the nature of
the task presented to the subjects, all cases of substitutions of one
word for another are considered real-word errors, even if they are
really agreement errors.

\begin{table}[ht]
\centerline{
\begin{tabular}{||ll|rr|rr|>{\bfseries}r>{\bfseries}r||}
\cline{3-8}
\multicolumn{2}{c}{}& \multicolumn{2}{|c}{Nonword error} & \multicolumn{2}{c}{Real-word
  error} & \multicolumn{2}{c||}{{\bfseries Total}}\\ \hline
& Single error                                              & 109 & 85\@.2\% & 19 & 86\@.4\% & 128 & 85\@.3\% \\
\raisebox{1.5ex}[0cm][0cm]{Missp.} & Multiple error   & 5   &  3\@.9\% &  2 &  9\@.1\% &   7 &  4\@.7\% \\
\hline
& Single error                                              & 14  & 10\@.9\% &  0 &      0\% &  14 &  9\@.3\% \\
\raisebox{1.5ex}[0cm][0cm]{Run-ons}      & Multiple error   & 0   &      0\% &  0 &      0\% &   0 &      0\% \\
\hline
& Single error                                              & 0   &      0\% &  1 &  4\@.5\% &   1 &  0\@.7\% \\
\raisebox{1.5ex}[0cm][0cm]{Splits} & Multiple error         & 0   &      0\% &  0 &      0\% &   0 &      0\% \\
\hline\hline
& Single error                                              & 123 & 96\@.1\% & 20 & 90\@.9\% & 143 & 95\@.3\% \\
Total & Multiple error                             & 5   &  3\@.9\% &  2 &  9\@.1\% &   7 &  4\@.7\% \\
& {\bfseries Total}             & {\bfseries 128} & {\bfseries 100\%} & {\bfseries 22} & {\bfseries 100\%} & {\bfseries 150} & {\bfseries 100\%} \\
\hline
\end{tabular}
}
\caption{Breakdown of lexical errors in \textsc{secretary}\label{tab:sec-lexcloseup}}
\end{table}

Table~\ref{tab:sec-lexcloseup} shows how the lexical errors are
distributed in \textsc{secretary}. The `easy' errors, the nonword
single error misspellings, make up a relatively large portion of the
errors in \textsc{secretary} compared to \textsc{cars}. Overall, the
figures in Table~\ref{tab:sec-lexcloseup} are more in line with what
others have found; there are more real-word errors, fewer segmentation
errors and there are more `easy' errors. \prognm{ctr}'s performance on
real-word errors was not really put to the test in the \textsc{cars}
experiments, but in \textsc{secretary} there is a sample to test the
recovery abilities of \prognm{ctr} on this error type.

There is a relatively large portion of really hard real-word errors in
\textsc{secretary}. Eight out of 22 real-word errors would be
impossible even for a human proof-reader to detect. An example:
\begin{list}{\textrm{\textsc{sec:}}}{\setlength{\rightmargin}{.5cm}}
        \outputitem{utterance}If you \hgl{select} Yes for Timer,
        indicate how long you want the menu displayed before the
        default operating system is started.\label{utt:ee-impossible}
\end{list}
The proper way to type this sentence, according to the original text,
would be to substitute \xpl{selected} for \xpl{select}. This is, of
course, a very harsh correctness criterion, but the task given to the
subject was to transcribe the text, not to convey the general meaning
of the text. There are examples of real-word errors that are not
impossible to detect, but still very hard to handle:
\begin{list}{\textrm{\textsc{sec:}}}{\setlength{\rightmargin}{.5cm}}
        \outputitem{utterance}Specifying Options for the OS/2 2\@.1
        Partition \hgl{of} Logical Drive\label{utt:ee-hard}
\end{list}
\xpl{of} in sentence~(\ref{utt:ee-hard}) should be \xpl{or}.

\begin{table}[ht]
\centerline{
\begin{tabular}{||lr|c|c|c|c|c||}
\cline{3-7}
\multicolumn{2}{c}{} & \multicolumn{4}{|c|}{Single Errors}  & \raisebox{-.4ex}[0cm][0cm]{Multiple}\\ \cline{3-6}
\multicolumn{2}{c|}{} & Del & Ins & Sub & Tra & \raisebox{.4ex}[0cm][0cm]{Errors} \\ \hline
Misspellings & 135 & 48 & 30 & 29 & 21 & 7 \\ \hline
Run-ons      &  14 & 14 &  0 &  0 &  0 & 0 \\ \hline
Splits       &   1 &  0 &  1 &  0 &  0 & 0 \\ \hline\hline
{\bfseries Total} & {\bfseries 150} & {\bfseries 41\@.3\%} &
{\bfseries 20\@.7\%} & {\bfseries 19\@.3\%} & {\bfseries
  14\@.0\%} & {\bfseries 4\@.7\%} \\ \hline
\end{tabular}
}
\caption{Comparison of the basic error types in \textsc{secretary}\label{tab:sec-primitive}}
\end{table}

The sort of sloppiness that was found in \textsc{cars}
(see utterance~(\ref{utt:prob-topevo}) in
Section~\ref{sec:corpus-cars}) is not present in \textsc{secretary} in
the same way. This shows in the lower multiple error rate of
\textsc{secretary}. Table~\ref{tab:sec-primitive} displays the low
multiple error rate and how the singletons are distributed over the
basic error types; deletion, insertion, substitution and
transposition. The ranking order of the basic error types is the
same as that found
by~Pollock and Zamora~\shortcite{pollock+zamora:83} in
their sample of 50,000 nonword misspellings.

It is not easy to say which corpus, \textsc{cars} or
\textsc{secretary}, is the more demanding from the point of view of
error recovery. \textsc{secretary} has more single error nonword
misspellings and less segmentation errors, but then, judging from the
results with the \textsc{cars} experiment, segmentation errors do not
seem to be much of a problem for \prognm{ctr}. \textsc{cars} has an
18\% multiple error rate while \textsc{secretary} only has 4\@.7\% and
this indicates that \textsc{cars} is more difficult. On the other hand
\textsc{secretary} has a 14\@.7\% real-word error rate compared to
5\@.5\% for \textsc{cars} and real-word errors are clearly the most
difficult error type.

\subsection{The Orthographic Decoder}
The training and test material taken together contain 1,223 word types
so the Orthographic Decoder contains 1,223 word modeling HMMs. There
are only 17 words in the test corpus that are not found in the
training corpus. In accordance with previous experiments these word
models are included in the OD\@.

The OD setup reported on in the \textsc{cars} experiment has been
evaluated here as well. However, the possible variation in the OD
setup has been somewhat more systematically evaluated in the
\textsc{secretary} experiment. We have tried both filtered and
unfiltered error corpora and three different values for
$\epsilon_{obs}$ has been tried, $10^{-4}$, $10^{-6}$ and
$10^{-8}$. Together this makes six alternative OD setups. The results
reported below concern the same setup that was used for
\textsc{cars}. The effect of the other setups are discussed in
Section~\ref{subsec:sec-res}.

In the experiments reported in Section~\ref{subsec:sec-res}, the error
corpora were generated with the error functions deletion,
substitution, and white space insertion. There are nine single
character `punctuation words' in the vocabulary and the training
material is a small set of hand-made corruptions that only involve the
space character. There are 49 numbers in the vocabulary. These special
words have corpora generated with only the white space insertion error
function. One set of ODs had their error corpus filtered for real-word
errors and the rest were trained on unfiltered corpora. The OD HMMs
were trained with the Baum-Welch reestimation algorithm. After
training the models had their observation symbol distribution smoothed
with the additive smoothing scheme with $\epsilon_{obs}=10^{-4}$,
$\epsilon_{obs}=10^{-6}$ and $\epsilon_{obs}=10^{-8}$.

\subsection{The POS Bigram Language Model}
The POS Bigram language model includes 50 tags. The tag-set consists
of traditional Part-Of-Speech with verbs and nouns subcategorized for
morphological features. Some frequent words with supposedly uniform
contextual distributions have been given their own tag. The complete
tag-set is listed in Appendix~\ref{app:A-sec}. The training corpus
consists of 19,975 word tokens. A tagged and an untagged version of it
have been used for estimating the model parameters.

Different Linguistic Decoder setups have been tried. We have used one
smoothed with $\epsilon_{trans}=10^{-3}$ and $\epsilon_{obs}=10^{-3}$
and one that was smoothed with $\epsilon_{trans}=10^{-4}$ and
$\epsilon_{obs}=10^{-4}$. Note that there is no way to analytically
determine the best smoothing value. Smoothing of unreliable
distributions is an important research topic, and to get a good
estimation of the parameter space it is necessary to use more advanced
methods than we are using here. However, we are content to see that
the techniques presented here work satisfactorily with a not so good
smoothing scheme, reassured that with a better smoothing method things
can only get better.

In the \textsc{secretary} experiments we are not that interested in
the tag sequence output from \prognm{ctr}, rather we would like to
maximize the predictive power of the (weak) language model. The LD
trained with the tagged training material have been contrasted with
the LD estimated from the untagged text using the Baum-Welch
reestimation algorithm. In Chapter~\ref{ch:rtr} it was mentioned that
the model trained with the Baum-Welch algorithm will converge to a
local maximum. Which of the optima the model will converge towards
depends (amongst other things) on the distributions of the initial
model. In the experiments reported here the initial model has a
uniform transition distribution and an observation distribution that
has uniformly distributed probabilities for the words that are members
of the tags (which are represented by the states of the model). In
other words: the initial model `knows' which words belong to which tag
and nothing else.

Together with the two smoothing setups, the supervised and
unsupervised\footnote{Supervised and unsupervised are not very precise
terms. The supervision that is provided for the LD that is extracted
from the tagged corpus applies to the proper tagging of words, not to
the prediction of the next word, which is what we are interested
in. The terms are used here because of their intuitive appeal.}
training methods yield four different LD configurations. The results
on the POS Bigram Linguistic Decoder reported in
Section~\ref{subsec:sec-res} refer to the LD that has been trained
unsupervised and has been smoothed with $\epsilon_{trans}=10^{-4}$ and
$\epsilon_{obs}=10^{-4}$. The most ambiguous word in the language
model is four ways ambiguous. In the experiments reported below
\prognm{ctr} has been evaluated using both 1-best and 4-best
search. The outcomes of these different search strategies were,
however, identical.

\subsection{Results}
\label{subsec:sec-res}
The evaluation scheme used here is the same as the one used for
\textsc{cars} except that nonword and real-word errors have their own
keys. Note that nonword and real-word errors on the one hand and
misspellings, run-ons and splits on the other hand are orthogonal,
i.e. $\mathbf{A}^{tot} = \mathbf{A}^{non}\cup \mathbf{A}^{real} =
\mathbf{A}^{m}\cup \mathbf{A}^{r}\cup \mathbf{A}^{s}$.

\begin{table}[ht]
\centerline{
\begin{tabular}{||c|l|r|r||}
\hline
Experiment           & Performance categories   & Recall       & Precision    \\ \cline{1-4}
                     & Sentences                & 51\@.3\%     & 51\@.3\%     \\ \cline{2-4}
                     & \hspace{1em}Misspellings & 54\@.1\%     & 54\@.1\%     \\ \cline{2-4}
\raisebox{-1.5ex}[0cm][0cm]{{\bfseries Baseline}} & \hspace{1em}Run-ons      &    100\%     &    100\%     \\ \cline{2-4}
                     & \hspace{1em}Splits       &      0\%     &      0\%     \\ \cline{2-4}
                     & \hspace{1em}Nonwords     &     68\%     &     68\%     \\ \cline{2-4}
                     & \hspace{1em}Real-words   &      0\%     &      0\%     \\ \cline{2-4}
                     & Total                    &     58\%     &     58\%     \\ \cline{2-4}
\hline
\end{tabular}
}

\caption{Baseline experiment\label{tab:sec-bas}}
\end{table}

The higher smoothing value worked best in the Baseline experiment,
i.e. $\epsilon_{obs}=10^{-4}$ outperformed $\epsilon_{obs}=10^{-6}$
and $\epsilon_{obs}=10^{-8}$. The filtered and the unfiltered versions
produced identical results. No errors were introduced.

There is just one split in the corpus. The split is also a real-word
error and it reads:
\begin{list}{\textrm{\textsc{sec:}}}{\setlength{\rightmargin}{.5cm}}
        \outputitem{utterance}If you \hgl{hav e} a Dual Boot partition
        containing\ldots\label{utt:ee-baseline-split}
\end{list}
\xpl{e} is a valid word in the vocabulary, it actually has three
meanings: the name of an appendix (in the manual), the name of a
disk-partition and the name of a logical
drive. Sentence~(\ref{utt:ee-baseline-split}) was changed into
\begin{list}{\textrm{\textsc{ctr:}}}{\setlength{\rightmargin}{.5cm}}
        \outputitem{utterance}if you have \hgl{e} a dual boot partition
        containing\ldots\label{utt:ee-baseline-split-change}
\end{list}
which is obviously not the desired output. (Recall that there is no
discrimination made between upper- and lowercase characters.) 

There is a considerable difference between the result of the
\textsc{cars} Baseline experiment and that of
\textsc{secretary}. Since no LD is involved and since the OD HMMs are
trained and smoothed in the same way, the only thing that can explain
the difference is the larger number of real-word errors in
\textsc{secretary} and the increased vocabulary size.

\begin{table}[ht]
\centerline{
\begin{tabular}{||c|l|r|r||}
\hline
Experiment             & Performance categories    & Recall       & Precision    \\ \cline{1-4}
                       & Sentences                 & 78\@.6\%     & 78\@.6\%     \\ \cline{2-4}
                       & \hspace{1em}Misspellings  &     80\%     & 79\@.4\%     \\ \cline{2-4}
\raisebox{-1.5ex}[0cm][0cm]{{\bfseries POS Bigram}} & \hspace{1em}Run-ons       &    100\%     &    100\%     \\ \cline{2-4}
                       & \hspace{1em}Splits        &    100\%     &    100\%     \\ \cline{2-4}
                       & \hspace{1em}Nonwords      &     93\%     &     93\%     \\ \cline{2-4}
                       & \hspace{1em}Real-words    & 18\@.2\%     & 17\@.4\%     \\ \cline{2-4}
                       & Total                     &     82\%     & 81\@.5\%     \\ \cline{2-4}
\hline
\end{tabular}
}

\caption{The unsupervised POS Bigram experiment smoothed with
  $\epsilon_{trans}=10^{-4}$ and $\epsilon_{obs}=10^{-4}$\label{tab:sec-bipos}}
\end{table}

When \prognm{ctr} is supplied with the POS Bigram LD, there is a
considerable boost in performance\footnote{A $\chi^{2}$-test showed
significance compared to Baseline on the 0\@.001 level.}. The OD used
in the experiment in Table~\ref{tab:sec-bipos} is that which has been
filtered for real-word errors.

The fact that the LD smoothed with the lower value outperforms the one
with the higher smoothing value indicates that the parameter values
arrived at by the Baum-Welch algorithm are not such a bad
estimation. The filtered and the unfiltered version give virtually the
same results. On one occasion the unfiltered version succeeded in
correcting a real-word error that the filtered version failed to
correct. However, the unfiltered version also introduced a couple of
errors so the net return of the filtered version is slightly better.

The real-word errors that \prognm{ctr} can handle are those where
there is an orthographic similarity between the error and the proposed
normalization, and, the real-word error is part of an unlikely tag
sequence. An example of a real-word error that \prognm{ctr}
successfully transformed is:
\begin{list}{\textrm{\textsc{sec:}}}{\setlength{\rightmargin}{.5cm}}
        \outputitem{utterance}\hgl{Of} you select advanced, your Boot
        Manager\ldots\label{utt:ee-real-word}
\end{list}
\prognm{ctr} normalized the sentence to
\begin{list}{\textrm{\textsc{ctr:}}}{\setlength{\rightmargin}{.5cm}}
        \outputitem{utterance}if you select advanced, your boot
        manager\ldots\label{utt:ee-real-word-corr}
\end{list}
which was the desired output.

The unsupervised LD performs better than the supervised. The optimum
reached under the restrictions imposed by the initial model with the
Baum-Welch algorithm is a minimum entropy point. The fact that this
model outperforms the model with the higher entropy is by no means
surprising.

The nonword error correcting rate of \textsc{secretary} is the same as
the total correction rate of \textsc{cars} (which basically only
contains nonword errors). One can speculate that the higher degree of
multiple errors in \textsc{cars} is compensated for by the larger
vocabulary in \textsc{secretary}.

A closer look at the errors that \prognm{ctr} failed to
properly correct revealed that an unrepresentative portion of the
problematic errors were transpositions. Since the transposition error
function was excluded from the functions that generated the training
corpora for the various ODs, we ran a series of complementary
experiments where every parameter was held stationary except that the
transposition error function was included in the generation of the
corpora. The result is shown in Table~\ref{tab:sec-bipos+tra}.

\begin{table}[ht]
\centerline{
\begin{tabular}{||c|l|r|r||}
\hline
Experiment             & Performance categories    & Recall       & Precision    \\ \cline{1-4}
                       & Sentences                 & 81\@.2\%     & 81\@.2\%     \\ \cline{2-4}
                       & \hspace{1em}Misspellings  &     83\%     & 82\@.4\%     \\ \cline{2-4}
\raisebox{-1.5ex}[0cm][0cm]{{\bfseries POS Bigram}} & \hspace{1em}Run-ons       &    100\%     &    100\%     \\ \cline{2-4}
                       & \hspace{1em}Splits        &    100\%     &     50\%     \\ \cline{2-4}
                       & \hspace{1em}Nonwords      & 94\@.5\%     & 94\@.5\%     \\ \cline{2-4}
                       & \hspace{1em}Real-words    & 27\@.3\%     &     25\%     \\ \cline{2-4}
                       & Total                     & 84\@.7\%     & 83\@.6\%     \\ \cline{2-4}
\hline
\end{tabular}
}

\caption{The unsupervised POS Bigram experiment smoothed with
  $\epsilon_{trans}=10^{-4}$ and $\epsilon_{obs}=10^{-4}$ where the OD
  has been trained on transpositions\label{tab:sec-bipos+tra}}
\end{table}

\section{Discussion}
\label{sec:ee-concl}
Clearly the \prognm{ctr} system can be used to normalize text input to
the vocabulary and language of a limited domain. The number of
utterances affected by lexical errors in the dialogue scenario are
brought down from 71 ($19.2\%$) to 13 ($3.5\%$) by \prognm{ctr} (POS
Bigram). In the transcription scenario, the error rate is brought down
from 117 ($19.5\%$) lexically ill-formed sentences to 22 ($3.7\%$)
(from the experiment in Table~\ref{tab:sec-bipos+tra}). Counting only
the nonword errors, there are only 7 (1\@.2\%) output sentences that
diverge from the key.

The results reported in the previous sections validates the assertions
made in section~\ref{sec:corpus-concl}. The wide error scope is
obviously beneficial. Particularly the dialogue scenario with its many
segmentation errors would be severly crippled without \prognm{ctr}'s
ability to handle run-ons and splits. The total recall would fall from
92\% to 59\% in both the \textsc{cars} Bigram experiments if neither
real-word errors nor segmentation errors could be fixed. The advantage
derived from modeling of the local context is obvious when comparing
the experiments in the two scenarios to their respective Baselines.

A rather ill-chosen, but still interesting comparison can be made
between the performance of \prognm{ctr} and commercially available
text processing tools. We ran the test data through the spell-checker
used in Microsoft \prognm{word} for Windows 95\footnote{The
spell-checker used in \prognm{word} is International Correct Spell
from INSO Corporation.}. Before testing, the spell-checker was given
the complete \textsc{secretary} vocabulary. We took the highest ranked
correction candidate from the spell-checker to be the suggested
correction. It is unfair to compare \prognm{ctr} to the \prognm{word}
spell-checker for two reasons: Firstly, the spell-checker is not
designed to be an automatic spelling corrector, its primary task is to
detect errors and bring the user's attention to them. Secondly, the
spell-checker has a vocabulary that is considerably larger than
\prognm{ctr}'s. There are more correction candidates to consider and
on ten occasions it turned out that errors that were nonwords relative
\prognm{ctr}'s vocabulary were missed because they were valid words in
\prognm{word}'s vocabulary.
\begin{table}[ht]
\centerline{
\begin{tabular}{||c|l|r|r||}
\hline
Experiment             & Performance categories    & Recall       & Precision    \\ \cline{1-4}
                       & Sentences                 & 49\@.6\%     & 42\@.3\%     \\ \cline{2-4}
                       & \hspace{1em}Misspellings  &     57\%     &     57\%     \\ \cline{2-4}
\raisebox{-1.5ex}[0cm][0cm]{{\bfseries MS Word}}    & \hspace{1em}Run-ons       &      0\%     &      0\%     \\ \cline{2-4}
                       & \hspace{1em}Splits        &      0\%     &      0\%     \\ \cline{2-4}
                       & \hspace{1em}Nonwords      & 59\@.4\%     & 59\@.4\%     \\ \cline{2-4}
                       & \hspace{1em}Real-words    &      0\%     &      0\%     \\ \cline{2-4}
                       & Total                     & 51\@.3\%     & 45\@.3\%     \\ \cline{2-4}
\hline
\end{tabular}
}

\caption{Microsoft \prognm{word} Experiment\label{tab:ms-word}}
\end{table}
The results from the experiment with the \prognm{word} spell-checker
is displayed in Table~\ref{tab:ms-word}. The spell-checker works fine
with the nonword misspellings but is incapable of handling any of the
segmentation- and real-word errors. On 20 occasions the spell-checker
stopped on something that was not in error, and was unable to suggest
a correction. These problems most certainly arise from faulty
assumptions made by the spell-checker's tokenizer. It will highlight
items like \xpl{Ctrl+Alt+Del}, assuming that it is one word. It is
simply the case that \xpl{+} does not delimit tokens in
\prognm{word}. Nevertheless, it actually performs slightly better than
the Baseline (without transposition-training) on the misspelling error
category.

It was pointed out above that there is virtually no other research
effort taking the holistic approach presented here, addressing the
entire problem area in a unified framework that uses both a model of
language production and one for typing behaviour and which makes
tokenization part of the recovery process. So the results presented
here do not lend themselves easily to comparisons to what others
have done. (Another major problem is, of course, that people uses
different test sets.) However, a couple of notes can be made.

As mentioned above, Kukich~\shortcite{kukich:92b}
made a comparative study of some of the more well-known approaches to
isolated-word spelling correction on 170 human-generated nonword
misspellings with vocabularies of three different sizes. Two of the
vocabularies, 521 words and 1,142 words, are quite close in size to
the vocabularies used in \textsc{cars} and \textsc{secretary} (584 and
1,223 words respectively). The OD component of \prognm{ctr} compares
quite favorably to the best isolated-word spelling correction
techniques. On the smaller vocabulary~Kukich reports
81\% accuracy for the best technique which is just about the same as
for the \textsc{cars} Baseline. The \textsc{secretary} Baseline
(without transposition-training) result on the nonwords is 68\% and
the result from the best isolated-word spelling correction technique
is 78\%\footnote{A technique called SVD (Singular Value Decomposition)
worked best on the smaller vocabulary. The accuracy for the different
programs ranged from 64\% to 81\%. The Technique of
Kernighan-Church-Gale~\cite{church+gale:91a,kernighan+al:90} worked
best on the larger vocabulary. The accuracy for the different programs
ranged from 54\% to 78\%.}. Note that the isolated-word spelling
correction programs do not need to tokenize the input, they are given
one word at a time whereas \prognm{ctr} need to find the word
boundaries by itself. The conspicuous fall in accuracy when the size
of the vocabulary grows gives yet another clear indication of the
positive impact of small vocabularies.

\prognm{ctr} does not compare well to other automatic spelling
correction techniques in terms of processing efficiency. The main
reason for the difference in speed is due to the fact that
\prognm{ctr} does not make any hard assumptions regarding the location
of word boundaries. However, \prognm{ctr} has the feature of character
incremental processing. In the dialogue application, which was the
first intended usage, this means that \prognm{ctr} performs real-time
text recognition, i.e. it processes the input as fast as the user can
type. With the 1,223 word vocabulary and \emph{without Beam Search
pruning} \prognm{ctr} processes the input (on a SUN Sparcstation~5)
with approximately the speed with which a skilled typewriter would
type it.

The results from these first two experiments are certainly
promising, all the more so since there is, on several accounts,
obviously room for improvement.

The transposition error type is a weak spot in the Orthographic
Decoder. The topology of the word models prohibits the transposition
errors to be processed as transposition errors. The lack of
back-chaining transitions in the HMM means that a word containing a
transposition error will be `diagnosed' as having a deletion
immediately followed by an insertion. This topology-related problem
can not be completely trained away. The OD trained on transpositions
(Table~\ref{tab:sec-bipos+tra}) could correct three transpositions and
one deletion that the other OD could not handle. (Two out of these four
were also real-word errors.) This shows that training certainly pays
off, but still, the restrictions imposed by the topology of the model
makes matters more difficult. There were three errors that the
\prognm{word} spell-checker managed to fix that \prognm{ctr} failed
on. These three were all transposition errors, \xpl{of} had been
spelled \xpl{\hgl{fo}} (on three occasions). Giving up the
left-to-right model topology may very well improve the system's
ability to deal with this error type.

Another important aspect concerning the OD is, of course, the way in
which it is trained. The only information available to the system
regarding the \emph{cause} of an error is the keyboard layout and even
this information is rather
sparse\footnote{Grudin~\shortcite{grudin:83} has
reported on a comprehensive study of errors produced by typists of
different skill-levels. The fact that the keypads are placed in close
proximity to each other can just partly explain all the things
that can go wrong.}. The generation of the error corpora for the OD
HMMs is entirely based on lexical errors that are accidental by
nature. More knowledge can be supplied to the system by studying
naturally occuring errors, especially those caused by cognitive and/or
phonetic misconceptions. Content words and function words should
probably not be treated alike in this respect. People (at least
adults) usually know how to spell the function words and consequently
the errors that affect this category are more likely to be mistakes.

The Linguistic Decoder is the most interesting component of the system
from the improvement point of view. It has been
shown~\cite{gale+church:90} that the additive smoothing scheme is
inferior to a number of alternative, more sophisticated smoothing
methods. We know that a better smoothing scheme will improve the LD in
\prognm{ctr}, the question is just how big the improvement will be.

The real-word error category is obviously the most
problematic. \prognm{ctr}'s ability to correct these errors hinges on
the predictive power of the Linguistic Decoder. With sparse data it is
necessary to chose a model with fewer parameters, such as a tag Bigram
language model for example. Such a model has only rather vague ideas
of what is likely to appear next in a given context. If a real-word
error happens to belong to the same tag as the intended word, the tag
Bigram is unable to detect the error. To improve the real-word error
correction rate it is necessary to deploy a more powerful language
model, a language model with lower (cross-word)
entropy. Mays~\emph{et al.}~\shortcite{mays+al:91} used the
trigram language model employed in the IBM speech recognition
project~\cite{bahl+al:83} to correct single error real-word
errors. They managed to detect and correct 73\% of the real-word
errors, and managed this with a rather primitive model of the channel
characteristics. These are reassuring results.

\chapter{Future Work}
\label{ch:fw}
\section{Practical Issues}
\label{sec:fw-pi}
There are a number of issues that need to be addressed if \prognm{ctr}
should be placed in the hands of an actual user.

One of \prognm{ctr}'s features is the incremental processing. The
process is monoto\-nous, however, so if the user goes back and edits the
input string, it will cause difficulties. Special care must be taken of
backspacing in the input.

\prognm{ctr}'s vocabulary (the OD HMMs) has a high memory space
demand. An application with tens of thousands of word-forms most
likely requires some kind of morphological processing, having a
network for each word-form is not a practicable path. Word-groups (as
described above) may marginally reduce the problem but prefixes,
suffixes and inflectional forms need probably be dealt with in a more
systematic fashion. It may prove profitable to look into Two-level
Morphology~\cite{koskenniemi:83}.

\prognm{ctr} does not discriminate between upper- and lowercase
characters which means that information is lost in the processing of
the input. This needs to be rectified should the system be used in a
more large scale application.

One way to look at \prognm{ctr} is to think of it as an intelligent
tokenizer. The system spends much effort calculating the the most
likely segmentation of the input. As the system presently runs it
spends a lot of time with this even if the input is error-free and
tokenization is trivial. There is however a simple way around this
waste of effort. The OD could easily be supplied with a more primitive
(standard) complementary tokenizer. This tokenizer would then run as
long as there is no problem, and when something goes wrong the OD would
take over. The LD would run as before, only it would receive one token
(as in Token Passing) per word-token instead of a number of tokens per
character. Note that with this technique real-word errors
would have to be spotted from the LD without any help from the OD.

\section{The Unknown Word}
\label{sec:fw-uw}
There is no general solution to the unknown word problem. Under
certain circumstances however there may be ways to limit the problem
somewhat. A short word generally has a lot of neighbors, words that
are, say, one edit distance away from it. With a longer word, say ten
characters, there are usually just a few neighbors. So if a long word
is correctly spelled and a moderately chosen beam is used, there will
be a relatively small number of viable hypotheses inside the beam. If
the long word is misspelled (not too severly), there will probably
still be a controllable number of hypotheses inside the beam. However,
if a previously unknown word is typed in, there is a good chance that
the best hypothesis is quite distant from the optimal hypothesis
(which is unknown). Since the best hypothesis controls the beam and
the best hypothesis is bad, the beam will in effect be wider than
normal and more distant neighbors will fit inside the beam. Thus, if a
long word is typed into the system and the number of viable hypotheses
inside the beam exceeds a threshold, the word is unknown. This
hypothesis should be pursued. (Note that shorter words are usually not
unkown.)

\section{Exploiting the Potential of the Framework}
\label{sec:fw-epf}
It was pointed out above that there are no restrictions in the Token
Passing framework as such as to how many layers there might be. The
layers can implement different probabilistic networks, actually they
need not even be probabilistic. The only requirement is that a layer
can communicate with its neighbors in a meaningful way. Hidden Markov
Models are appealing since they have one distribution (the
transitions) for network internal operations and one (the observables)
for communication.

Connected Text Recognition with layered HMM networks in the Token
Passing framework is quite a machinery. It might even seem like bit of
an overkill to use this rather complex system just to tokenize an
input string and output the normalization of it. However, the system
leaves room for more complex tasks to be performed as well. We stated
above that \prognm{ctr} could be told to output the tag sequence along
with the normalized utterance. An utterance like:
\begin{list}{\texttt ==>}{\setlength{\rightmargin}{.5cm}}
        \inputitem{utterance}show volvo with impact-safety higher
        than 3\label{utt:fw-epf-input}
\end{list}
would then be processed and output
as\footnote{Cf. Appendix~\ref{app:A-dom} for an explanation of the tags.}:
\begin{list}{\textrm{\textsc{ctr:}}}{\setlength{\rightmargin}{1cm}}
        \outputitem{utterance}show/CH volvo/OH with/R
        impact-safety/AH higher/VH than/R 3/VH\label{utt:fw-epf-first}
\end{list}
The portion \xpl{with impact-safety higher than 3} is obviously a
restriction that the subject wants to have placed on the Volvos that he
wants extracted from the database. It would clearly be a great help if
the system could identify such phrases. Envisage a third layer,
inbetween the word modeling Orthographic Decoder and the utterance
modeling HMM in the Linguistic Decoder, that models phrases of the
type just exemplified. The additional layer would make the LD
two-layered. This layer would consist of a number of Token Passing
networks (e.g. HMMs) that would segment the stream of words from
beneath into phrases, phrases with a meaning in the application at
hand. One network in the new layer would then be the \emph{conditional}
network. The output from \prognm{ctr} could now look like:
\begin{list}{\textrm{\textsc{ctr:}}}{\setlength{\rightmargin}{2cm}}
        \outputitem{utterance}show/CH volvo/OH [with/R
        impact-safety/AH higher/VH than/R 3/VH]/COND\label{utt:fw-epf-second}
\end{list}
One can also imagine a fourth layer (making the LD three-layered) that
classifies utterances in terms of their dialogue function (see
e.g.~J{\"o}nsson~\shortcite{Jonsson:93}). This fourth
layer would be the new topmost layer and the output could look like:
\begin{list}{\textrm{\textsc{ctr:}}}{\setlength{\rightmargin}{.5cm}}
        \outputitem{utterance}[show/CH volvo/OH [with/R
        impact-safety/AH higher/VH than/R
        3/VH]/COND]/EXTRACT\label{utt:fw-epf-third}
\end{list}
The step from~(\ref{utt:fw-epf-third}) to the SQL-query below is not
very long to take.
\begin{tabbing}
\texttt{select} \=manufacturer.model.year from CARS w\kill \\
\texttt{select}\>\texttt{manufacturer.model.year.impact-safety from CARS where}\\
\>\texttt{model = `volvo' and} \\
\>\texttt{impact-safety > 3} \\
\end{tabbing}
The three fields \xpl{manufacturer}, \xpl{model} and \xpl{year}
together make up the unique identifier for each car in the database.
These fields are listed in the output by default while the other
fields referred to in the question are appended.

\bibliographystyle{named}

\appendix
\chapter{Tag-sets Used in \sc ctr}
\label{app:A}

\section{The Domain-Tags used in \sc cars}
\label{app:A-dom}
\begin{itemize}
\item 19 tags
\item 584 words
\item 55 words are ambiguous
\begin{itemize}
\item 51 words are two ways ambiguous
\item 4 words are three ways ambiguous
\end{itemize}
\end{itemize}

\setlongtables
\begin{longtable}[l]{||l|l|l||}
\hline Tag & Explanation & Example\\ \hline
\endfirsthead
\hline\multicolumn{3}{||l||}{\small \slshape continued from previous page}\\
  \hline Tag & Explanation & Example\\ \hline
\endhead
\hline\multicolumn{3}{||r||}{\small \slshape continued on next
  page}\\\hline
\endfoot
\hline
\endlastfoot
AH         & Aspect Head              & \xpl{fuel-consumption} \\
CC         & Coordinating Conjunction & \xpl{and} \\
CH         & Communicative Head       & \xpl{find} \\
CM         & Communicative Modifier   & \xpl{help} \\
K          & Non-sentence Delimiters  & \xpl{,} \\
M          & Determiner               & \xpl{each} \\
N          & Numeral                  & \xpl{4} \\
OH         & Object Head              & \xpl{mazda 323} \\
OW         & Object Wh-word           & \xpl{which} \\
P          & Sentence Delimiters      & \xpl{?} \\
R          & Relation Word            & \xpl{instead} \\
RS         & Response Word            & \xpl{yes} \\
SH         & Semiotic Head            & \xpl{mean} \\
SM         & Semiotic Modifier        & \xpl{not} \\
TH         & Table Head               & \xpl{table} \\
VH         & Value Head               & \xpl{0,9} \\
VM         & Value Modifier           & \xpl{about} \\
VW         & Value Wh-word            & \xpl{how} \\
X          & Others                   & \xpl{thanks} \\
\end{longtable}

\section{The POS used in \sc cars}
\label{app:A-pos}
\begin{itemize}
\item 31 tags
\item 584 words
\item 31 words are ambiguous
\begin{itemize}
\item 30 words are two ways ambiguous
\item 1 word is three ways ambiguous
\end{itemize}
\end{itemize}

\begin{longtable}[l]{||l|l|l||}
\hline Tag & Explanation & Example\\ \hline
\endfirsthead
\hline\multicolumn{3}{||l||}{\small \slshape continued from previous page}\\
  \hline Tag & Explanation & Example\\ \hline
\endhead
\hline\multicolumn{3}{||r||}{\small \slshape continued on next
  page}\\\hline
\endfoot
\hline
\endlastfoot
AB           & Adverb                    & \xpl{quickly} \\
CIT          & Citation Mark             & \verb!`'''! \\
COM          & Comma                     & \xpl{,} \\
DSH          & Dash                      & \xpl{-} \\
DT           & Determiner                & \xpl{all} \\
HA           & Wh Adverb                 & \xpl{why} \\
HD           & Wh Determiner             & \xpl{which} \\
HP           & Wh Pronoun                & \xpl{what} \\
IM           & Infinitive Marker         & \xpl{to} \\
IN           & Interjection              & \xpl{ok} \\
JJ           & Adjective                 & \xpl{lower} \\
KN           & Coordinating Conjunction  & \xpl{both} \\
LP           & Left Parenthesis          & \xpl{(} \\
NN           & Noun                      & \xpl{car} \\
PC           & Participle                & \xpl{enumerated} \\
PM           & Proper Noun               & \xpl{audi 100} \\
PN           & Pronoun                   & \xpl{these} \\
PNO          & Object Pronoun            & \xpl{them} \\
PNS          & Subject Pronoun           & \xpl{you} \\
PP           & Preposition               & \xpl{with} \\
PRT          & Particle                  & \xpl{away} \\
PS           & Possessive Pronoun        & \xpl{their} \\
QUE          & Question Mark             & \xpl{?} \\
REL          & Relative Marker           & \xpl{which} \\
RG           & Number                    & \xpl{1988} \\
RP           & Right Parenthesis         & \xpl{)} \\
SN           & Subordinating Conjunction & \xpl{if} \\
VBF          & Finite Verb               & \xpl{classified} \\
VBI          & Verb Infinitive           & \xpl{see} \\
VBP          & Verb Imperative           & \xpl{add} \\
VBS          & Supine Verb               & \xpl{shown} \\
\end{longtable}

\section{The POS used in \sc secretary}
\label{app:A-sec}
\begin{itemize}
\item 50 tags
\item 1223 words
\item 78 words are ambiguous
\begin{itemize}
\item 74 words are two ways ambiguous
\item 3 words are three ways ambiguous
\item 1 word is four ways ambiguous
\end{itemize}
\end{itemize}

\begin{longtable}[l]{||l|l|l||}
\hline Tag & Explanation & Example\\ \hline
\endfirsthead
\hline\multicolumn{3}{||l||}{\small \slshape continued from previous page}\\
  \hline Tag & Explanation & Example\\ \hline
\endhead
\hline\multicolumn{3}{||r||}{\small \slshape continued on next
  page}\\\hline
\endfoot
\hline
\endlastfoot
AB           & Adverb                          & \xpl{actually} \\
ACL          & Article                         & \xpl{a} \\
AND          & ---                             & \xpl{and} \\
ARE          & ---                             & \xpl{are} \\
BE           & ---                             & \xpl{be} \\
BSL          & Backslash                       & \verb!`\'! \\
COL          & Colon                           & \xpl{:} \\
COM          & Comma                           & \xpl{,} \\
CT           & Citation Mark                   & \verb!`'''! \\
DO           & ---                             & \xpl{do} \\
DSH          & Dash                            & \xpl{-} \\
DT           & Determiner                      & \xpl{both} \\
DZ           & ---                             & \xpl{does} \\
HA           & Wh Adverb                       & \xpl{how} \\
HD           & Wh Determiner                   & \xpl{which} \\
HP           & Wh Pronoun                      & \xpl{what} \\
IJ           & Interjection                    & \xpl{please} \\
IM           & Infinitive Marker               & \xpl{to} \\
IS           & ---                             & \xpl{is} \\
IT           & ---                             & \xpl{it} \\
JJ           & Adjective                       & \xpl{available} \\
KN           & Coordinating Conjunction        & \xpl{either} \\
LB           & Left Parenthesis                & \xpl{(} \\
MD           & Model Auxiliary                 & \xpl{should} \\
NEG          & Negation                        & \xpl{not} \\
NN           & Noun                            & \xpl{backup} \\
NNS          & Plural Noun                     & \xpl{actions} \\
OF           & ---                             & \xpl{of} \\
PKT          & Period                          & \xpl{.} \\
PM           & Proper Noun                     & \xpl{autoexec.bat} \\
PN           & Pronoun                         & \xpl{some} \\
PNO          & Object Pronoun                  & \xpl{them} \\
PNS          & Subject Pronoun                 & \xpl{they} \\
PP           & Preposition                     & \xpl{by} \\
PS           & Possessive Pronoun              & \xpl{its} \\
QUE          & Question Mark                   & \xpl{?} \\
RB           & Right Parenthesis               & \xpl{)} \\
REL          & Relative Marker                 & \xpl{that} \\
RG           & Number                          & \xpl{1024} \\
RO           & Ordinal Number                  & \xpl{first} \\
SN           & Subordinating Conjunction       & \xpl{after} \\
SYM          & Symbol                          & \xpl{\%} \\
THE          & ---                             & \xpl{the} \\
UO           & Unknown                         & \xpl{md} \\
VB           & Bare Verb Form                  & \xpl{change} \\
VBD          & Verb Past Tense                 & \xpl{selected} \\
VBG          & Verb ing-form                   & \xpl{according} \\
VBL          & Past Participle                 & \xpl{accessed} \\
VBZ          & Verb Third Person Present Tense & \xpl{installs} \\
YOU          & ---                             & \xpl{you} \\
\end{longtable}

%

\end{document}